\documentclass[10pt,journal,compsoc]{IEEEtran}
\ifCLASSOPTIONcompsoc
  \usepackage[nocompress]{cite}
\else
  \usepackage{cite}
\fi
\ifCLASSINFOpdf
\else
\fi
\usepackage[utf8]{inputenc} 
\usepackage[T1]{fontenc}    
\usepackage{hyperref}       
\usepackage{url}            
\usepackage{booktabs}       
\usepackage{amsfonts}       
\usepackage{nicefrac}       
\usepackage{microtype}      
\usepackage{xcolor}         
\usepackage{breqn}

\usepackage[english]{babel}
\usepackage{amsmath,amssymb}
\usepackage{mathtools}
\usepackage{enumerate}
\usepackage{comment}
\usepackage{algorithm,algorithmicx}
\usepackage[noend]{algpseudocode}
\usepackage{wrapfig}
\usepackage{subfigure}
\usepackage{multirow}
\usepackage{diagbox}
\usepackage{enumitem}
\usepackage{amsmath,bm}
\usepackage{amsthm}
\newtheorem{theorem}{Theorem}

\usepackage{bbm}
\usepackage{ulem}

\def\shownotes{1}  
\ifnum\shownotes=1
\newcommand{\authnote}[2]{[#1: #2]}
\else
\newcommand{\authnote}[2]{}
\fi

\newcommand{\E}{\mathbb{E}}
\newcommand{\Esub}[1]{\underset{#1}{\E}}

\newcommand{\cS}{\mathcal{S}}

\newcommand{\cA}{\mathcal{A}}

\newcommand{\cond}{\mathbbm{1} \left\{\left| \frac{\tilde{\pi}(a|s)}{\pi(a|s)} - 1 \right| < \delta \right\}}
\newcommand{\leftcond}{\mathbbm{1} \left\{ 1 - \frac{\tilde{\pi}(a|s)}{\pi(a|s)} < \delta \right\}}
\newcommand{\rightcond}{\mathbbm{1} \left\{ \frac{\tilde{\pi}(a|s)}{\pi(a|s)} - 1 < \delta \right\}}
\newcommand{\klcond}{\mathbbm{1} \left\{ \textup{KL} \left( \pi(a|s), \tilde{\pi}(a|s) \right)  < \rho \right\}}

\renewcommand{\paragraph}[1]{{\noindent {\bf #1.}}}

\begin{document}
%
\title{Sample Dropout: A Simple yet Effective Variance Reduction Technique in Deep Policy Optimization}

\author{Zichuan~Lin,
        Xiapeng~Wu,
        Mingfei Sun, 
        Deheng~Ye,
        Qiang~Fu,
        Wei~Yang,
        and~Wei~Liu
\IEEEcompsocitemizethanks{
\IEEEcompsocthanksitem Zichuan Lin, Xiapeng Wu, Deheng Ye, Qiang Fu, Wei Yang, and Wei Liu are with Tencent Inc., Shenzhen, China.
Email: \{zichuanlin, happenwu, dericye, leonfu, willyang\}@tencent.com, wl2223@columbia.edu. Mingfei Sun is with the University of Manchester, Manchester, United Kingdom. Email: mingfei.sun@manchester.ac.uk

\IEEEcompsocthanksitem Deheng Ye is the corresponding author. 
}
}

\IEEEtitleabstractindextext{%

\begin{abstract}
Recent success in Deep Reinforcement Learning (DRL) methods has shown that 
policy optimization with respect to an off-policy distribution via importance sampling 
is effective for sample reuse. 
In this paper, we show that the use of importance sampling could introduce high variance in the objective estimate. 
Specifically, we show in a principled way that the variance of importance sampling estimate grows quadratically with importance ratios 
and the large ratios could consequently jeopardize the effectiveness of surrogate objective optimization. 
We then propose a technique called \textit{sample dropout} to bound the estimation variance 
by dropping out samples when their ratio deviation is too high. 
We instantiate this sample dropout technique on representative policy optimization algorithms, 
including TRPO, PPO, and ESPO, 
and demonstrate that it consistently boosts the performance of those DRL algorithms on both continuous and discrete action controls, including MuJoCo, DMControl and Atari video games. 
Our code is open-sourced at \url{https://github.com/LinZichuan/sdpo.git}.

\end{abstract}

}

\maketitle

\IEEEdisplaynontitleabstractindextext

%
\IEEEpeerreviewmaketitle

\section{Introduction}
\IEEEPARstart{P}{olicy} optimization methods in deep reinforcement learning
have achieved notable success in a variety of applications, 
including complicated board games~\cite{silver2016mastering}, 
high-dimensional robotic controls~\cite{levine2016end}, 
and challenging modern video games~\cite{mnih2015human,vinyals2019alphastar,berner2019dota,ye2020mastering,lin2021juewu}. 
This family of methods features in an iterative update procedure, 
with simultaneously a \textit{surrogate objective} to optimize 
and a \textit{trust region constraint} to enforce in each iteration~\cite{kakade2002approximately,pirotta2013safe,schulman2015trust,schulman2017proximal,sun2022you}. 
The surrogate objective prescribes a practical objective to maximize, 
for which the policy is directly optimized through multiple epochs with empirical samples.
The trust region constraint, on the other hand, specifies an explicit constraint over the update step-size, 
which can be implemented as either KL divergence~\cite{schulman2015trust} or ratio deviations~\cite{sun2022you}. 
Combined, the policy optimization is guaranteed a monotonic improvement~\cite{schulman2015trust}.

One intriguing feature in the surrogate objective is the use of policy ratios. 
Specifically, the theory in the seminal work~\cite{kakade2002approximately} 
strictly requires the surrogate objective to be estimated over the on-policy distribution
-- the distribution that is induced by the policy to be optimized. 
This requirement significantly increases the sample complexity 
because the samples can only be used once in each iteration~\cite{schulman2017proximal}. 
Trust Region Policy Optimization (TRPO)~\cite{schulman2015trust} and the follow-up studies, 
including Proximal Policy Optimization (PPO)\cite{schulman2017proximal} and Early Stopping Policy Optimization (ESPO)~\cite{sun2022you}, 
leverage the \textit{importance sampling} scheme to re-formulate the objective estimate with respect to an off-policy distribution
-- the distribution from which the samples are collected, 
which yields importance ratios in the surrogate objective. 
Namely, in practical optimization, 
the surrogate objective is empirically estimated with samples generated by a different policy 
and then re-weighted via the importance weights, 
i.e., the policy ratios between the policy to be optimized and the policy used to collect samples. 
The use of this importance sampling enables the sample reuse through multiple optimization epochs 
and significantly improves the sample efficiency.

However, we show in this paper that the use of importance sampling does not come for free:
the variance introduced by this importance sampling to the objective estimate could be very high in each iteration
and would consequently jeopardize the effectiveness of the surrogate objective optimization. 
We prove in a principled way that the variance in the importance sampling estimate grows quadratically with policy ratios. 
This side-effect is detrimental to the policy optimization 
since the importance sampling subjective may not be a valid substitute for the true surrogate objective. 
The monotonic improvement guarantee can thus no longer hold even if the trust region constraint is satisfied. 
Many works point out a fact that the ratio clipping can fail to enforce the trust region~\cite{ilyas2018deep,engstrom2019implementation,wang2020truly,tomar2020mirror,sun2022you}, 
but largely overlook this detrimental high variance in the policy optimization.

We propose a simple technique to overcome the high variance in the surrogate objective estimate, called \textit{sample dropout}. 
We first present an analysis for the variance in the objective estimate
through deriving an upper bound of the variance, which involves a quadratic term of policy ratios. 
This theoretical result indicates that the variance of estimating the true surrogate objective could grow quadratically with the importance ratios. 
To alleviate this quadratic growth, 
we propose the sample dropout technique, which discards samples when their ratio deviation is too large. 
By doing so, the ratios can be bounded and consequently a low variance can be achieved for the objective estimate. 

Our technique is simple to implement and can be easily combined with existing policy optimization algorithms. 
We instantiate it on a range of representative policy optimization algorithms, including TRPO~\cite{schulman2015trust}, PPO~\cite{schulman2017proximal}, and ESPO~\cite{sun2022you}.
We evaluate our proposed method in many high-dimensional continuous control benchmarks as well as discrete-action environments 
and show that our technique consistently improves a wide range of representative policy optimization algorithms across a variety of benchmarks.

We summarize our contributions as follows:
1) We prove that the use of importance sampling could introduce high variance to the objective estimate, which potentially harms the surrogate objective optimization.
2) We propose \textit{sample dropout}, a simple yet very effective technique for variance reduction that can be easily integrated into many existing policy optimization methods. 
3) We demonstrate that the proposed method consistently improves the performance of a wide range of policy optimization algorithms across many benchmarks including MuJoCo~\cite{todorov2012mujoco}, DeepMind Control Suites~\cite{tassa2018deepmind} and Atari games~\cite{bellemare2013arcade}.
\section{Related Work}

This work is related to: 1) policy optimization methods, variance in policy gradient, and their implementation tricks; 2) experience replay methods, since we work on sample usage in reinforcement learning. 

\subsection{Policy optimization with monotonic improvement guarantee}
Conservative Policy Iteration (CPI)~\cite{kakade2002approximately} optimizes the objective defined as policy advantage (i.e., a surrogate objective) by linearly mixing the behavior policy with the learned policy. 
To relax this linear mixing mechanism, Trust Region Policy Optimization (TRPO)~\cite{schulman2015trust} uses the Total Deviation (TV) divergence between behavior policy and target policy as policy regularization and presents that the policy has guaranteed a monotonic improvement with this regularization. Similarly, Constrained Policy Optimization (CPO)~\cite{achiam2017constrained} also presents a monotonic improvement guarantee by using average TV divergence over all empirical samples. Mirror Descent Policy Optimization (MDPO)~\cite{tomar2020mirror} proposes a more generic regularization, which uses a Bregman divergence to regularize the policy update.
Unlike explicit policy regularization, Proximal Policy Optimization (PPO)~\cite{schulman2017proximal} adopts a simple ratio clipping technique which clips probability ratios to form a pessimistic surrogate objective and performs multiple epochs of policy optimization on the same set of sampled data. 
Despite the strong empirical performance of PPO, existing studies show that under this clipping mechanism, the ratios could still become unbounded~\cite{wang2020truly,engstrom2019implementation,tomar2020mirror}. 
To address this problem, existing works choose to either modify the surrogate objective or enforce the trust-region constraint.
For example, \cite{wang2020truly} proposes Truly PPO, which modifies the clipping mechanism of PPO by applying a negative incentive to prevent the policy from being pushed away during training. 
And, \cite{sun2022you} proposes ratio-regularized policy optimization algorithms with an improvement guarantee, optimizing the policy until the estimate of ratio deviation exceeds a threshold to ensure that the correction term is bounded.
By comparison, our work directly operates on samples rather than policy radio. 
To expand, while algorithms like PPO use ratio clipping to derive a lower bound of the original surrogate objective, we drop samples using the criterion we develop, which can further be combined with PPO.
Our criterion of sample dropout relies on ratio deviation. Similar to ours, ESPO~\cite{sun2022you} uses ratio deviation estimates as a condition to early stop the multi-epoch optimization. Our method differs from ESPO in the following aspects: 
1) The early-stopping mechanism in ESPO only considers the expectation of ratio deviation rather than the distribution of ratio deviation. 
Consequently, samples with small ratio deviation could be wasted due to the early-stop operation. 
2) ESPO uses a hard constraint on the correction term while we directly operate on samples. 
3) Our method has a wide application scope and can be applied in ESPO itself.
In short, our technique is orthogonal to prior works on improving policy optimization, as evidenced by the fact that it can be combined with existing policy optimization algorithms, which will be shown in the experiment section in detail. 

\subsection{Variance reduction in policy gradient}
Our work is closely related to variance reduction techniques in policy gradient methods. One simple yet effective method to reduce variance without introducing bias is through using a baseline, which has been widely studied in prior works~\cite{sutton2018reinforcement,weaver2013optimal,greensmith2004variance,schulman2015high}. The original baseline function is only dependent on the current state, which helps remove the influence of future actions from the total reward. Several works have made attempts on exploiting the factorizability of the policy probability distribution to further reduce variance. \cite{wu2018variance} incorporates additional action information into the baseline which fully exploits the structural form of the stochastic policy, hence reducing variance of the policy gradient estimator. \cite{gu2016q} makes use of an action-dependent control variate to combine the advantages of low-variance of off-policy critic and low bias of on-policy Monte Carlo returns. Some other works develop techniques to help reduce the variance of policy gradient estimates at the cost of some bias. Inspired by TD($\lambda$)~\cite{sutton1998introduction}, \cite{schulman2015high} proposes an exponentially-weighted estimator of the advantage function, denoted as generalized advantage estimation (GAE), that substantially reduces the variance of policy gradient with introduced bias. Similar to GAE, our proposed sample dropout technique also reduces variance with some bias. It is worth noting that the contribution of sample dropout is orthogonal to that of GAE --- while GAE helps trade off variance and bias by mixing various $k$-step advantage estimators, sample dropout reduces the variance by disposing samples whose ratio deviation exceeds a threshold. We also demonstrate in our experiments that our sample dropout technique can boost the sample efficiency of GAE-based policy optimization algorithms.
\cite{metelli2018policy} and \cite{metelli2020importance} propose an actor-only policy optimization algorithm that alternates online and offline optimization via important sampling. To capture the uncertainty induced by importance sampling, they propose a surrogate objective function derived from a statistical bound on the estimated performance, which helps bound the variance of surrogate objective in terms of the Renyi divergence.

\subsection{Implementation and reproducibility in policy optimization}
Apart from the above algorithmic efforts, there are also a lot of works investigating the implementation tricks in policy optimization algorithms. 
\cite{henderson2018deep} points out the reproducibility issues in RL, including the performance differences between code bases, hyperparameter tuning, and the stochasticity due to random seeds. 
Considering these concerns raised, we have verified sample dropout over a range of well-known policy optimization algorithms using different environments with a large number of random seeds. 
Recently, \cite{engstrom2019implementation} and \cite{andrychowicz2020matters} investigate code-level improvements in policy optimization methods like PPO/TRPO and conclude that the implementation tricks, such as reward scaling, orthogonal initialization, etc., are responsible for the most of the performance boost. 
Our work differs from these implementation tricks in the sense that sample dropout is motivated by the theoretical analysis on the variance of the surrogate objective estimates, resulting in a practical accelerator that can be plugged-into algorithms like PPO and TRPO, rather than tuning these algorithms themselves. 

\subsection{Policy optimization with (prioritized) sample reuse}
Our method is related to policy optimization with (prioritized) sample reuse since our technique operates on samples during training. 
Some prior works combine on-policy and off-policy data to boost the sample efficiency of policy optimization methods~\cite{wang2016sample,queeney2021generalized,oh2018self}. 
\cite{wang2016sample} proposes importance weight truncation with bias correction to enable experience replay in actor-critic methods. 
\cite{queeney2021generalized} develops policy improvement guarantees that are suitable for the off-policy data. 
\cite{oh2018self} proposes an advantage clipping technique to bias the policy towards good behaviors. 
Different from these works, sample dropout aims to improve the multi-epoch optimization for importance sampling based policy optimization methods. 
Our work is also related to prioritized experience replay~\cite{schaul2015prioritized,liang2021ptr,libardi2021guided,holubar2020continuous,sovrano2019combining} and re-weighted learning~\cite{lin2018episodic,zhu2020episodic}, which adopt heuristic algorithms to preferentially select samples for learning by increasing the sampling probability of important samples. 
Also, our work is reminiscent of prior works that use truncated importance weights to eliminate the contribution of the poorly explored data~\cite{bottou2013counterfactual}.
Different from these works, our sample dropout criterion is derived from the theoretical analysis for the variance of surrogate objective. 




\section{Preliminaries}

\subsection{Markov Decision Process}
Consider a Markov Decision Process (MDP) with state space $\cS$ and action space $\cA$. A policy $\pi(\cdot|s)$ specifies the conditional distribution over the action space given a state $s$. The transition dynamics $P(s'|s,a)$ specifies the conditional distribution of the next state $s'$ given the current state $s$ and $a$. A reward function $r:\cS \times \cA \rightarrow \mathbb{R}$ defines the reward at each step. We also consider a discount $\gamma \in (0,1)$ and an initial state distribution $\rho_0$. We define the value function $V^{\pi}: \cS \rightarrow \mathbb{R}$ at state $s$ for a policy $\pi$:
$V^{\pi}(s) = \E_{\pi,P} \left[\sum_t \gamma^t r(s_t,a_t) | s_0 = s\right]$. 
The goal of reinforcement learning (RL) is to seek a policy that maximizes the expected discounted return $J(\pi)=\E_{\pi,P,\rho_0} \left[\sum_t \gamma^t r(s_t,a_t)\right]$.

\subsection{Policy Optimization}

Built upon the classic policy gradient based methods, e.g., vanilla policy gradient \cite{sutton1999policy}, the representative policy optimization methods at present are \textit{trust-region based} (e.g., TRPO) and  \textit{proximal-based} (e.g., PPO, ESPO) methods. 
We briefly cover their basics in this section. 

Trust Region Policy Optimization (TRPO)~\cite{schulman2015trust} derives a Total Variation (TV) divergence between target policy and behavior policy and presents guaranteed monotonic improvement with this TV divergence as regularization. Formally, TRPO uses Kullback–Leibler (KL) divergence and optimizes the following objectives:
\begin{align}\label{eq:obj_trpo}
\begin{split}
    \underset{\tilde{\pi}}{\textup{max}}
    \quad \E_{\pi,P} 
    \left[ 
    \frac{\tilde{\pi}(a|s)}{\pi(a|s)} A_{\pi}(s,a) \right], \quad \\
    \textup{s.t.} 
    \quad \E_{\pi,P}
    \left[
    \textup{KL}(\pi(a|s), \tilde{\pi}(a|s)) < \rho_{tr} \right],
\end{split}
\end{align}
where $A_{\pi}(s,a)$ denotes the advantage estimates, $\pi(a|s)$ denotes old policy used to sample data, $\tilde{\pi}(a|s)$ denotes new policy, and $\rho_{tr}$ is a threshold of KL divergence.

Proximal Policy Optimization (PPO)~\cite{schulman2017proximal} derives a lower bound of the original surrogate objective, which uses ratio clipping to restrict large policy update:
\begin{align}\label{eq:obj_ppo}
\begin{split}
    \underset{\tilde{\pi}}{\textup{max}}
    \quad \E_{\pi,P} 
    \bigg[ 
    \textup{min} \bigg( \frac{\tilde{\pi}(a|s)}{\pi(a|s)} A_{\pi}(s,a), \\
    \textup{clip}\bigg(\frac{\tilde{\pi}(a|s)}{\pi(a|s)}, 1-\epsilon, 1+\epsilon \bigg) A_{\pi}(s,a)
    \bigg) 
    \bigg].
\end{split}
\end{align}

Recently, \cite{sun2022you} revisits the ratio clipping technique proposed in PPO and find that the ratios in PPO are not bounded such that the resulting maximum TV divergence is non-trivially bounded. They thus present a new ratio-regularized improvement guarantee with the following theorem. 
\begin{theorem}\label{theorem_sun}
For any policies $\tilde{\pi}$ and $\pi$, the following bound holds:
\begin{align}\label{eq:bound_sun}
\begin{split}
    J(\tilde{\pi}) - J(\pi) 
    \ge 
    \frac{1}{1-\gamma} 
        \bigg\{ 
        \underbrace{\E_{\pi,P} \bigg[ \frac{\tilde{\pi}(a|s)}{\pi(a|s)} A_{\pi}(s,a)\bigg]}_{surrogate\;term} \\ - \underbrace{C\cdot \E_{\pi,P} \bigg| \frac{\tilde{\pi}(a|s)}{\pi(a|s)} - 1 \bigg|}_{correction\;term} \bigg\},
\end{split}
\end{align}
where $C=\frac{\xi \gamma}{1-\gamma}, \xi=\textup{max}_{s,a} |A_{\pi}(s,a)|$.
\end{theorem}
Theorem~\ref{theorem_sun} shows that policy improvement can be bounded by a surrogate term (i.e., policy advantage) and a correction term (i.e., ratio deviation).
This theorem shows that one can maximize the original surrogate objective without any ratio clipping. The key is to ensure that the ratio deviations are bounded when optimizing the surrogate objective. 
The authors thus propose an algorithm called early stopping policy optimization (ESPO) which early stops the multi-epoch policy optimization when the estimated ratio deviation exceeds a threshold $\delta_{es}$. Formally, the objective of ESPO is: 
\begin{align}\label{eq:obj_espo}
\begin{split}
    \underset{\tilde{\pi}}{\textup{max}}
    \quad \E_{\pi,P}
    \left[ 
    \frac{\tilde{\pi}(a|s)}{\pi(a|s)} A_{\pi}(s,a) \right], \quad \\
    \textup{s.t.} 
    \quad \E_{\pi,P}
    \left[
        \left| \frac{\tilde{\pi}(a|s)}{\pi(a|s)} - 1 \right|
    \right] < \delta_{es}.
\end{split}
\end{align}

\section{Methodology} \label{sec:methodology}

\subsection{Variance analysis of surrogate objective estimate}

One important and intriguing feature adopted by TRPO, PPO and ESPO 
is the use of importance sampling in the surrogate objective. 
Different from greed policy solver~\cite{kakade2002approximately}, 
which directly maximizes the policy advantage via on-policy samples, 
the use of importance sampling enables us to reuse samples generated by the previous policy, 
but also introduces high variance into the surrogate objective estimate.

We present an analysis of the variance in the surrogate objective estimate. 
Without importance sampling, the original surrogate objective is
$\max_{\tilde{\pi}} \mathbb{E}_{s\sim d_{\pi}}\left[\sum_{a}\tilde{\pi}(a|s)A_{\pi}(s, a) \right]$~\cite{kakade2002approximately}. 
Without loss of generality,
we consider in the subsequent analysis only the action-summation part $\sum_{a}\tilde{\pi}(a|s)A_{\pi}(s, a)$. 
Specifically, the empirical estimate involves on-policy samples induced by $\tilde{\pi}$. 
In discrete action control, 
the maximal of this empirical estimate is achieved when $\tilde{\pi}(a|s) = \arg\max A_{\pi}(s, a)$~\cite{kakade2002approximately,pirotta2013safe}. 
However, in continuous action control for which the summation is integrated over the whole action space, 
computing the maximal of this empirical estimate could be challenging especially when 
the advantage estimates are only available at a limited set of sample points. 
As an alternative, we consider the \textit{importance sampling estimate}:
\begin{equation*}
\hat{\mu}_{\pi} = \frac{1}{n} \sum_{i=1}^{n} \frac{\tilde{\pi}(a|s)}{\pi(a|s)}A_{\pi}(s, a), \quad a\sim\pi(\cdot|s). 
\end{equation*}
Importantly, this importance sampling form provides an unbiased estimate of the \textit{true} surrogate objective $\mathbb{E}_{a\sim\tilde{\pi}}\left[ A_{\pi}(s, a) \right]$. 
The following theorem provides a bound for the estimation variance. 
\begin{theorem}\label{theo:variance}
Let $\mu\triangleq\mathbb{E}_{a\sim\pi}\big[\frac{\tilde{\pi}(a|s)}{\pi(a|s)}A_{\pi}(s, a)\big]$, 
the variance of importance sampling estimate for the surrogate objective is bounded as follows:
\begin{equation*}
    \sigma_{\pi} \leq \xi^2 \cdot \mathbb{E}_{a\sim\pi} \left[ \Big(\frac{\tilde{\pi}(a|s)}{\pi(a|s)}\Big)^2 \right] - \left[\mathbb{E}_{a\sim\pi}\big[\frac{\tilde{\pi}(a|s)}{\pi(a|s)}A_{\pi}(s, a)\big]\right]^2. 
\end{equation*}
\end{theorem}
The proof is straightforward according to the definition of variance: 
$\sigma_{\pi} \triangleq \mathbb{E}_{a\sim\pi}\left[ \Big(\frac{\tilde{\pi}(a|s)}{\pi(a|s)}A_{\pi}(s, a)\Big)^2 \right] - \left[\mathbb{E}_{a\sim\pi}\big[\frac{\tilde{\pi}(a|s)}{\pi(a|s)}A_{\pi}(s, a)\big]\right]^2$, 
and $\xi \triangleq \max_{s, a}A_{\pi}(s, a)$. 


This theorem indicates that, 
when importance sampling is introduced as a substitute for the surrogate objective,
the variance of estimating the \textit{true} surrogate objective grows quadratically w.r.t the policy ratios. 
Namely, importance sampling scheme can perform quite poorly 
when $\mathbb{E}_{a\sim\pi}\left[ \Big(\frac{\tilde{\pi}(a|s)}{\pi(a|s)}\Big)^2 \right] = +\infty$\cite{robert1999monte}. 
This could happen in policy optimization. 
We show in the experiment section that the policy ratios in PPO and ESPO can easily grow to a large value, 
which could invalidate the surrogate objective estimate based on the above theorem.
We will provide diagnostic analysis of the estimate variance in the experiment and
consider checking the ratio distribution due to the fact 
that $\mathbb{E}_{a\sim\pi}\left[ \frac{\tilde{\pi}(a|s)}{\pi(a|s)} \right] = 1$:
If the sample mean of the weights is far from $1$ then that is a sign that importance sampling could poorly perform. 
That is, $\bar{\omega} = \frac{1}{n}\sum_{i=1}^{n} \omega_i$ is another diagnostic~\cite{mcbook}.


\subsection{Sample dropout in surrogate objective maximization}\label{sec:theory}
The upper bound of variance in Theorem~\ref{theo:variance} is determined by two terms: 
the first term involves the ratios, 
and the second term corresponds to the surrogate objective to be optimized. 
As the surrogate objective is being maximized, 
the ratios will be increased, 
which in turn would also increase the variance. 
In PPO and ESPO, the surrogate objective is often optimized via multiple mini-epochs. 
The ratios, if without any clipping or explicit bound, will usually become very large 
(see our empirical results of Fig.~\ref{fig:ratio_range} in the experiment section). 
The variance could thus become very high and using the importance sampling estimate might be no longer effective. 
In the ideal case, the surrogate objective should be optimized in a way that the variance is also kept to be small.

By Theorem~\ref{theo:variance}, one way to achieve a low variance 
when optimizing the surrogate objective is to keep ratios small. 
We thus propose a strategy by disposing samples whose ratio deviation exceeds the threshold $\delta$. 
We call this strategy \textit{sample dropout} and define this operation as \textit{ratio-based dropout indicator}:
\begin{equation} \label{dropout_indicator}
    \cond.
\end{equation}
This indicator returns 1 when the ratio deviation is within the threshold $\delta$ and returns 0 otherwise, which is a sign indicating which samples are used for training.


Besides using ratio deviation as a criteria for sample dropout, an alternative criteria is using the KL divergence. 
Therefore, we define a new operation called \textit{KL-based dropout indicator}:
\begin{equation} \label{kl_based_dropout_indicator}
    \klcond, 
\end{equation}
where $\rho$ is a threshold of KL divergence. This indicator returns 1 when the KL divergence is within the threshold $\rho$ and returns 0 otherwise. 
In principle, the ratio-based dropout indicator can be used in both trust-region based (e.g., TRPO) and proximity-based (e.g., PPO, ESPO) policy optimization methods. Practically, we use the KL-based one to align with the constraints in TRPO and empirically find that it works slightly better than the ratio-based one for TRPO. 
Therefore, for proximity-based policy optimization methods (e.g., PPO, ESPO), we use the ratio-based dropout indicator while for trust-region based methods (e.g., TRPO), we use the KL-based one. 

\begin{table*}[t]
    \centering
    \begin{tabular}{ccc|cc|cc} 
    \toprule
    Tasks &   TRPO score & SD-TRPO score & PPO score & SD-PPO score & ESPO score & SD-ESPO score \\
    \midrule
    Ant & 2848 & \textbf{4021} & 5304 & \textbf{5356} & 4094 & \textbf{5124} \\
    Humanoid & 1835 & \textbf{2112} & 1277 & \textbf{1478} & 4176 & \textbf{4671} \\
    HumanoidStandup & 102174 & \textbf{106480} & 150934 & \textbf{157595} & 159110 & \textbf{177585} \\
    Inv.Dou.Pen. & 7635 & \textbf{7992} & 8950 & \textbf{8956} & 6815 & \textbf{8071} \\
    Walker2d & 3503 & \textbf{4045} & 4322 & \textbf{5145} & 3455 & \textbf{3501} \\
    HalfCheetah & 2121 & \textbf{3044} & 2166 & \textbf{2869} & 2771 & \textbf{3177} \\
    \midrule
    Dog.fetch & 33 & \textbf{48} & 23 & \textbf{96} & 75 & \textbf{77} \\
    Dog.run & 92 & \textbf{105} & 142 & \textbf{163} & 193 & \textbf{196} \\
    Dog.stand & 375 & \textbf{425} & 276 & \textbf{698} & 770 & \textbf{803} \\
    Dog.trot & 93 & \textbf{120} & 270 & \textbf{345} & 264 & \textbf{486} \\
    Dog.walk & 206 & \textbf{215} & 371 & \textbf{544} & 421 & \textbf{593} \\
    Walker.walk & 643 & \textbf{655} & 630 & \textbf{742} & 539 & \textbf{587} \\
    \bottomrule
    \end{tabular}
    \caption{Performance results for MuJoCo and DMControl. Each number is averaged over 10 random seeds. The scores are undiscounted average returns. The results show that sample dropout greatly boosts the performance of policy optimization methods across a wide range of continuous control tasks.} \label{table:all}
\end{table*}

\subsection{Practical Algorithm}

Our idea is simple to implement and can be easily combined with existing policy optimization algorithms. 
Backed by the analysis above, in this section, we develop algorithmic instantiations of sample dropout. 
Specifically, we instantiate our idea in TRPO, PPO, and the recently proposed ESPO, resulting in a new family of algorithms, i.e., SD-TRPO, SD-PPO, and SD-ESPO, respectively. 
Our algorithmic framework is shown in Algorithm~\ref{algo:sdpo}. In each iteration, we use the policy $\pi$ to roll out in the environments and collect a batch of data (line\ref{algo:init_buffer}-\ref{algo:append}), and then we compute advantage estimate (line~\ref{algo:compute_advantage}) using GAE~\cite{schulman2015high}. Hence, we perform updates for policy and value networks according to the instantiated algorithm (SD-TRPO, SD-PPO, SD-ESPO) (line\ref{algo:pi_start}-\ref{algo:pi_end}).
We elaborate these instantiated methods below.

\begin{algorithm}[t]
\caption{Policy Optimization with Sample Dropout}
\label{algo:sdpo}
\begin{algorithmic}[1]
    \State Randomly initialize policy $\pi$, value network $V_{\phi}$
    \For {iterations $i=1,2,...$}
        \State Initialize sample buffer $B=\emptyset$ \label{algo:init_buffer}
        \For {step=$1,2,...,N$} \Comment{Sampling data}
            \State Receive current state $s$ and take action $a$ according to policy $\pi$
            \State Receive reward $r$ and append tuple $(s,a,r)$ into buffer $B$  \label{algo:append}
        \EndFor
        \State Compute advantage estimate $\hat{A}_{\pi}$ using GAE\label{algo:compute_advantage}
        \For {epoch=$1,2,...,K$} \Comment{Policy and value updates} \label{algo:pi_start}
            \State Sample $L$ samples from the buffer
            \State Compute the dropout indicator as defined in Eq. \ref{dropout_indicator} and Eq. \ref{kl_based_dropout_indicator}
            \State Optimize policy $\tilde{\pi}_{\theta}$ and value $V_{\phi}$ with instantiated algorithms \label{algo:pi_end} 
        \EndFor
        \State $\pi \leftarrow \tilde{\pi}_{\theta}$
    \EndFor
\end{algorithmic}
\end{algorithm}

\paragraph{Sample Dropout TRPO (SD-TRPO)} We combine the idea of sample dropout with the TRPO objective (Eq.~\ref{eq:obj_trpo}) by simply multiplying the TRPO's surrogate objective by the ratio-based dropout indicator, and form an algorithm called sample dropout TRPO (SD-TRPO). 
Formally, the objective of SD-TRPO is as follows:
\begin{equation}\label{eq:obj_cd_trpo}
\begin{split}
\footnotesize
    \underset{\tilde{\pi}}{\textup{max}}
    \quad \E_{\pi,P}
    \left[
            \textup{sg} \left( \klcond \right)
    \frac{\tilde{\pi}(a|s)}{\pi(a|s)} A_{\pi}(s,a) \right], \\
    \textup{s.t.} 
    \quad \E_{\pi,P}
    \bigg[
    \textup{KL}(\pi(a|s), \tilde{\pi}(a|s)) < \rho_{tr} \bigg].
\end{split}
\end{equation}
Here $\textup{sg}$ denotes ``stop gradient'' to avoid gradient propagating through the dropout indicator.

\paragraph{Sample Dropout PPO (SD-PPO)} PPO derives a lower bound of the original surrogate objective to get a pessimistic estimate of the policy performance (as shown in Eq.~\ref{eq:obj_ppo}). We directly multiply this lower bound by our ratio-based dropout indicator and form an objective for SD-PPO:
\begin{equation}\label{eq:obj_cd_ppo}
\begin{split}
\footnotesize
    \underset{\tilde{\pi}}{\textup{max}}
    \quad \E_{\pi,P}
    \bigg[ 
        \textup{sg} \left( \cond \right)
    \textup{min} \bigg( \frac{\tilde{\pi}(a|s)}{\pi(a|s)} A_{\pi}(s,a), \\
    \textup{clip}\bigg(\frac{\tilde{\pi}(a|s)}{\pi(a|s)}, 1-\epsilon, 1+\epsilon \bigg) A_{\pi}(s,a)
    \bigg) \bigg].
\end{split}
\end{equation}
Similar to PPO, SD-PPO performs multi-epoch optimization on the same batch of data. In our implementation, for each epoch, we sample a mini-batch of data, compute the ratio-based dropout indicator on the mini-batch data, and then optimize the policy with the objective in Eq.~\ref{eq:obj_cd_ppo}.

\paragraph{Sample Dropout ESPO (SD-ESPO)} ESPO optimizes the original surrogate objective and early stops the multi-epoch policy optimization according to the value of ratio deviation (as shown in Eq.~\ref{eq:obj_espo}). We apply the ratio-based dropout indicator to the objective of ESPO and form sample dropout ESPO (SD-ESPO) with the following objective:
\begin{equation}\label{eq:obj_cd_espo}
\begin{split}
\footnotesize
\underset{\tilde{\pi}}{\textup{max}}
    \quad \E_{\pi,P}
    \left[ 
        \textup{sg} \left( \cond \right)
    \frac{\tilde{\pi}(a|s)}{\pi(a|s)} A_{\pi}(s,a) \right], \\
    \textup{s.t.} 
    \quad \E_{\pi,P}
        \left| \frac{\tilde{\pi}(a|s)}{\pi(a|s)} - 1 \right|
    < \delta_{es}.
\end{split}
\end{equation}

The objectives described above are used for policy updates. For value function updates, we also apply the idea of sample dropout --- this helps us to avoid fitting value networks to those samples with large ratio deviation. Formally, the value function is updated as follows:
\begin{equation}\label{eq:obj_cd_espo_value}
\begin{split}
\footnotesize
\underset{\phi}{\textup{min}}
    \quad \E_{\pi,P}
    \left[ 
        \textup{sg} \left(
        \mathbbm{1} \left\{ 
            \left| \frac{\tilde{\pi}(a|s)}{\pi(a|s)} - 1 \right| < \delta
        \right\} \right)
        \left( V_{\phi}(s) - R \right)^2 \right],
\end{split}
\end{equation}
where $R$ is the discounted sum of rewards, $V_{\phi}(s)$ denotes value networks parameterized by $\phi$.

\section{Experiments}

In our experiments, we aim to answer: 
(1) Can sample dropout improve the performance of existing policy optimization algorithms? 
(2) Does sample dropout help reduce variance of surrogate objective estimates? 
(3) How do different degrees of sample dropout affect the performance?

\begin{figure*}[t]
	\centering
	\includegraphics[width=0.25\textwidth]{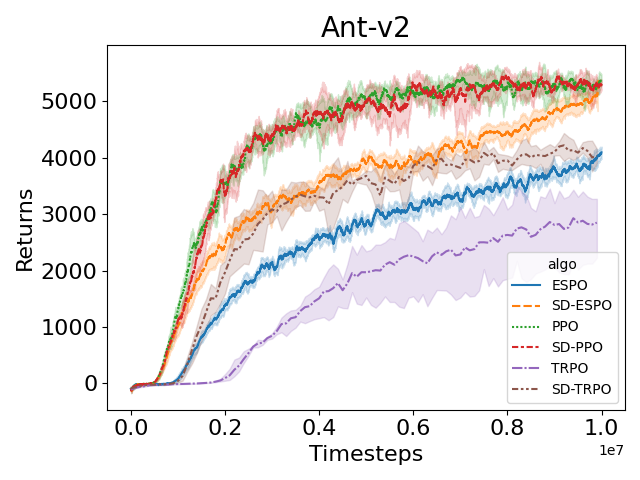}
	\includegraphics[width=0.25\textwidth]{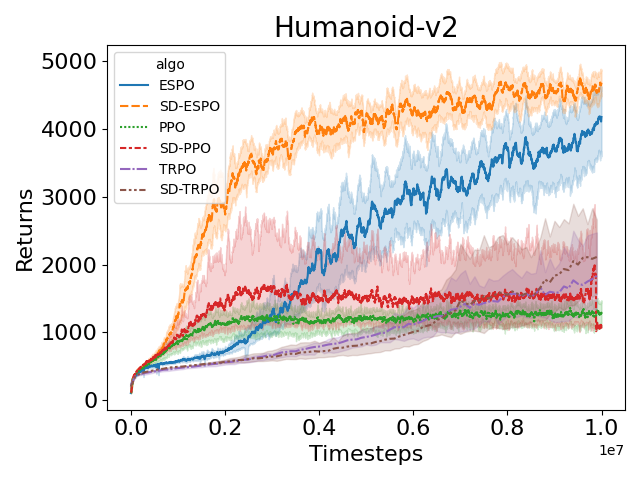}
	\includegraphics[width=0.25\textwidth]{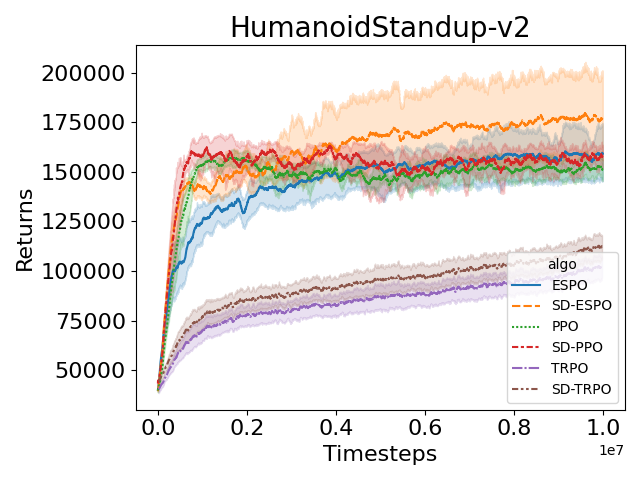}
	\includegraphics[width=0.25\textwidth]{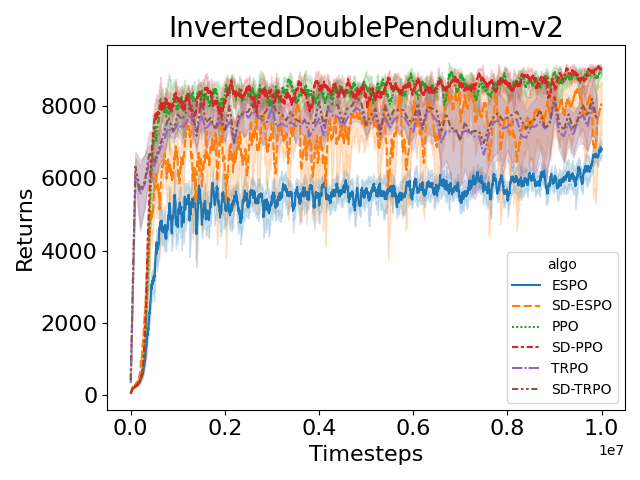}
	\includegraphics[width=0.25\textwidth]{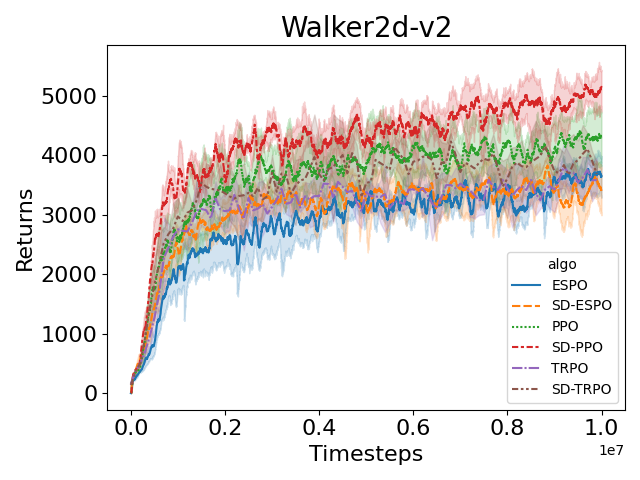}
	\includegraphics[width=0.25\textwidth]{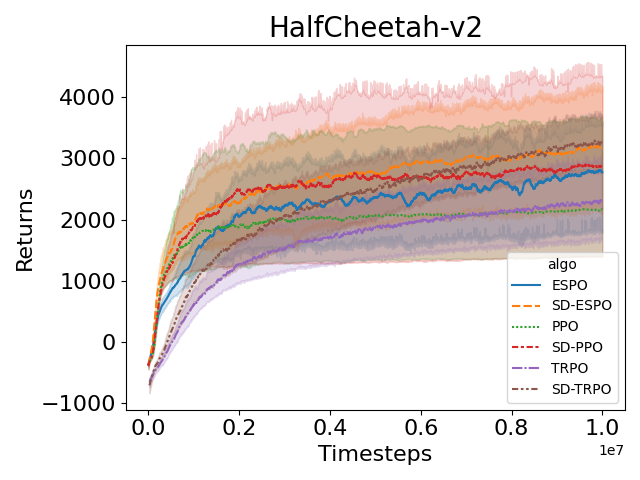}
 	\caption{Training curves (plotted using 10 random seeds) of MuJoCo tasks. The results show that sample dropout (SD) consistently boosts the performance and sample efficiency of TRPO, PPO, and ESPO across a wide range of continuous control tasks. }
	\label{fig:mujoco}
\end{figure*}

\subsection{Setup}
\begin{table}[ht]
\centering
    \begin{tabular}{ll}
		\toprule
		Environment & $\mathcal{S} \times \mathcal{A}$ \\
		\midrule
        Ant & $ \mathbb{R}^{111} \times \mathbb{R}^{8} $ \\
        Humanoid & $ \mathbb{R}^{376} \times \mathbb{R}^{17} $ \\
        HalfCheetah & $ \mathbb{R}^{17} \times \mathbb{R}^{6} $ \\
        Inv.Dou.Pen. & $ \mathbb{R}^{11} \times \mathbb{R}^{1} $ \\
        HumanoidSta. & $ \mathbb{R}^{376} \times \mathbb{R}^{17} $ \\
        Walker2d & $ \mathbb{R}^{17} \times \mathbb{R}^{6} $ \\
		\midrule
		dm\_dog.fetch & $ \mathbb{R}^{223} \times \mathbb{R}^{38} $ \\
		dm\_dog.run & $ \mathbb{R}^{223} \times \mathbb{R}^{38} $ \\
		dm\_dog.stand & $ \mathbb{R}^{223} \times \mathbb{R}^{38} $ \\
		dm\_dog.trot & $ \mathbb{R}^{223} \times \mathbb{R}^{38} $ \\
		dm\_dog.walk & $ \mathbb{R}^{223} \times \mathbb{R}^{38} $ \\
		\midrule
		Atari domains & $ \mathbb{R}^{7056} \times \mathbb{R}^{18} $ \\
		\bottomrule
	\end{tabular}
	\caption{Control complexity of different domains.}
	\label{app:domain_complexity}
	\vspace{-0.2in}
\end{table}
We conduct experiments using environments that have been widely used in prior works \cite{schulman2017proximal,sun2022you}. Specifically, we consider continuous control MuJoCo benchmarks~\cite{todorov2012mujoco}, higher-dimensional tasks in the DeepMind Control Suite (DMControl)~\cite{tassa2018deepmind}, as well as Atari video  games~\cite{bellemare2013arcade}. 
We provide an overview of the control complexity in terms of state-action dimensions in Table~\ref{app:domain_complexity}. 
For continuous control tasks, we use Gaussian distribution for the policy whose mean and covariance are parameterized by neural networks. 
For discrete Atari domains, we use the same policy network architecture as used in \cite{mnih2015human}. 
We evaluate SD-TRPO, SD-PPO, and SD-ESPO by comparing them with baseline counterparts, i.e., TRPO, PPO, and ESPO. 
Following a common practice in \cite{schulman2017proximal,sun2022you}, we normalize observations, rewards and advantages for all algorithms. 
Given the concerns in reproducibility~\cite{henderson2018deep}, we run all experiments across a large number of random seeds (10 random seeds) and report the mean and standard deviation of training curves. 
Our code is built upon the OpenAI baselines~\cite{baselines}.

We provide the implementation details and all hyperparameter settings. Table~\ref{app:hyper:trpo} presents hyperparameters of TRPO and SD-TRPO. Note that the main difference between TRPO and SD-TRPO is the additional hyperparameter threshold of KL divergence (i.e., $\rho$) used in the KL-based dropout indicator. For TRPO and SD-TRPO, the epoch number $K$ in Algorithm~\ref{algo:sdpo} is set to 1 and the sampling number $L$ is set to $N$.
Table~\ref{app:hyper:ppo} shows hyperparameters of PPO and SD-PPO, where SD-PPO involves an additional hyperparameter $\delta$ representing the threshold of ratio deviation. 
Table~\ref{app:hyper:espo} shows hyperparameters of ESPO and SD-ESPO. SD-ESPO also involves an extra hyperparameter $\delta$ compared to ESPO. 


\begin{figure*}[t!]
	\centering
	\includegraphics[width=0.25\textwidth]{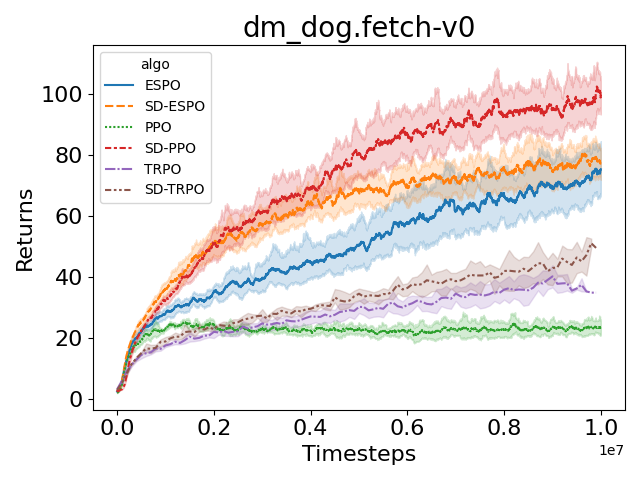}
	\includegraphics[width=0.25\textwidth]{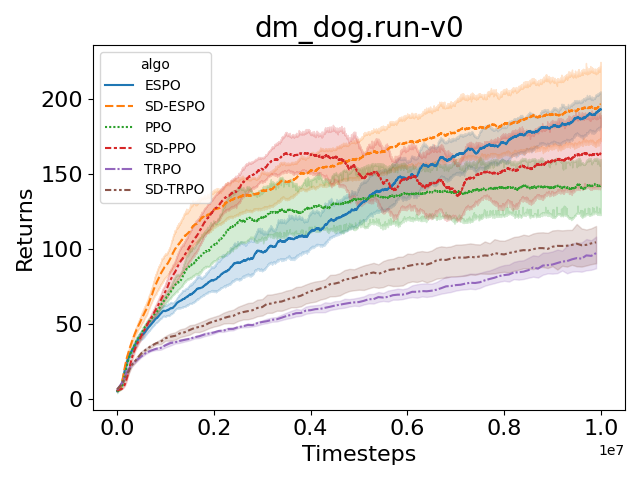}
	\includegraphics[width=0.25\textwidth]{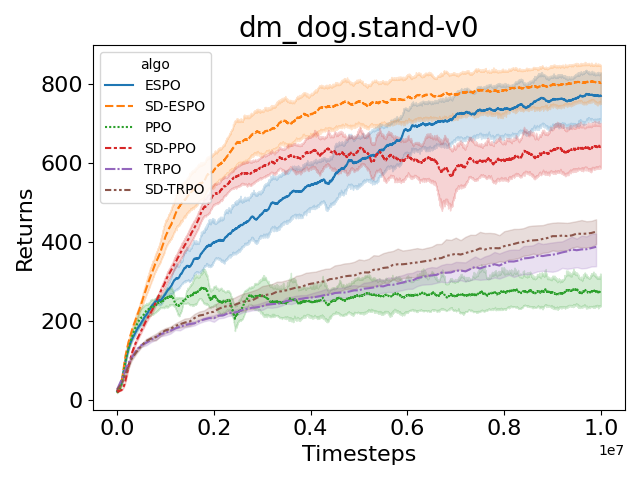}
	\includegraphics[width=0.25\textwidth]{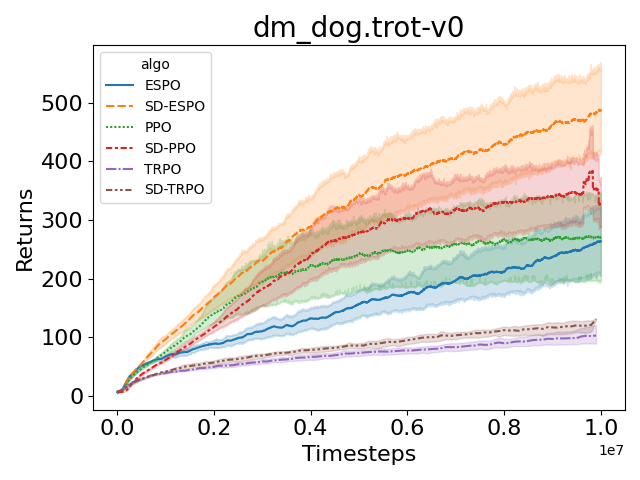}
	\includegraphics[width=0.25\textwidth]{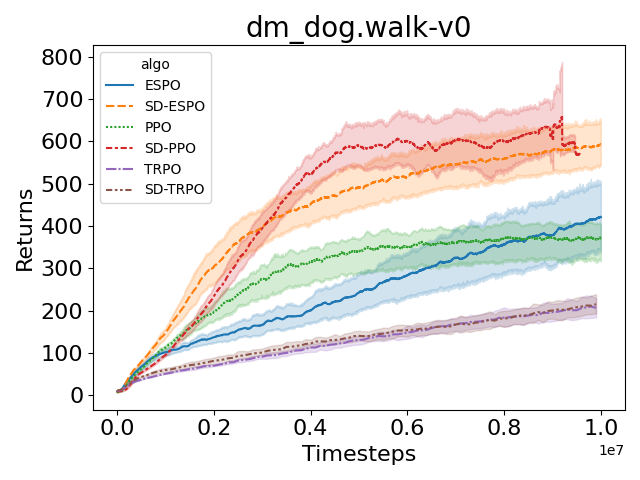}
	\includegraphics[width=0.25\textwidth]{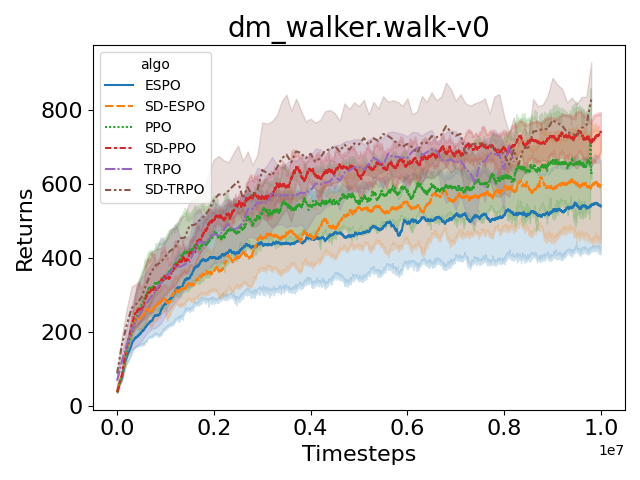}
	\caption{Training curves (plotted using 10 random seeds) in DMControl tasks. The results show that sample dropout (SD) consistently boosts the performance and sample efficiency of TRPO, PPO and ESPO across a wide range of continuous control tasks. }
	\label{fig:dmcontrol}
\end{figure*}

\begin{table}[h]
    \centering
    \tiny
    \begin{tabular}{ccccccc} 
    \toprule
    & \multicolumn{2}{c}{\textbf{MuJoCo}} & \multicolumn{2}{c}{\textbf{DMControl}} & \multicolumn{2}{c}{\textbf{Atari}} \\
    \toprule
    Hyperparameter & TRPO & SD-TRPO & TRPO & SD-TRPO & TRPO & SD-TRPO \\
    \midrule
    Batch size ($N$) & 4000 & 4000 & 4000 & 4000 & 4000 & 4000 \\
    $\lambda$ & 0.97  & 0.97 & 0.97 & 0.97 & 0.97 & 0.97\\
    $\gamma$ & 0.99  & 0.99 & 0.99 & 0.99 & 0.99 & 0.99\\
    backtrack coefficient & 0.8 & 0.8 & 0.8 & 0.8 & 0.8 & 0.8 \\
    backtrack iterations & 10 & 10 & 10 & 10 & 10 & 10 \\
    cg iterations & 10 & 10 & 10 & 10 & 10 & 10 \\
    damping coefficient & 0.1 & 0.1 & 0.1 & 0.1 & 0.1 & 0.1 \\
    value update iteration & 80 & 80 & 80 & 80 & 80 & 80 \\
    value update learning rate & 1e-3 & 1e-3 & 1e-3 & 1e-3 & 1e-3 & 1e-3 \\
    $\rho_{tr}$ & 0.001 & 0.001 & 0.001 & 0.001 & 0.001 & 0.001 \\
    $\rho$ & - & 0.001 & - & 0.001 & - & 0.001 \\
    \bottomrule
    \end{tabular}
\caption{Hyperparameters for TRPO and SD-TRPO.} \label{app:hyper:trpo}
\end{table}

\begin{table}[h]
    \centering
    \tiny
    \begin{tabular}{ccccccc} 
    \toprule
    & \multicolumn{2}{c}{\textbf{MuJoCo}} & \multicolumn{2}{c}{\textbf{DMControl}} & \multicolumn{2}{c}{\textbf{Atari}} \\
    \toprule
    Hyperparameter & PPO & SD-PPO & PPO & SD-PPO & PPO & SD-PPO \\
    \midrule
    Batch size ($N$) & 2048 & 2048 & 2048 & 2048 & 2048 & 2048 \\
    Mini-batch size ($L$) & 512 & 512 & 512 & 512 & 512 & 512 \\
    $\lambda$ & 0.95  & 0.95 & 0.95 & 0.95 & 0.95 & 0.95\\
    $\gamma$ & 0.99  & 0.99 & 0.99 & 0.99 & 0.99 & 0.99\\
    Optimization epochs ($K$) & 10  & 10 & 10 & 10 & 10 & 10\\
    Entropy coefficient & 0.0  & 0.0  & 0.0  & 0.0  & 0.0  & 0.0 \\
    Initial learning rate & 0.0003 & 0.0003 & 0.0003 & 0.0003 & 0.0003 & 0.0003 \\
    Learning rate decay & linear & linear & linear & linear & linear & linear \\
    $\delta$ & - & 1.0 & - & 0.5 & - & 0.5 \\
    \bottomrule
    \end{tabular}
\caption{Hyperparameters for PPO and SD-PPO.} \label{app:hyper:ppo}
\end{table}

\begin{table}[h]
    \centering
    \tiny
    \begin{tabular}{ccccccc} 
    \toprule
    & \multicolumn{2}{c}{\textbf{MuJoCo}} & \multicolumn{2}{c}{\textbf{DMControl}} & \multicolumn{2}{c}{\textbf{Atari}} \\
    \toprule
    Hyperparameter & ESPO & SD-ESPO & ESPO & SD-ESPO & ESPO & SD-ESPO \\
    \midrule
    Batch size ($N$) & 2048 & 2048 & 2048 & 2048 & 2048 & 2048 \\
    Mini-batch size ($L$) & 64 & 64 & 64 & 64 & 64 & 64 \\
    $\lambda$ & 0.95  & 0.95 & 0.95 & 0.95 & 0.95 & 0.95\\
    $\gamma$ & 0.99  & 0.99 & 0.99 & 0.99 & 0.99 & 0.99\\
    Epochs ($K$) & 10  & 10 & 10 & 10 & 10 & 10\\
    Entropy coefficient & 0.0  & 0.0  & 0.0  & 0.0  & 0.0  & 0.0 \\
    Init learning rate & 0.0003 & 0.0003 & 0.0003 & 0.0003 & 0.0003 & 0.0003 \\
    Learning rate decay & linear & linear & linear & linear & linear & linear \\
    $\delta_{es}$ & 0.25 & 0.25 & 0.25 & 0.25 & 0.25 & 0.25 \\
    $\delta$ & - & 0.25 & - & 0.25 & - & 0.25 \\
    \bottomrule
    \end{tabular}
\caption{Hyperparameters for ESPO and SD-ESPO.} \label{app:hyper:espo}
\end{table}

\subsection{Main results}
\paragraph{MuJoCo benchmarks}
Table~\ref{table:all} shows the performance on MuJoCo tasks. 
We see that sample dropout can consistently boost the performance of existing policy optimization algorithms. Figure~\ref{fig:mujoco} compares the training curves between different algorithms on MuJoCo. We find that the sample dropout technique consistently improves TRPO, PPO and ESPO across a wide range of MuJoCo benchmarks. 
In particular, sample dropout achieves significant performance boost in the high-dimensional tasks (e.g., Humanoid, Ant). 
Compared with TRPO, we find that SD-TRPO achieves clear performance boost in three MuJoCo tasks (Ant, HalfCheetah, HumanoidStandup). 
Compared with PPO, we find that SD-PPO outperforms PPO in three tasks (Humanoid, HalfCheetah, and Walker2d).
Comared with ESPO, SD-ESPO wins with a clear margin in five tasks (Ant, Humanoid, InvertedDoublePendulumm, HalfCheetah, and HumanoidStandup).
In general, algorithms with sample dropout (SD-) have advantage over those without sample dropout in terms of both sample efficiency and the final performance. 

\paragraph{DMControl benchmarks} We further compare algorithms on more complex control domains: the Dog and Walker domains in the DMControl, including tasks of fetch, run, stand, trot and walk, which have higher-dimensional action space than the tasks in MuJoCo (as shown in Table~\ref{app:domain_complexity}) and poses more challenges to the algorithms.
As shown in Table~\ref{table:all}, the sample dropout technique also works quite well in DMControl. Figure~\ref{fig:dmcontrol} shows the training curves comparison of all algorithms. 
We observe that: 1) SD-PPO outperforms PPO by a large margin in three tasks (dog.fetch, dog.stand and dog.walk); 2) SD-ESPO beats ESPO in all tested tasks and demonstrates clear advantage over ESPO in four tasks (dog.trot, dof.walk, dog.fetch and dog.stand); 3) SD-TRPO achieves slight performance boost over TRPO. We conjecture that this is because TRPO poses hard strict constraints to the policy update and thus struggle to learn efficiently in DMControl.

\begin{figure*}[t!]
	\centering
	\includegraphics[width=0.25\textwidth]{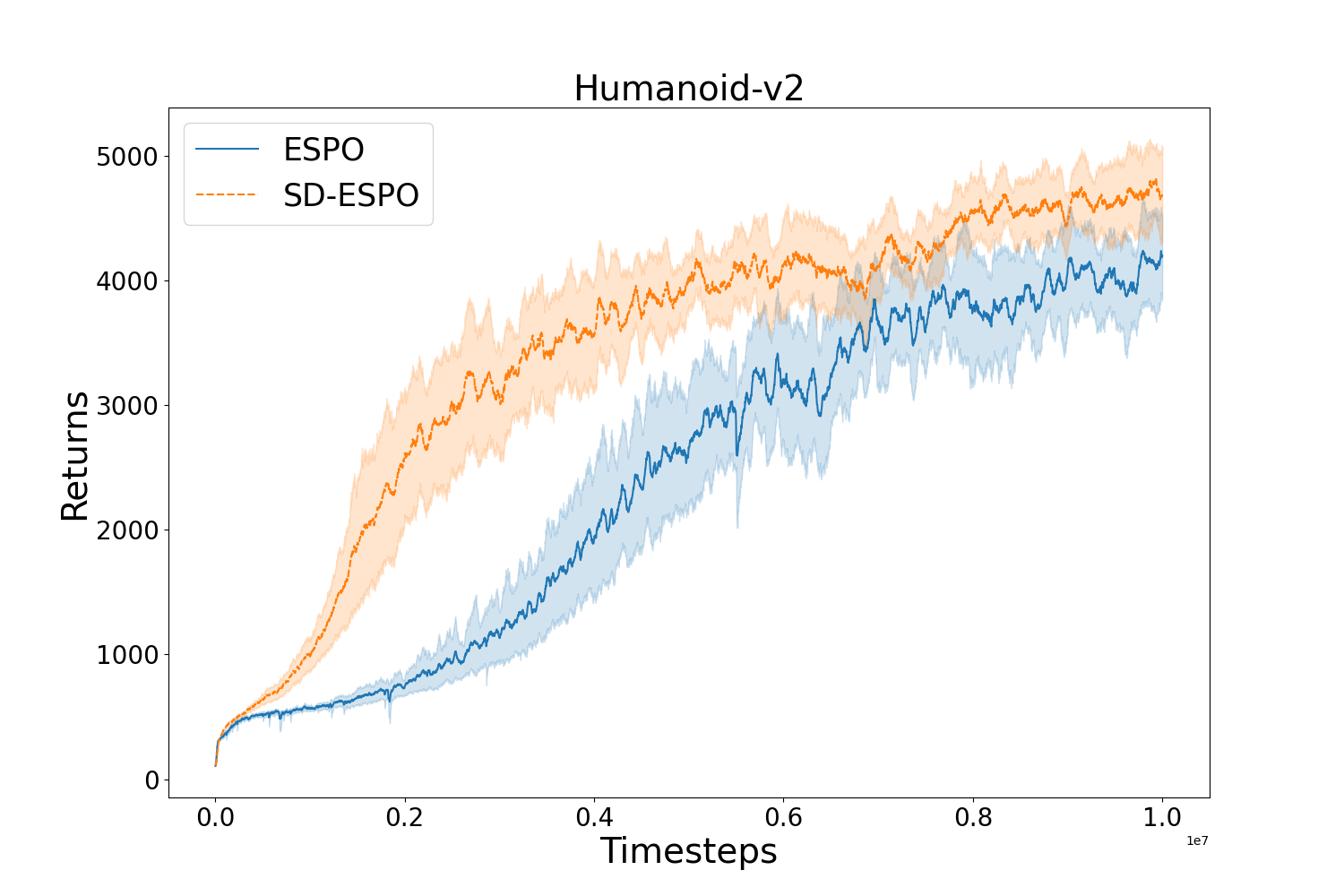}
	\includegraphics[width=0.25\textwidth]{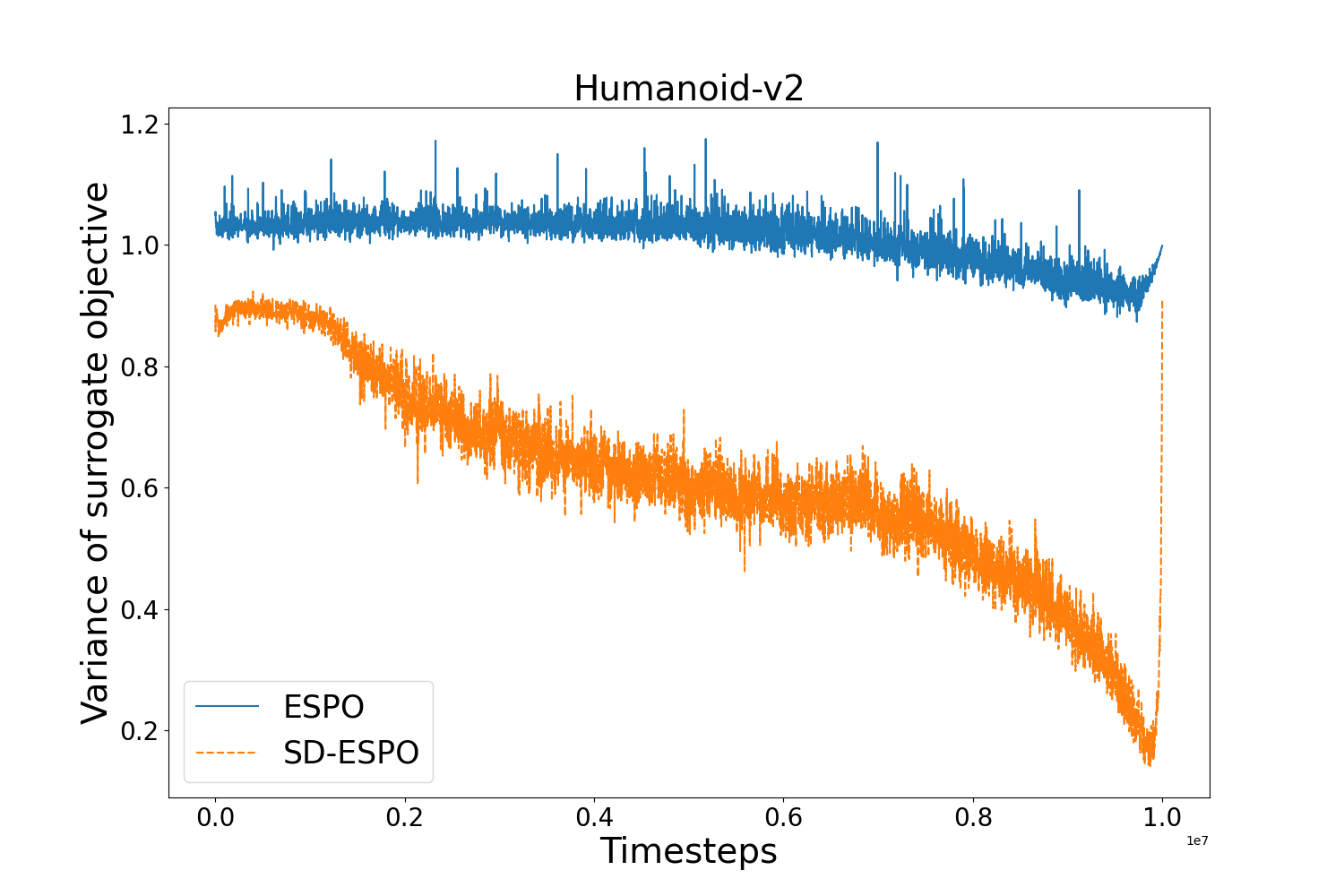}
	\includegraphics[width=0.25\textwidth]{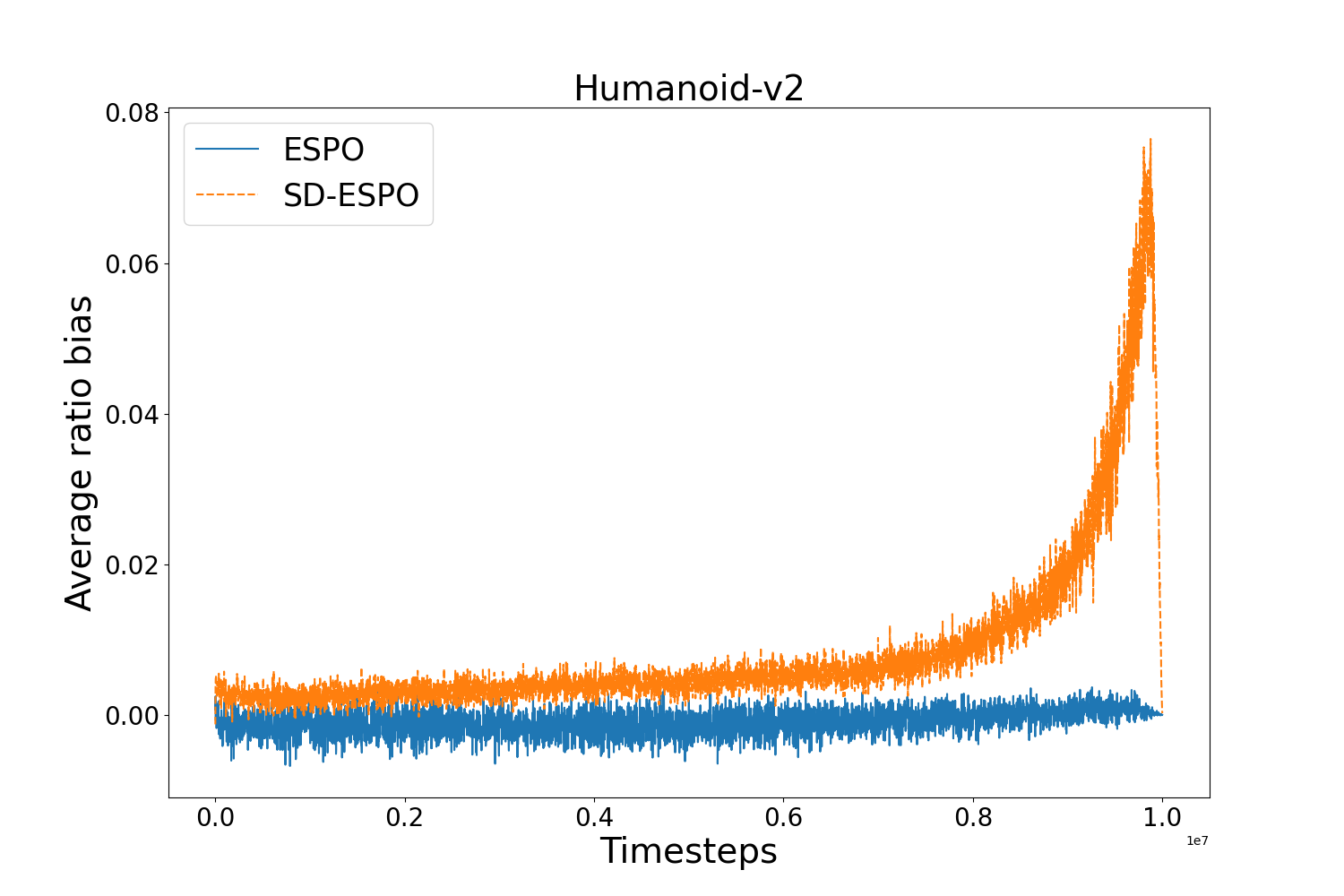}

	\includegraphics[width=0.25\textwidth]{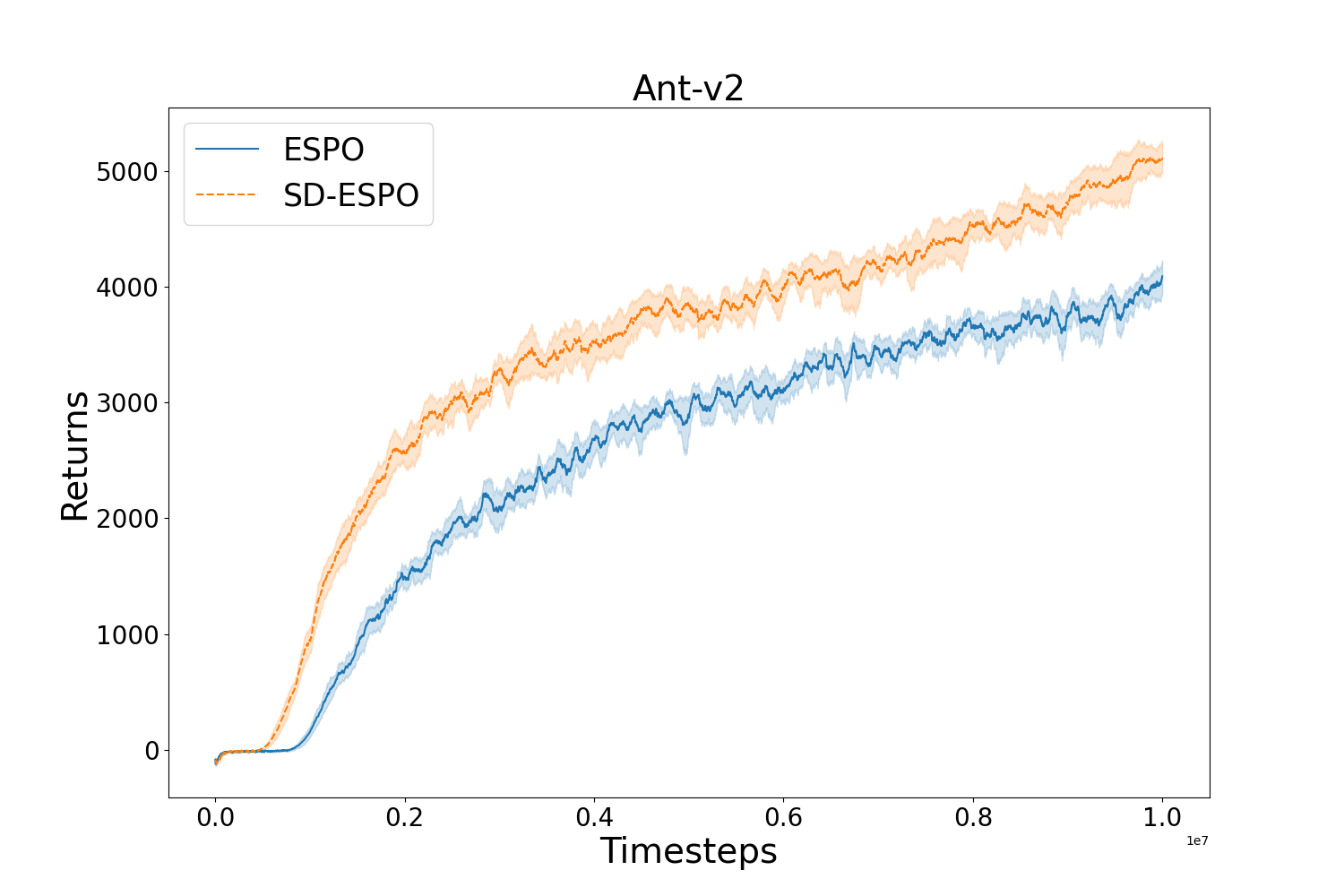}
 	\includegraphics[width=0.25\textwidth]{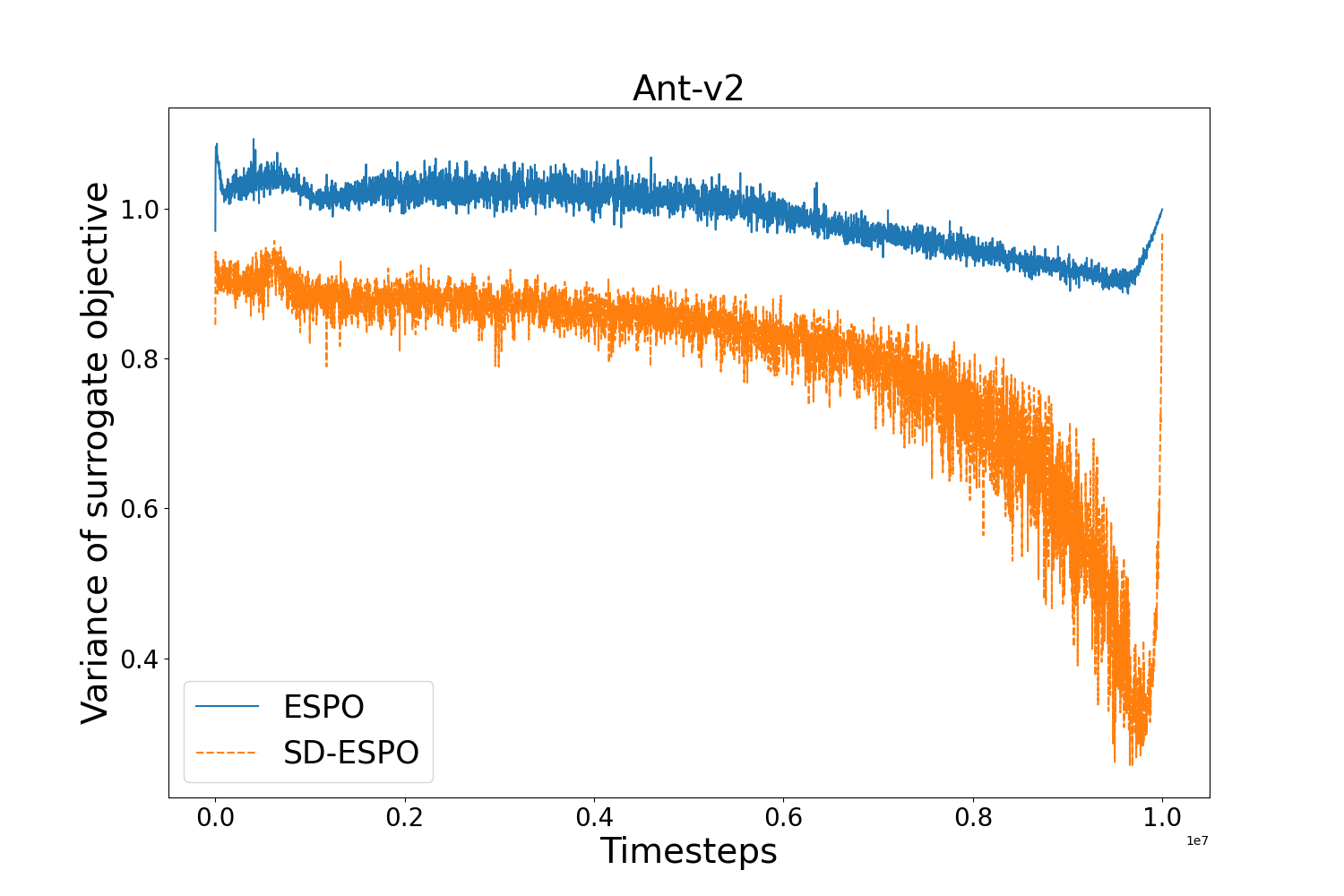}
 	\includegraphics[width=0.25\textwidth]{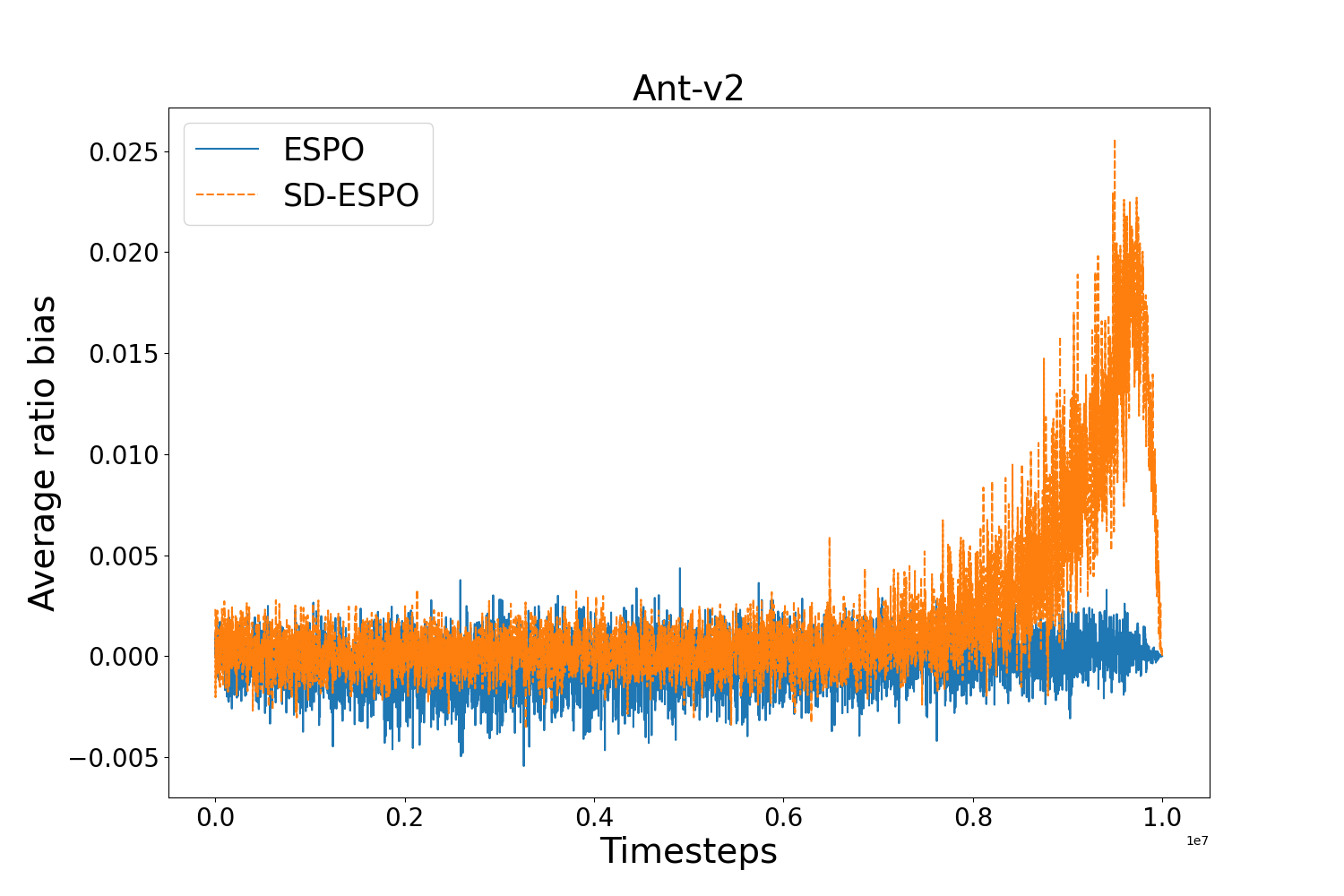}
 
 
    \caption{Comparison between ESPO and SD-ESPO. Left column: agent performance on achieved returns. Middle column: variance of the surrogate objective estimate. Right column: average ratio bias which reflects the ratio deviation from 1.0. These results establish that the sample dropout technique reduces the variance of surrogate objective and significantly improves sample efficiency. While sample dropout introduces slight bias, we find that the introduced bias is tolerable. Overall, this technique benefits the training process.}
	\label{fig:reduce_variance}
\end{figure*}

\begin{figure*}[th!]
\centering
\includegraphics[width=0.17\textwidth]{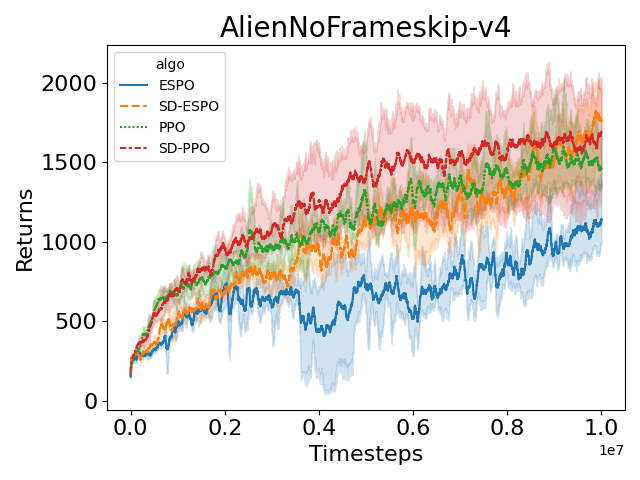}
\includegraphics[width=0.17\textwidth]{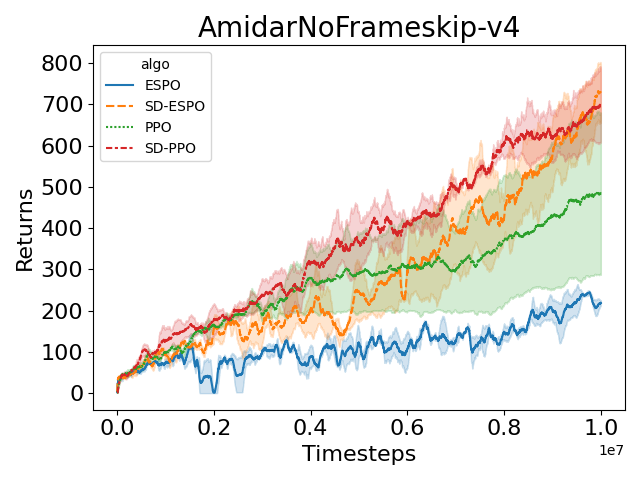}
\includegraphics[width=0.17\textwidth]{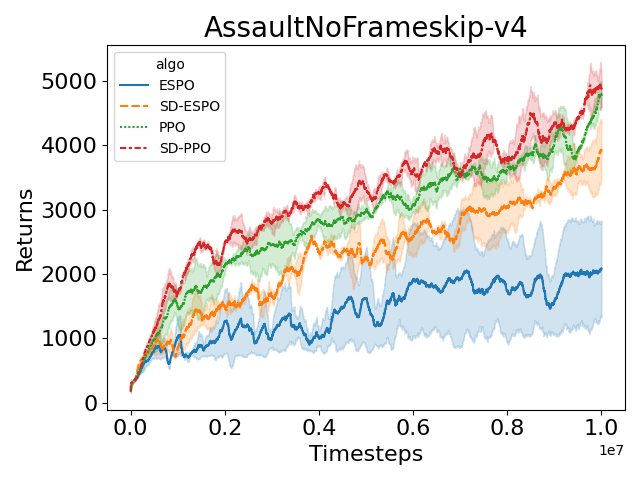}
\includegraphics[width=0.17\textwidth]{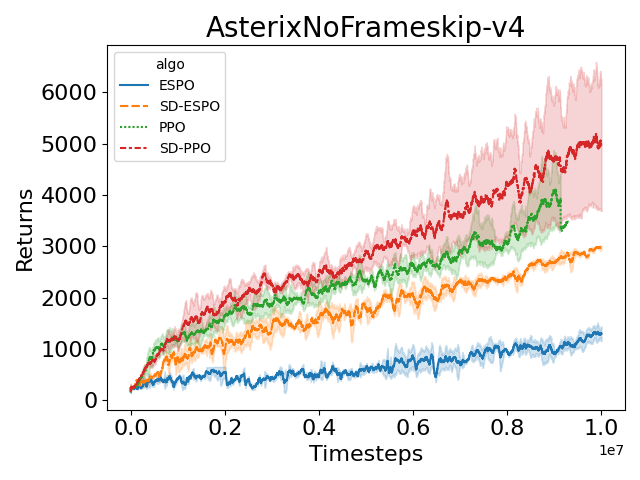}
\includegraphics[width=0.17\textwidth]{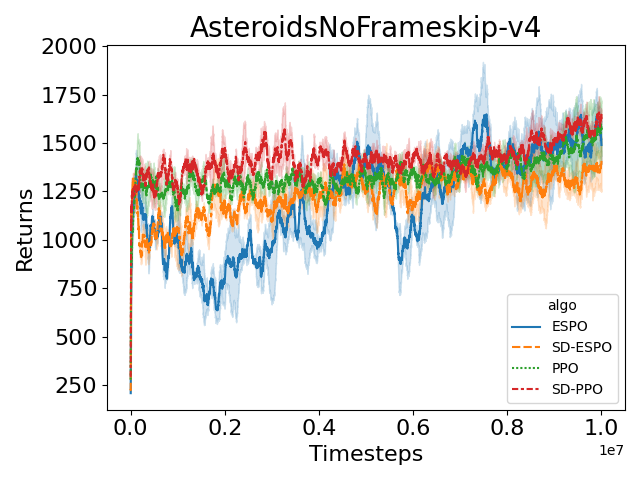}
\includegraphics[width=0.17\textwidth]{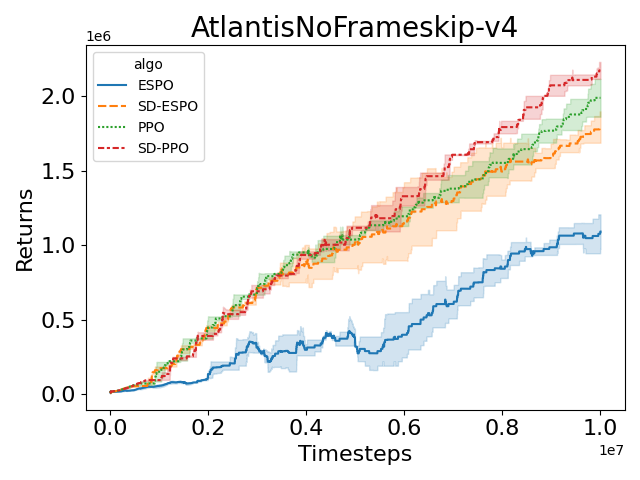}
\includegraphics[width=0.17\textwidth]{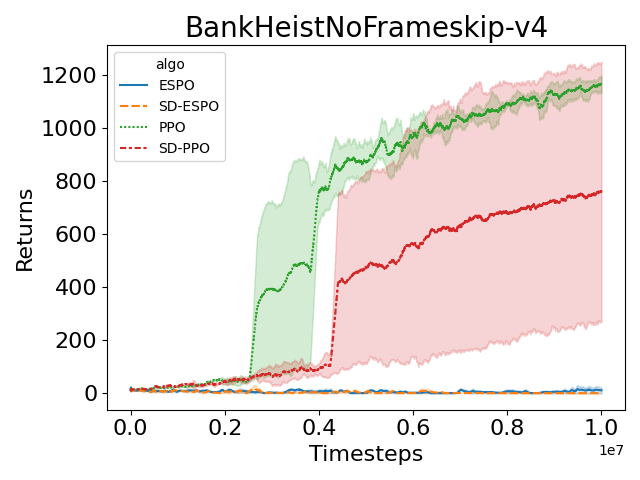}
\includegraphics[width=0.17\textwidth]{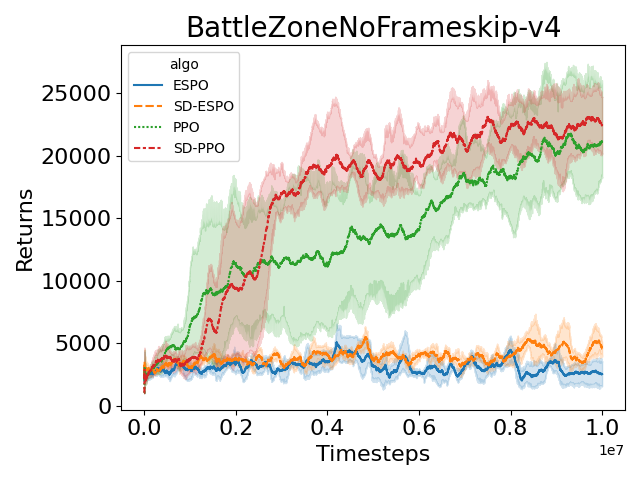}
\includegraphics[width=0.17\textwidth]{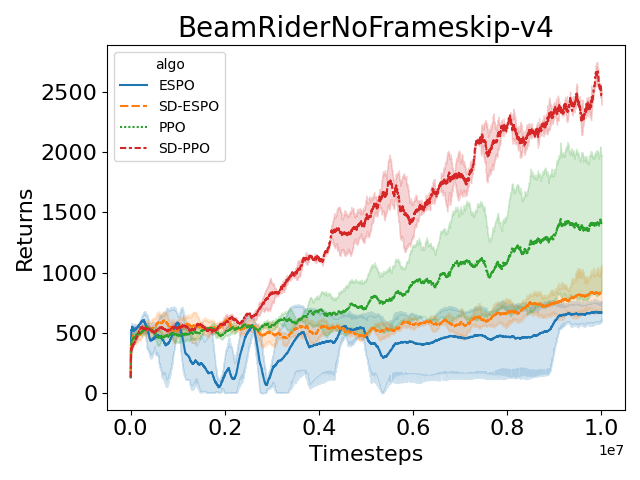}
\includegraphics[width=0.17\textwidth]{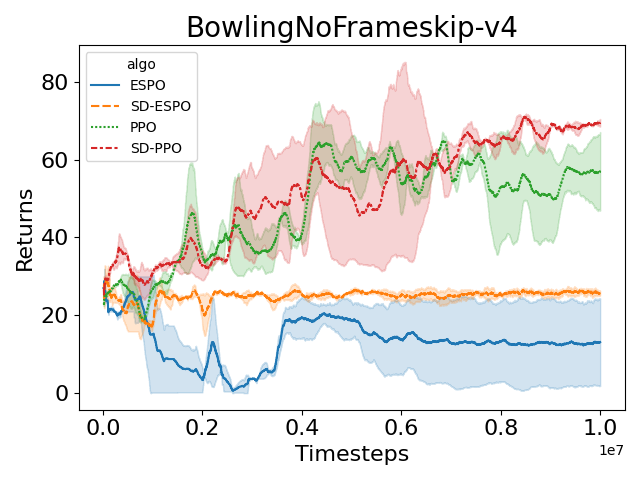}
\includegraphics[width=0.17\textwidth]{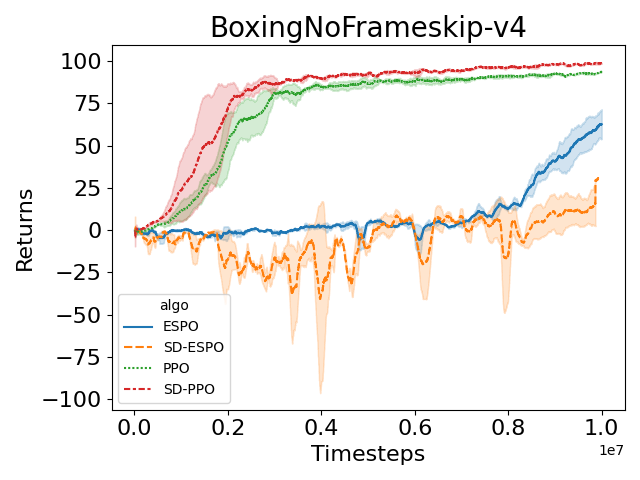}
\includegraphics[width=0.17\textwidth]{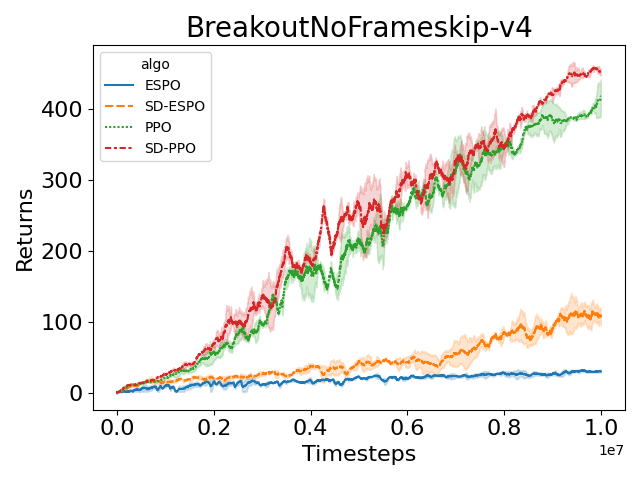}
\includegraphics[width=0.17\textwidth]{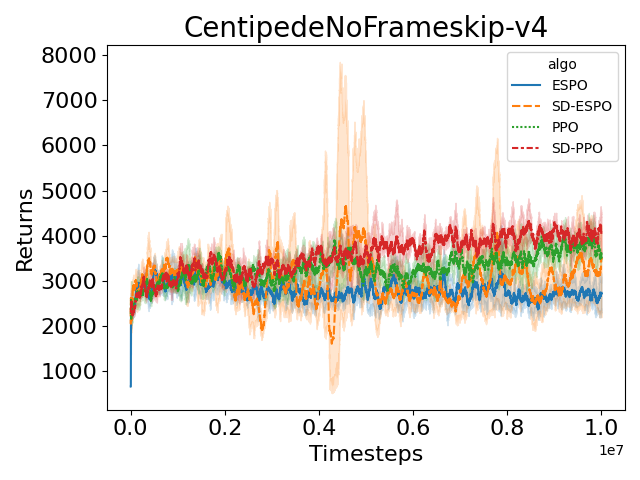}
\includegraphics[width=0.17\textwidth]{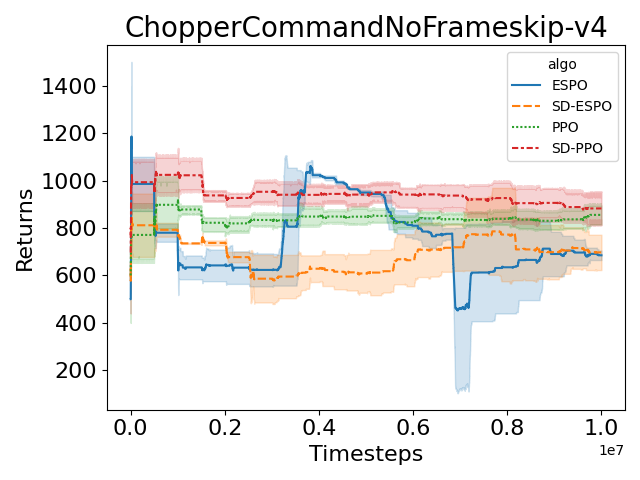}
\includegraphics[width=0.17\textwidth]{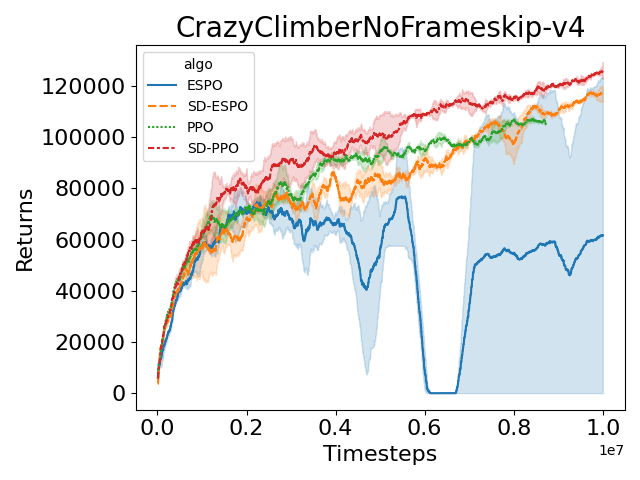}
\includegraphics[width=0.17\textwidth]{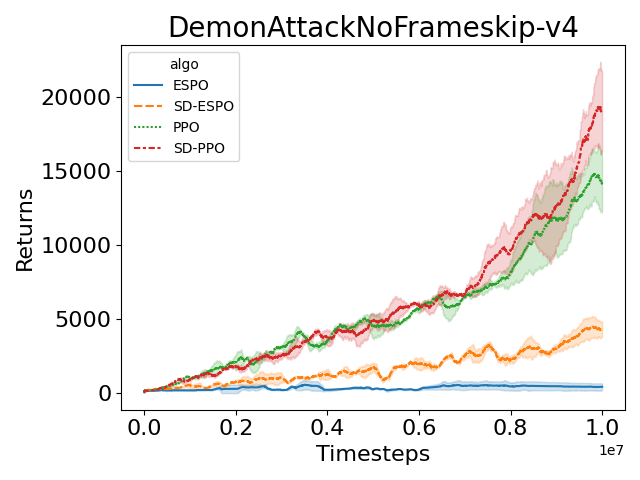}
\includegraphics[width=0.17\textwidth]{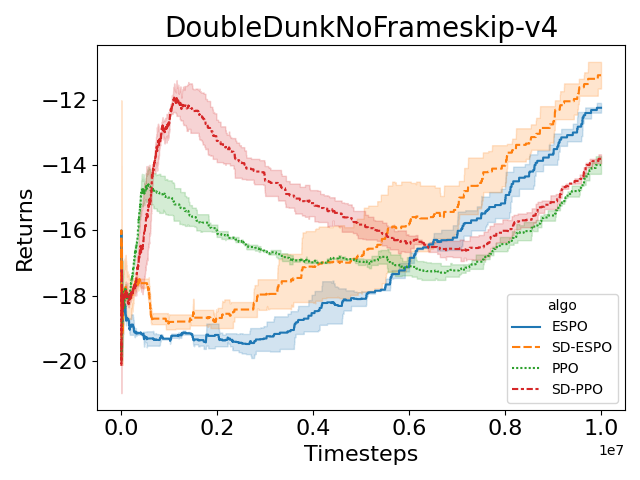}
\includegraphics[width=0.17\textwidth]{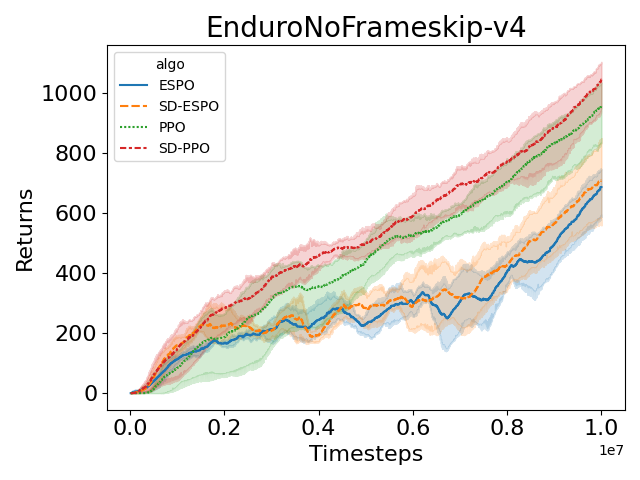}
\includegraphics[width=0.17\textwidth]{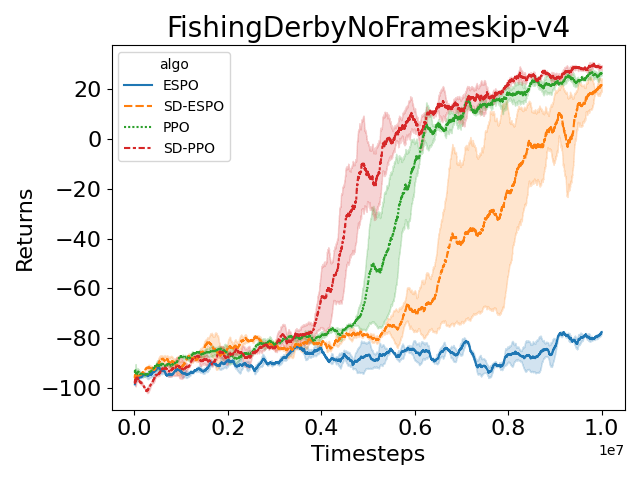}
\includegraphics[width=0.17\textwidth]{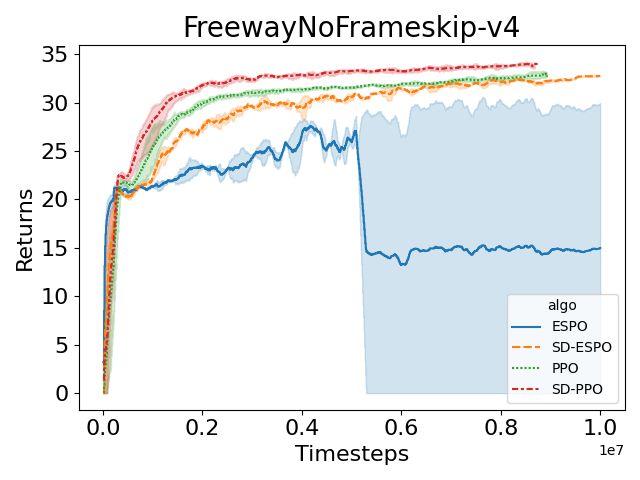}
\includegraphics[width=0.17\textwidth]{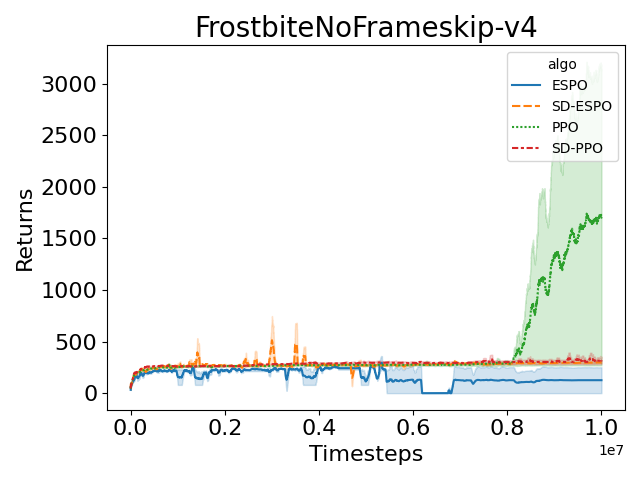}
\includegraphics[width=0.17\textwidth]{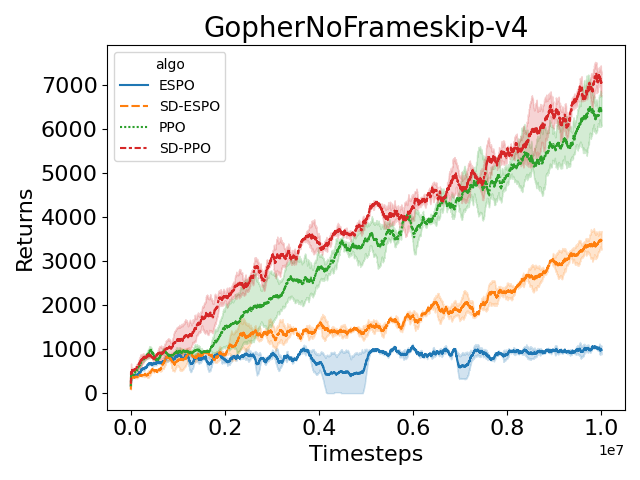}
\includegraphics[width=0.17\textwidth]{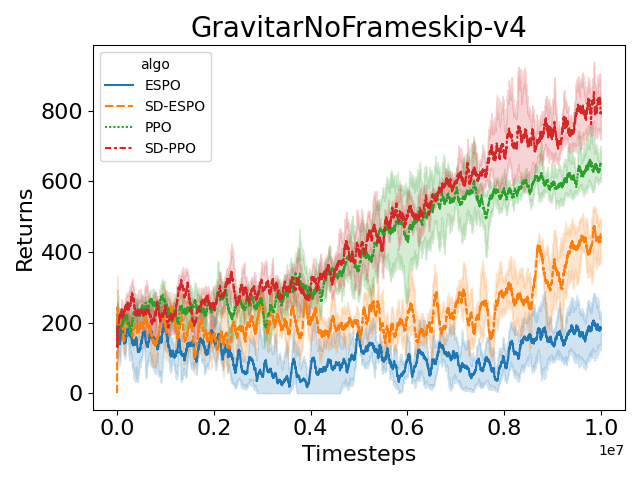}
\includegraphics[width=0.17\textwidth]{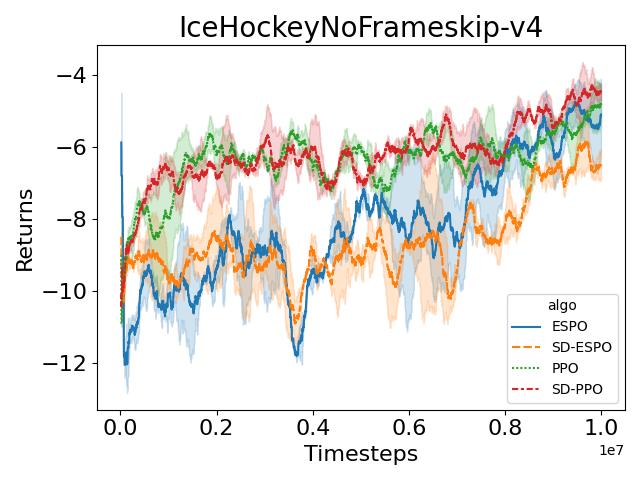}
\includegraphics[width=0.17\textwidth]{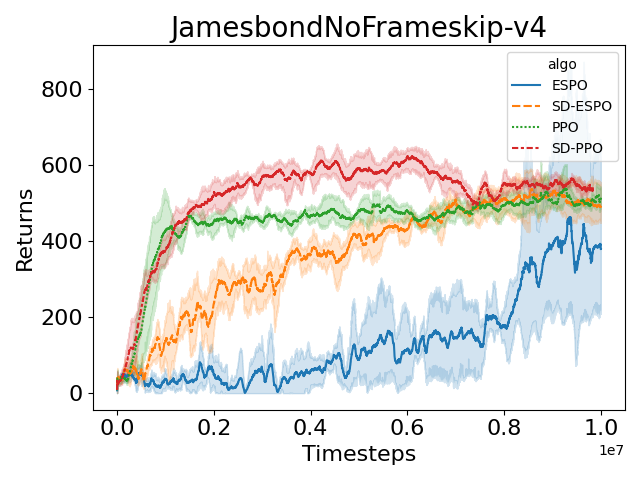}
\includegraphics[width=0.17\textwidth]{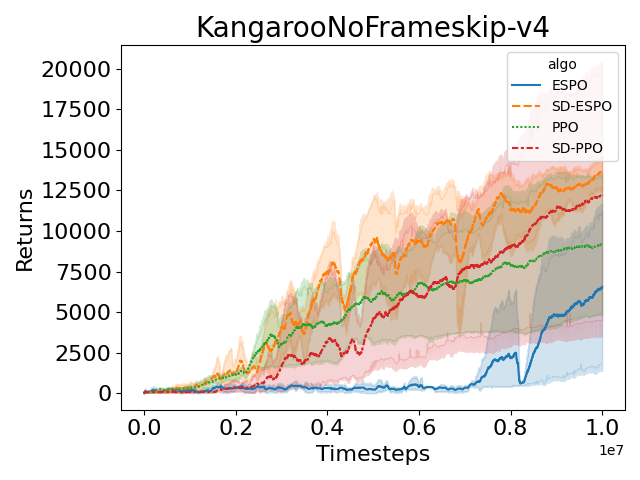}
\includegraphics[width=0.17\textwidth]{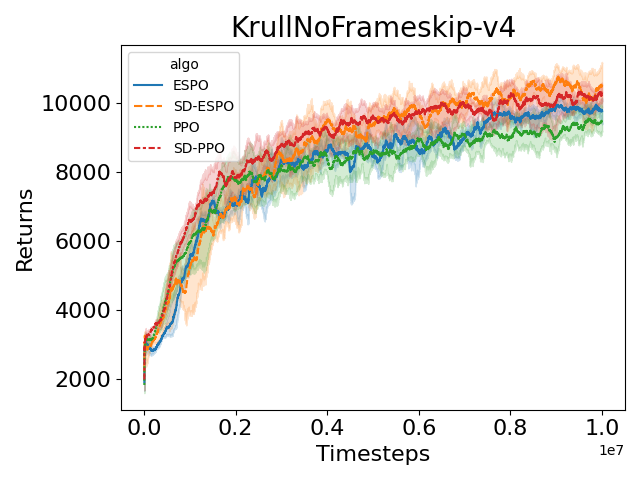}
\includegraphics[width=0.17\textwidth]{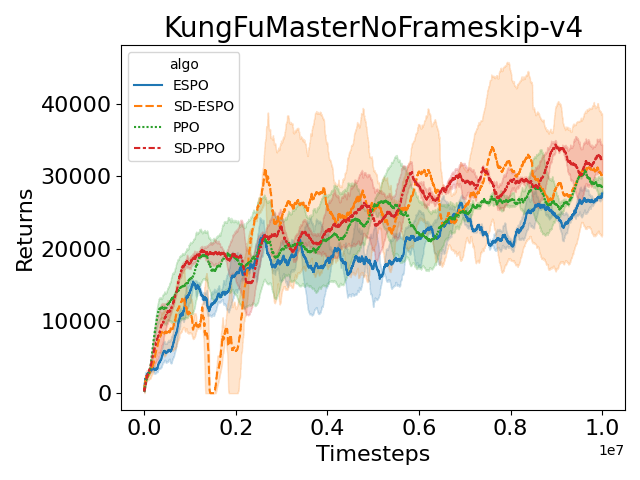}
\includegraphics[width=0.17\textwidth]{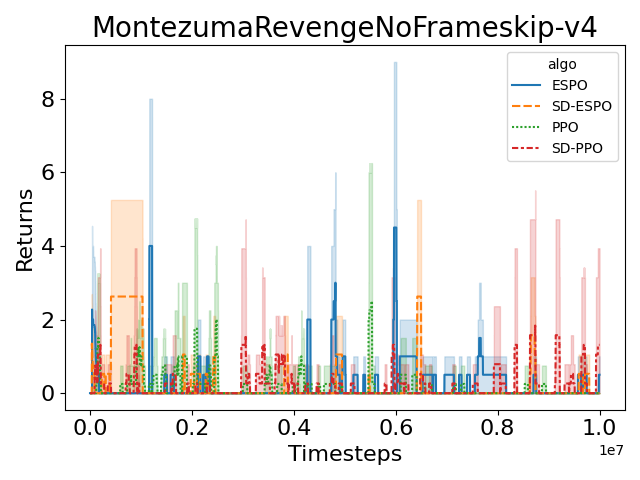}
\includegraphics[width=0.17\textwidth]{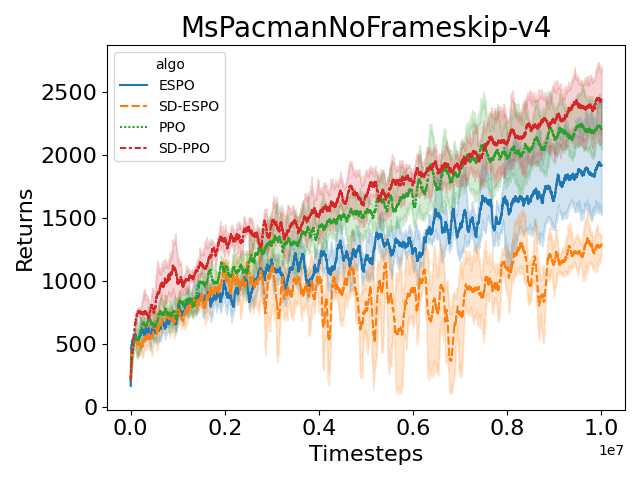}
\includegraphics[width=0.17\textwidth]{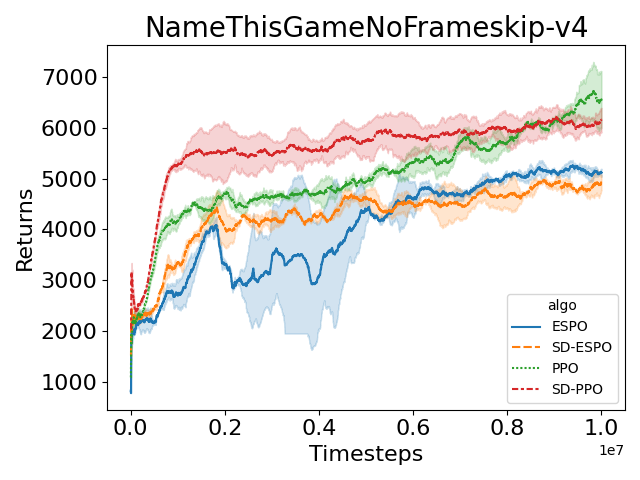}
\includegraphics[width=0.17\textwidth]{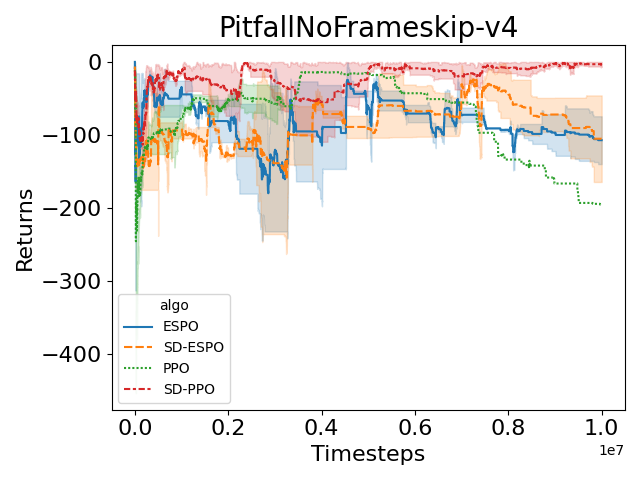}
\includegraphics[width=0.17\textwidth]{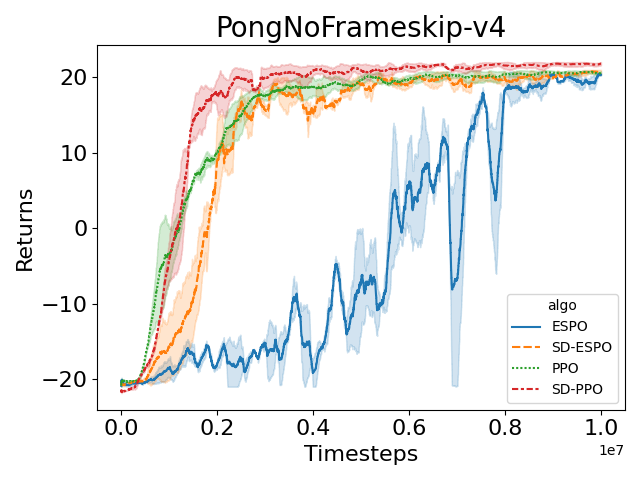}
\includegraphics[width=0.17\textwidth]{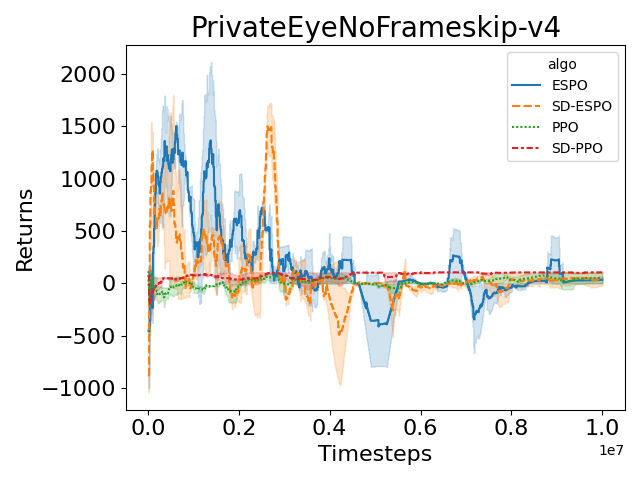}
\includegraphics[width=0.17\textwidth]{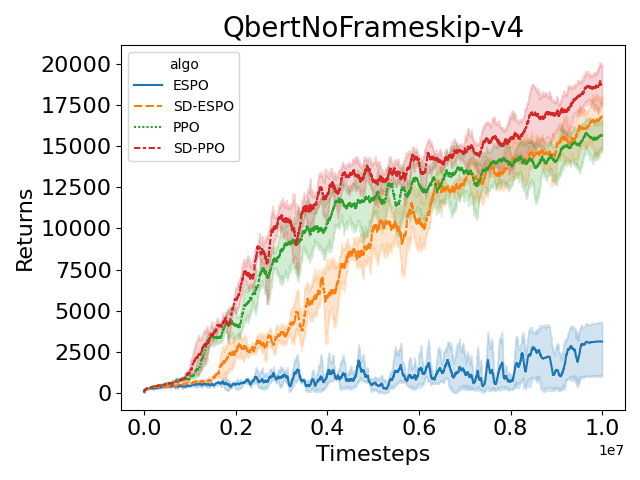}
\includegraphics[width=0.17\textwidth]{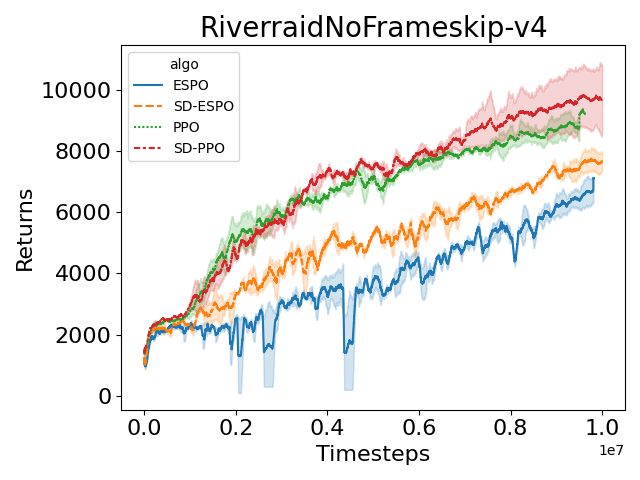}
\includegraphics[width=0.17\textwidth]{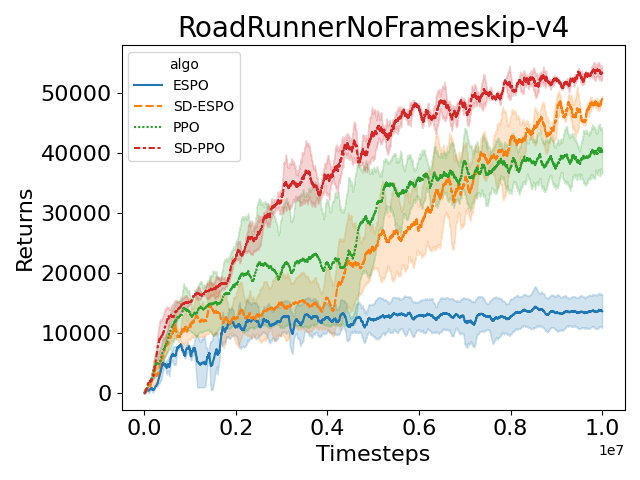}
\includegraphics[width=0.17\textwidth]{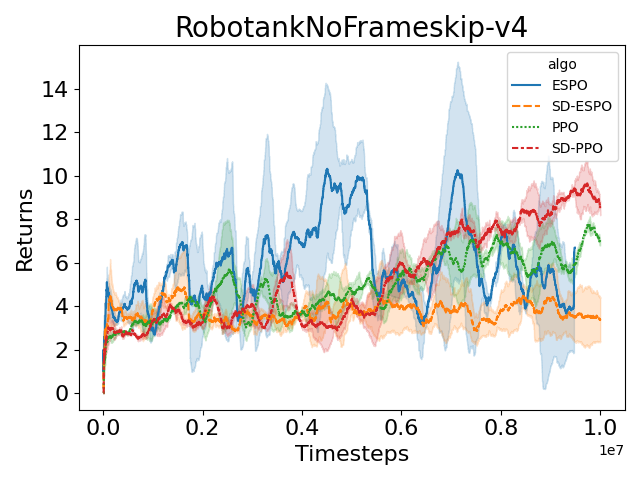}
\includegraphics[width=0.17\textwidth]{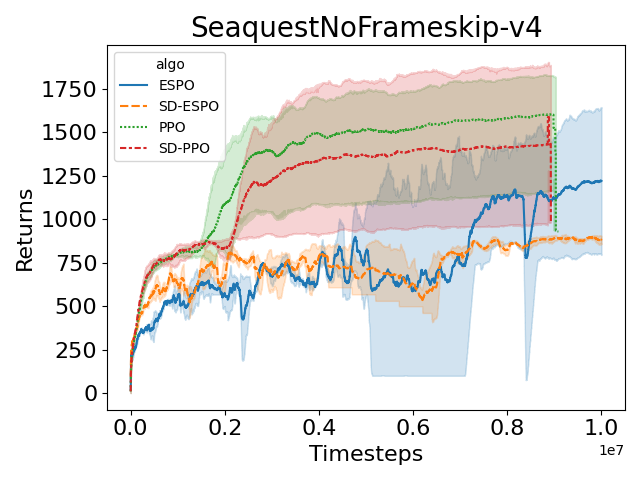}
\includegraphics[width=0.17\textwidth]{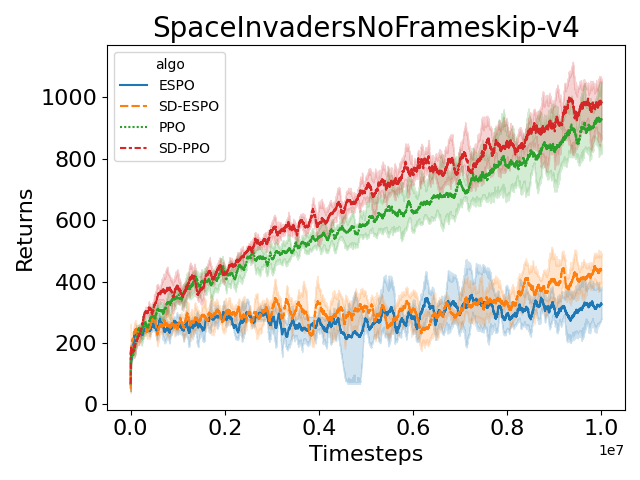}
\includegraphics[width=0.17\textwidth]{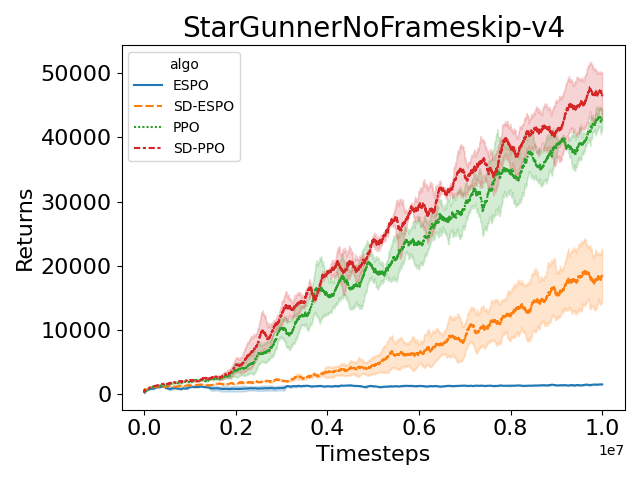}
\includegraphics[width=0.17\textwidth]{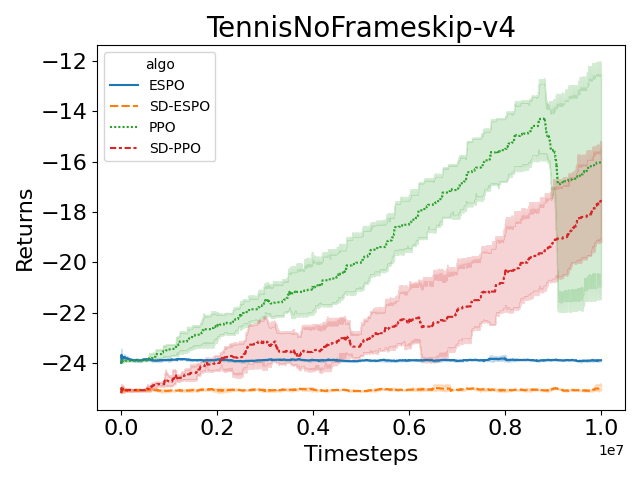}
\includegraphics[width=0.17\textwidth]{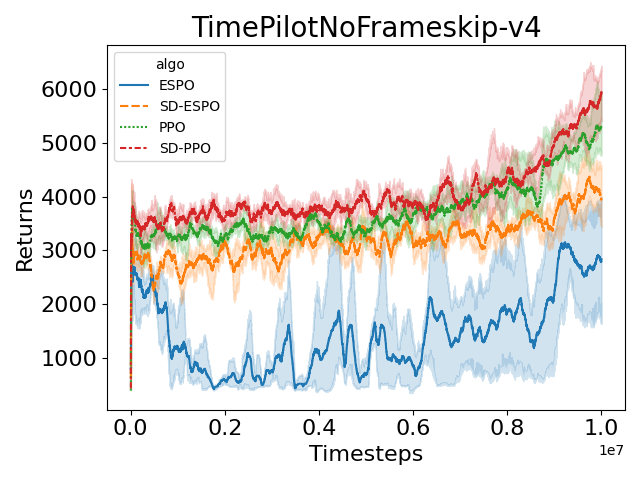}
\includegraphics[width=0.17\textwidth]{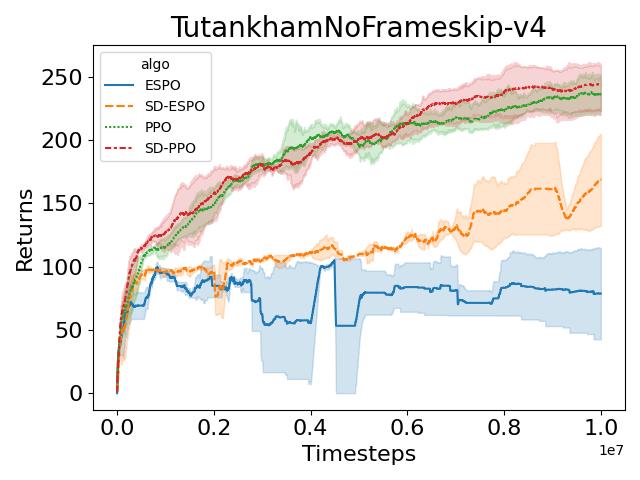}
\includegraphics[width=0.17\textwidth]{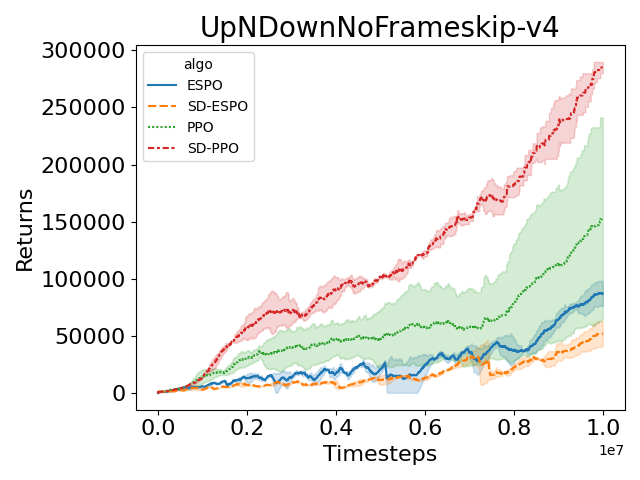}
\includegraphics[width=0.17\textwidth]{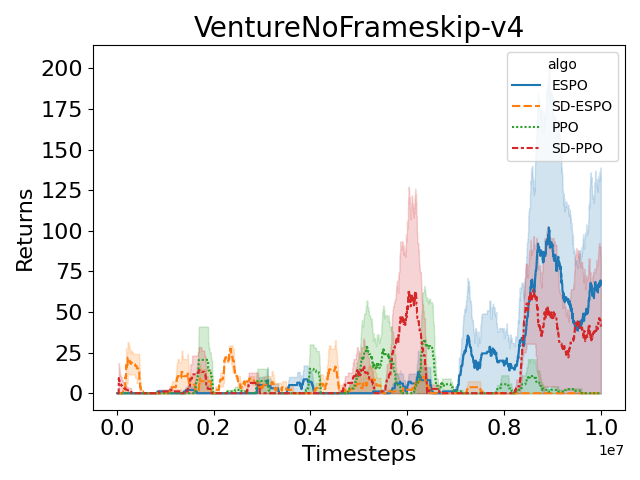}
\includegraphics[width=0.17\textwidth]{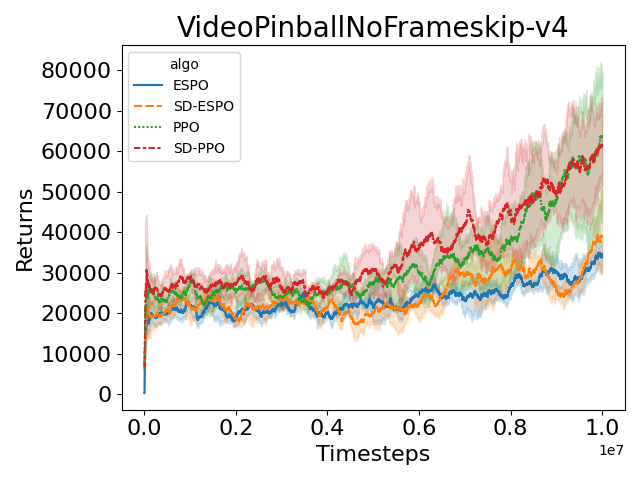}
\includegraphics[width=0.17\textwidth]{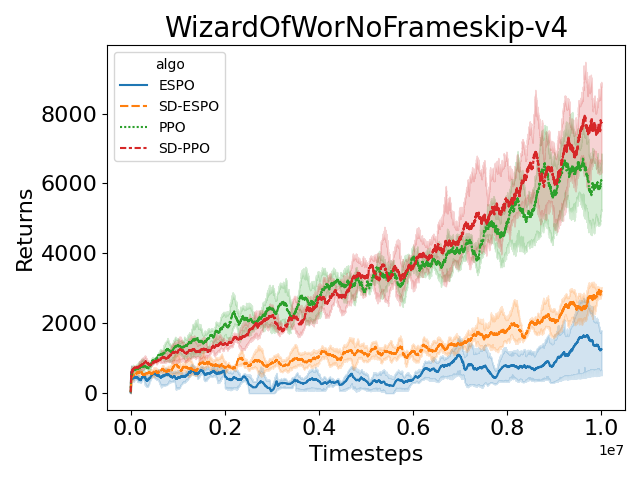}
\includegraphics[width=0.17\textwidth]{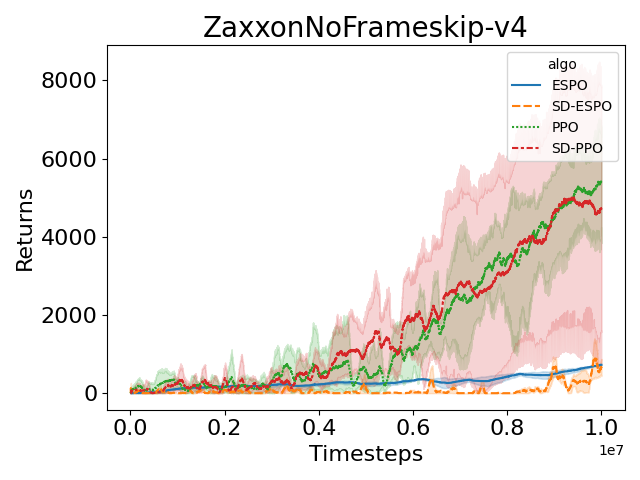}
\caption{Training curves of PPO, SD-PPO, ESPO, and SD-ESPO in Atari games. The curves show that sample dropout successfully boosts the sample efficiency of PPO and ESPO in most tasks.}
\label{fig:atari_all_figs}
\end{figure*}

\begin{figure*}[th!]
\centering
\includegraphics[width=0.17\textwidth]{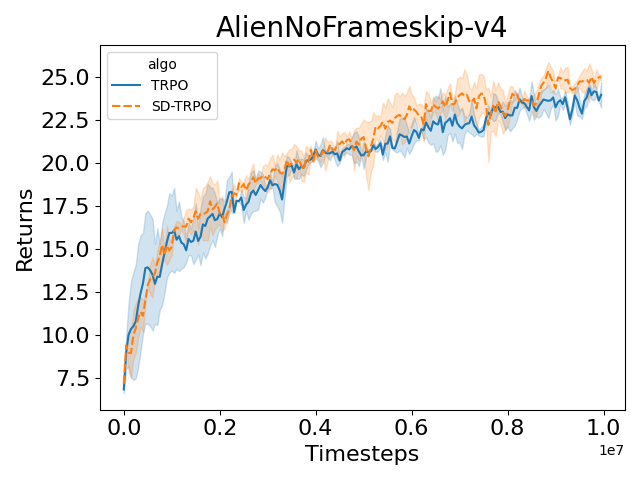}
\includegraphics[width=0.17\textwidth]{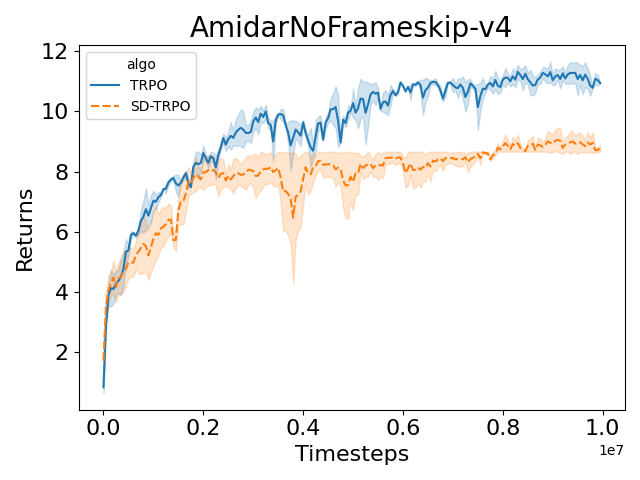}
\includegraphics[width=0.17\textwidth]{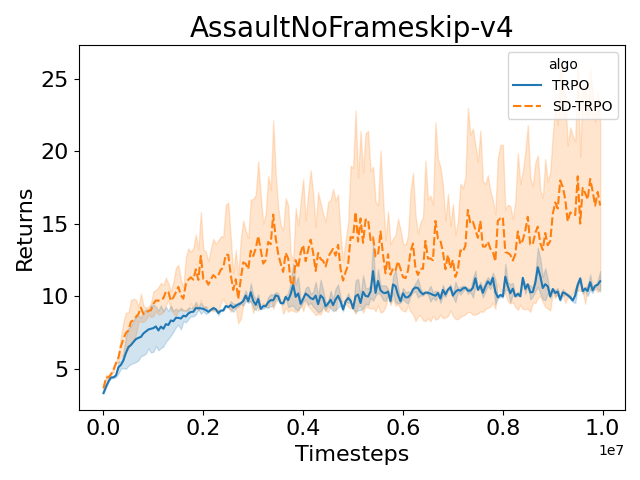}
\includegraphics[width=0.17\textwidth]{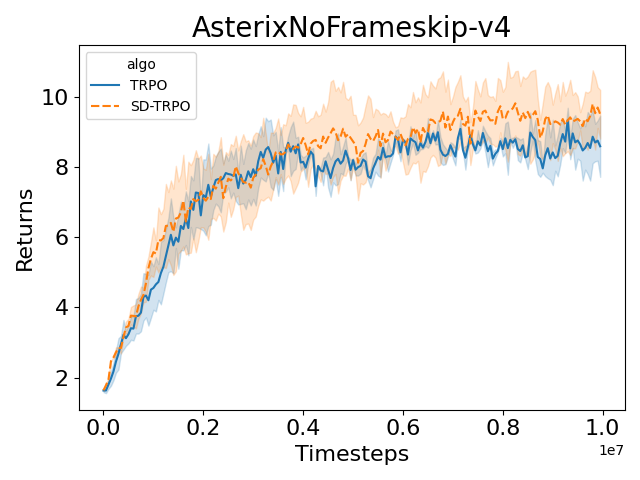}
\includegraphics[width=0.17\textwidth]{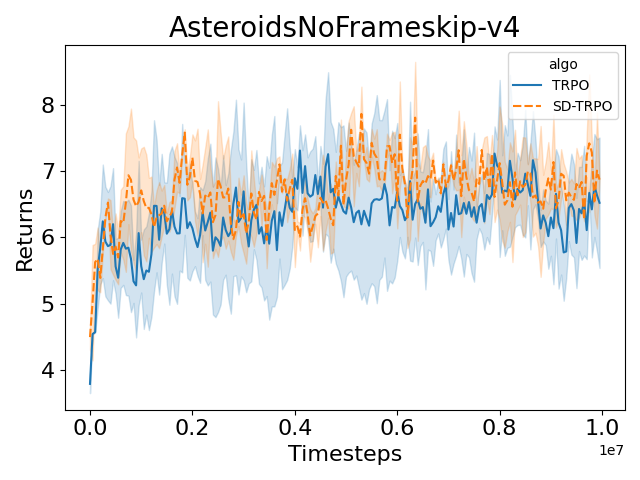}
\includegraphics[width=0.17\textwidth]{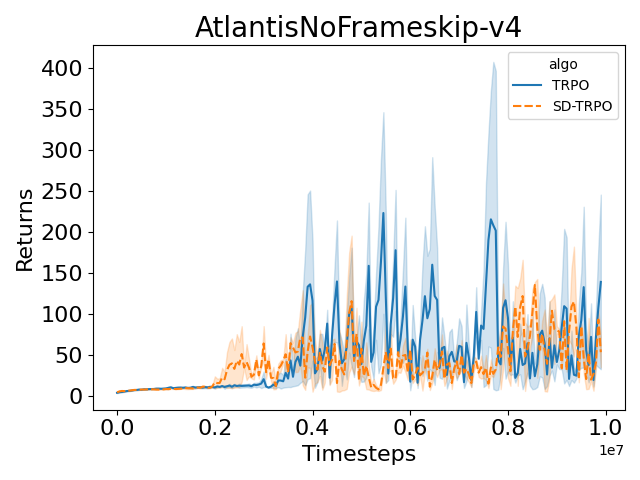}
\includegraphics[width=0.17\textwidth]{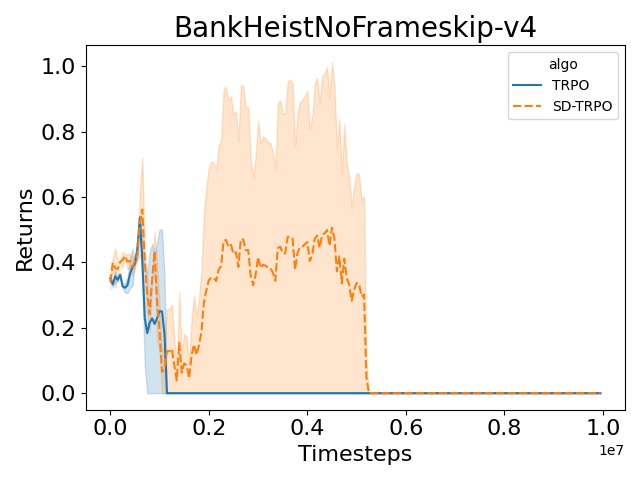}
\includegraphics[width=0.17\textwidth]{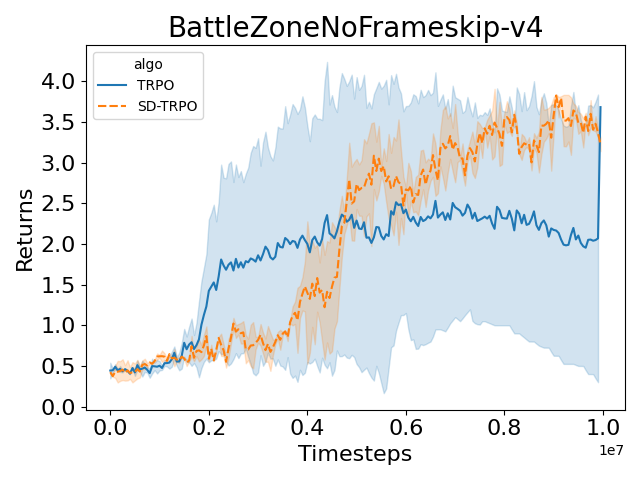}
\includegraphics[width=0.17\textwidth]{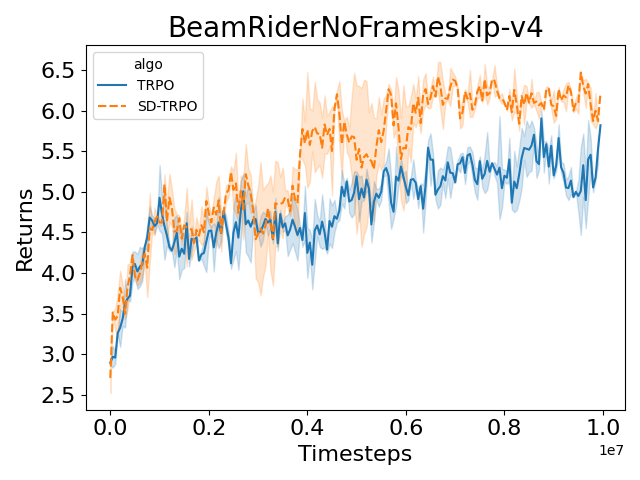}
\includegraphics[width=0.17\textwidth]{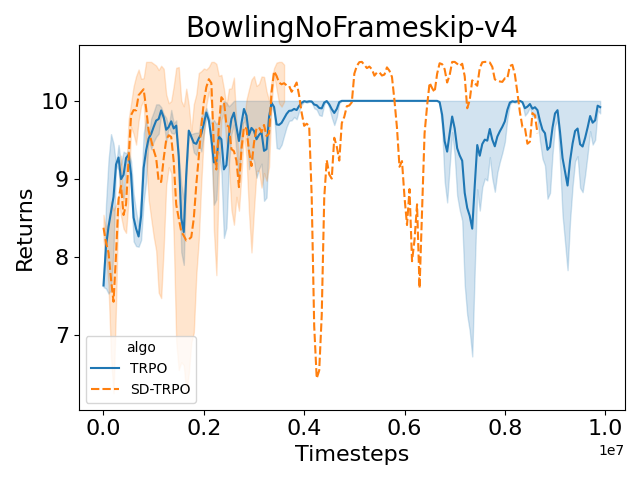}
\includegraphics[width=0.17\textwidth]{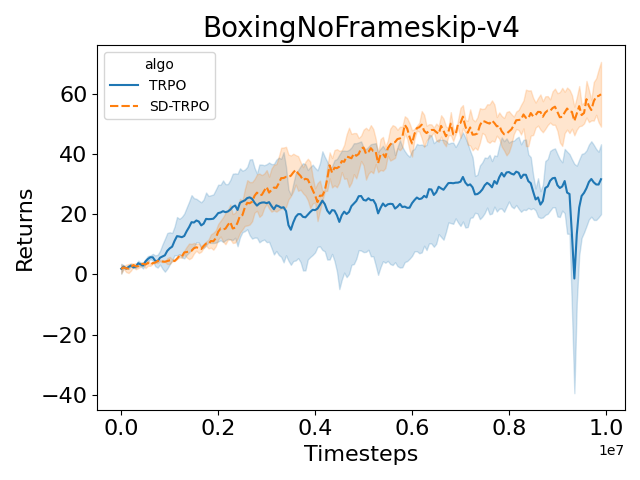}
\includegraphics[width=0.17\textwidth]{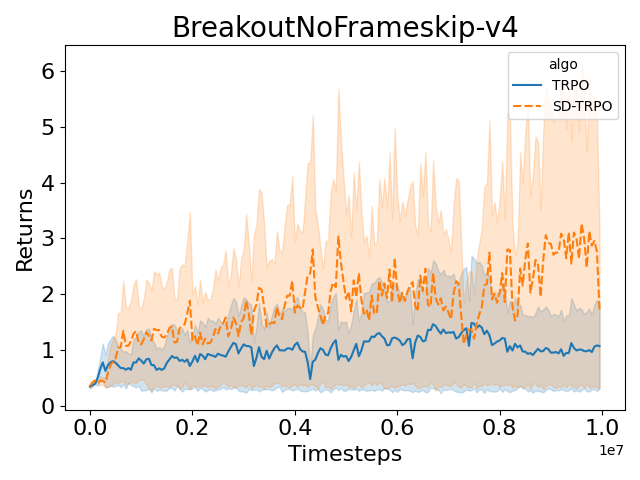}
\includegraphics[width=0.17\textwidth]{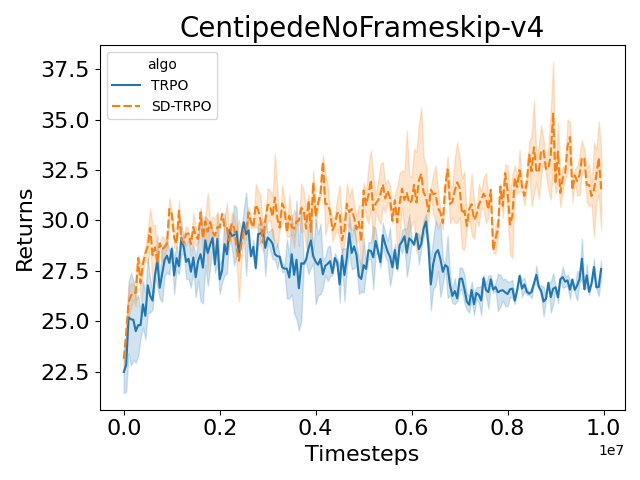}
\includegraphics[width=0.17\textwidth]{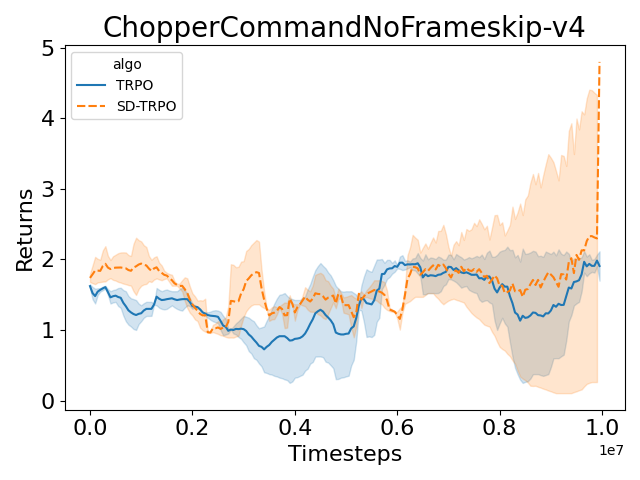}
\includegraphics[width=0.17\textwidth]{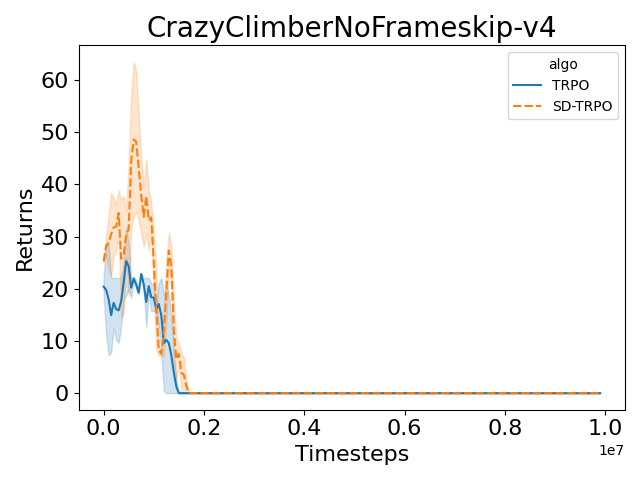}
\includegraphics[width=0.17\textwidth]{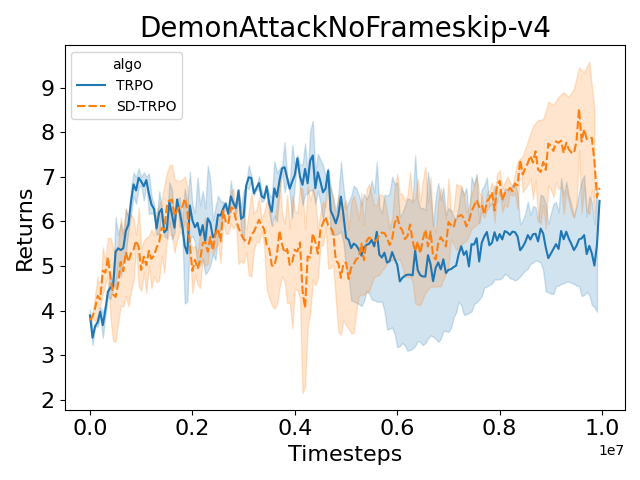}
\includegraphics[width=0.17\textwidth]{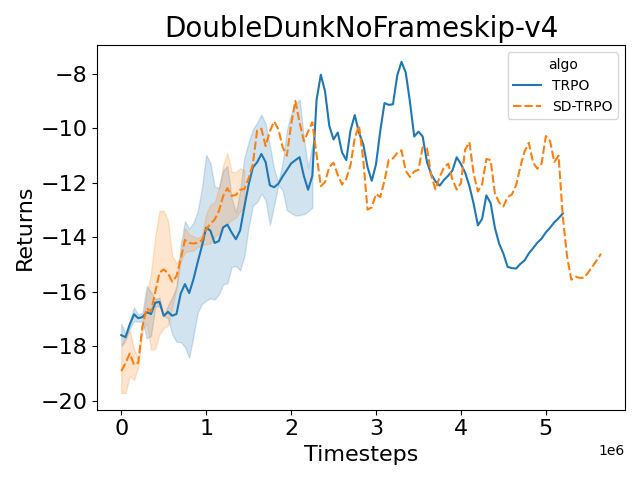}
\includegraphics[width=0.17\textwidth]{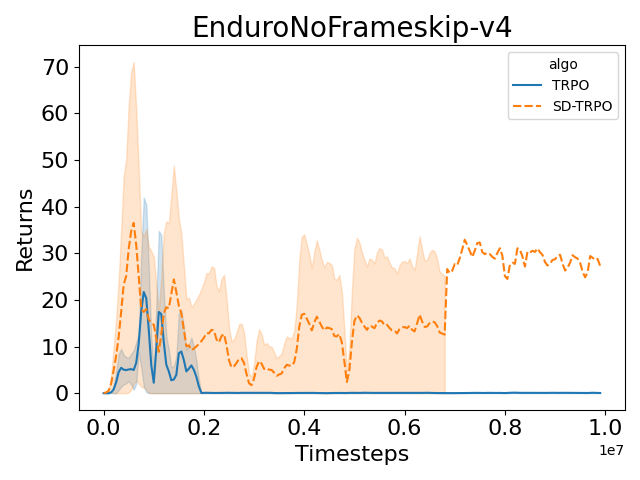}
\includegraphics[width=0.17\textwidth]{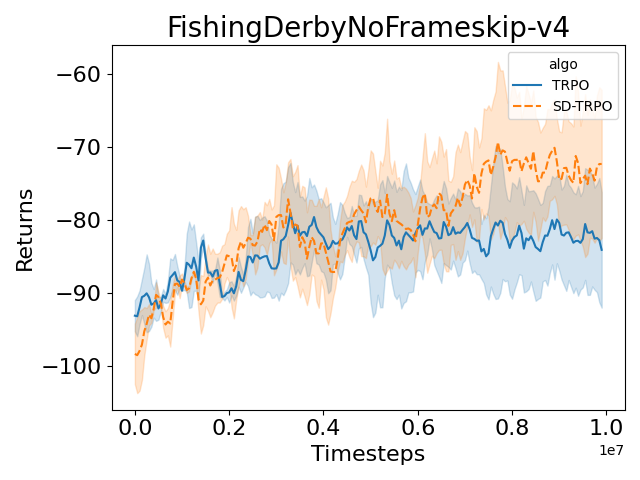}
\includegraphics[width=0.17\textwidth]{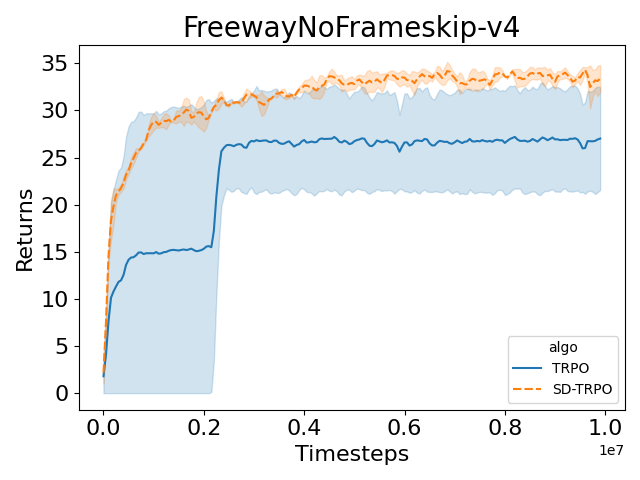}
\includegraphics[width=0.17\textwidth]{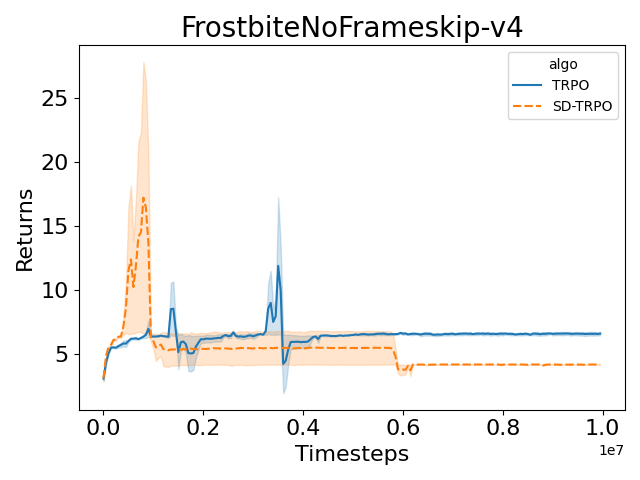}
\includegraphics[width=0.17\textwidth]{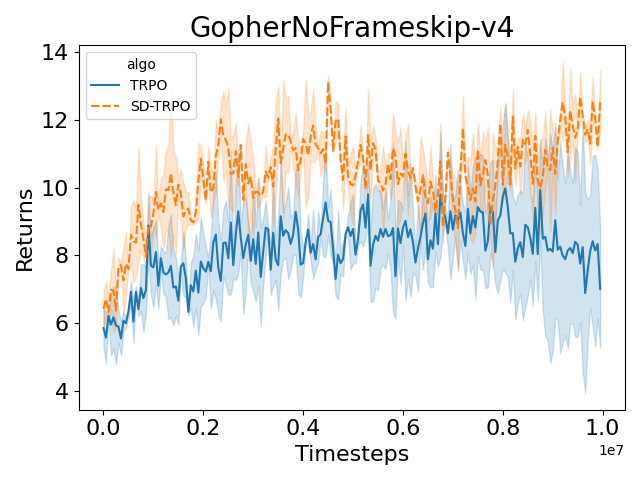}
\includegraphics[width=0.17\textwidth]{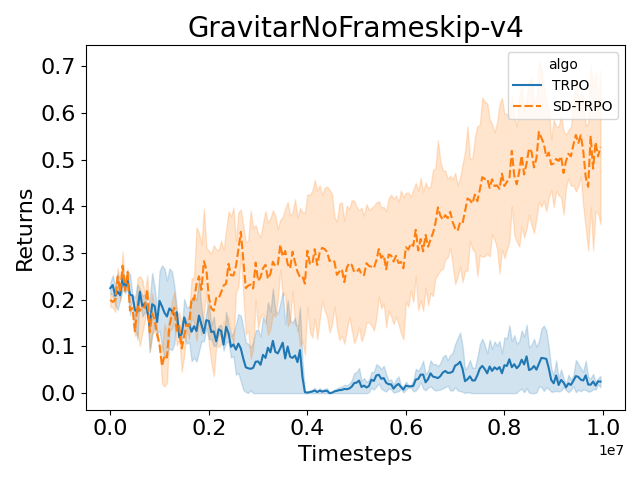}
\includegraphics[width=0.17\textwidth]{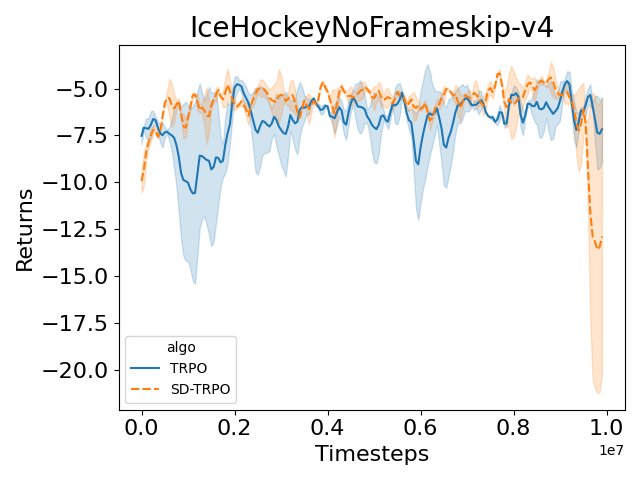}
\includegraphics[width=0.17\textwidth]{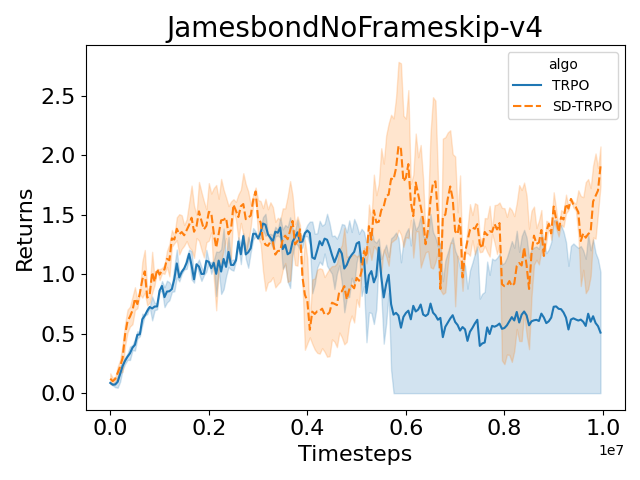}
\includegraphics[width=0.17\textwidth]{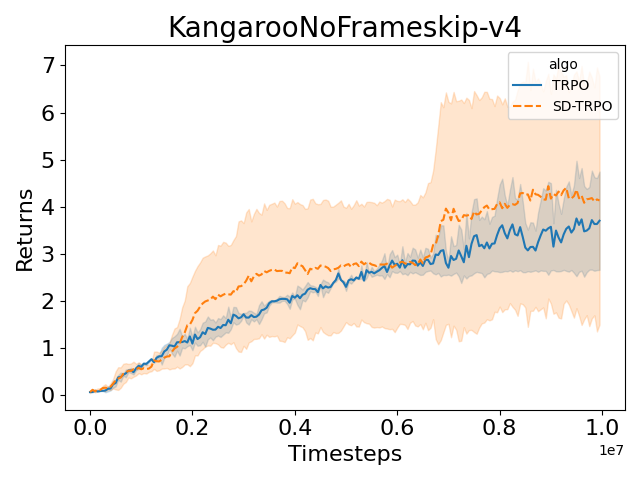}
\includegraphics[width=0.17\textwidth]{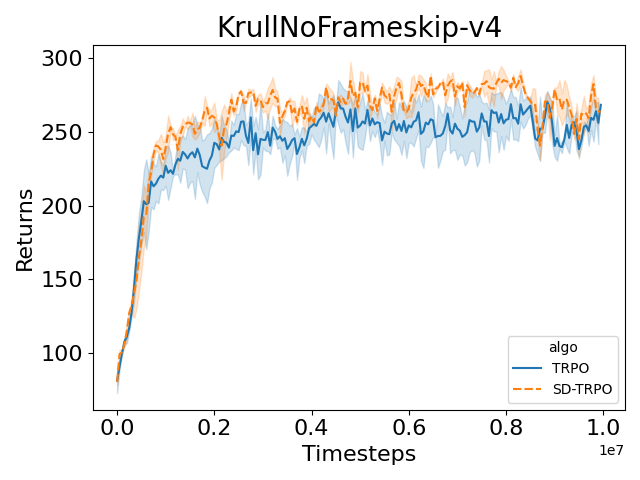}
\includegraphics[width=0.17\textwidth]{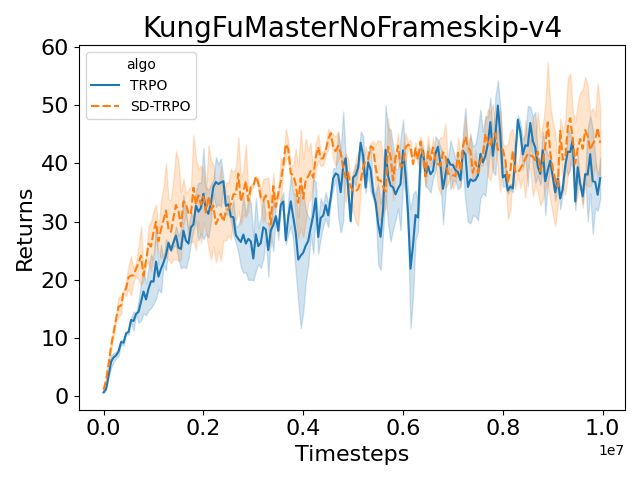}
\includegraphics[width=0.17\textwidth]{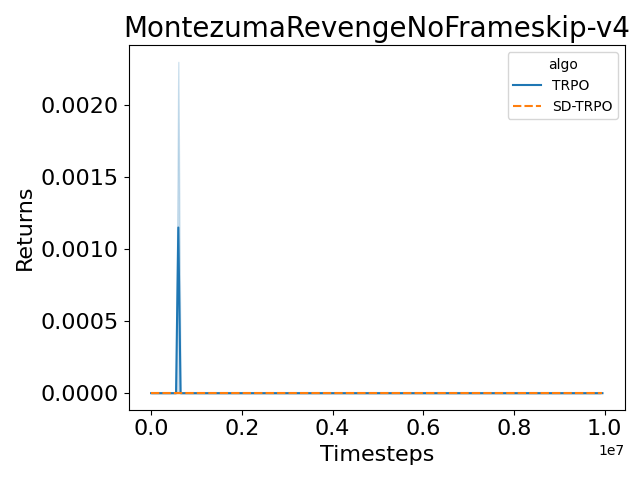}
\includegraphics[width=0.17\textwidth]{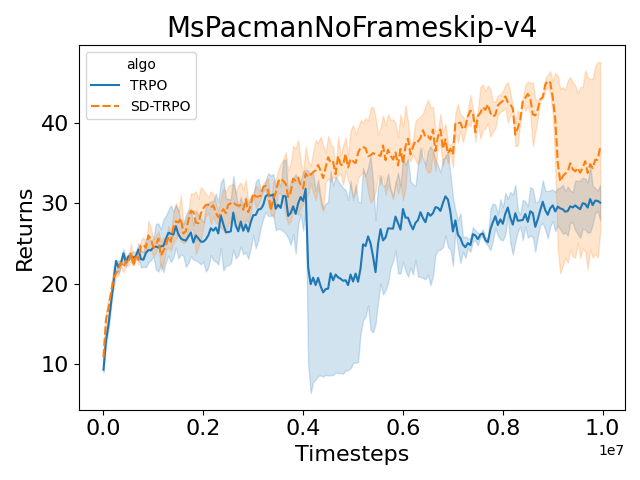}
\includegraphics[width=0.17\textwidth]{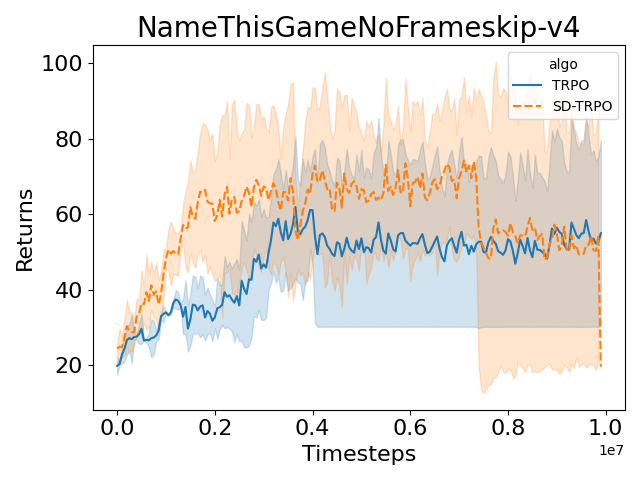}
\includegraphics[width=0.17\textwidth]{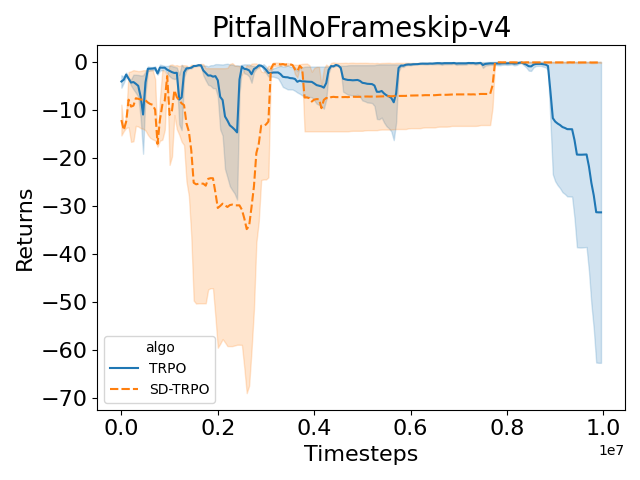}
\includegraphics[width=0.17\textwidth]{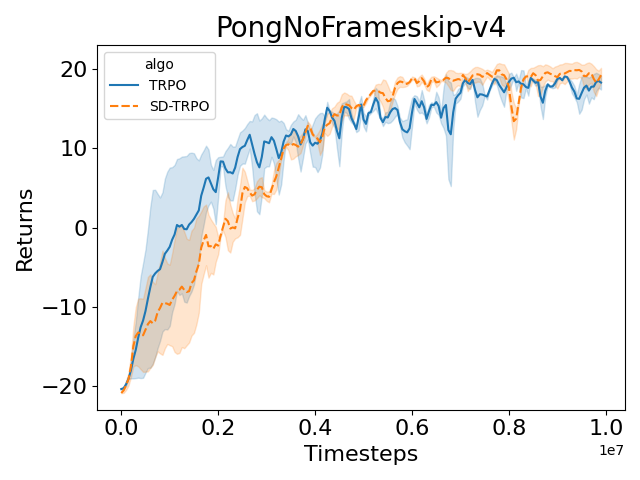}
\includegraphics[width=0.17\textwidth]{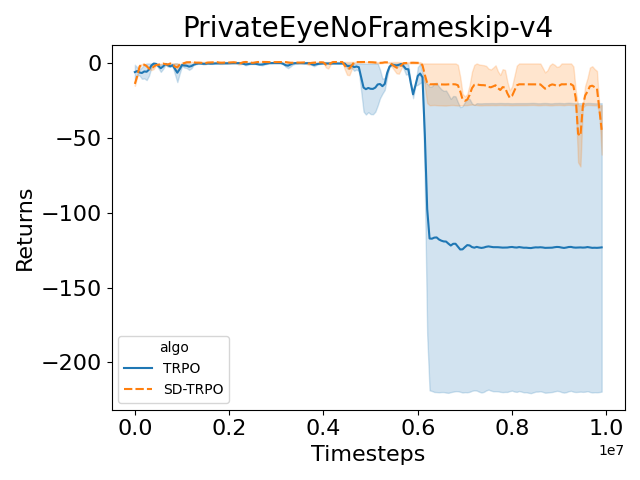}
\includegraphics[width=0.17\textwidth]{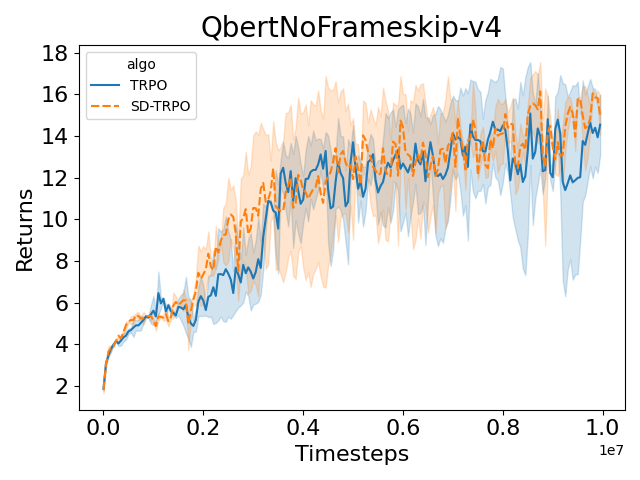}
\includegraphics[width=0.17\textwidth]{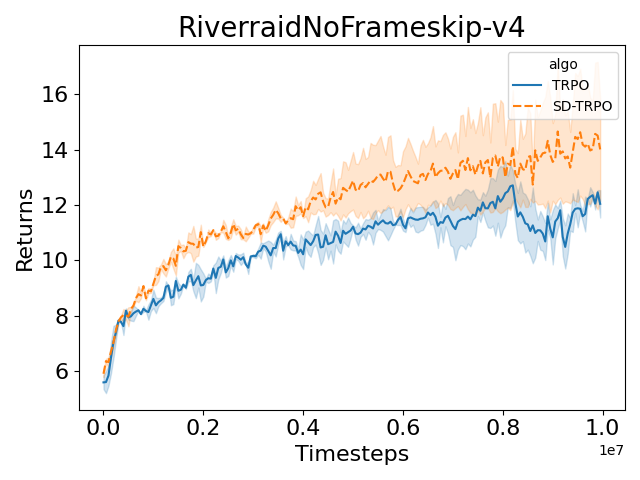}
\includegraphics[width=0.17\textwidth]{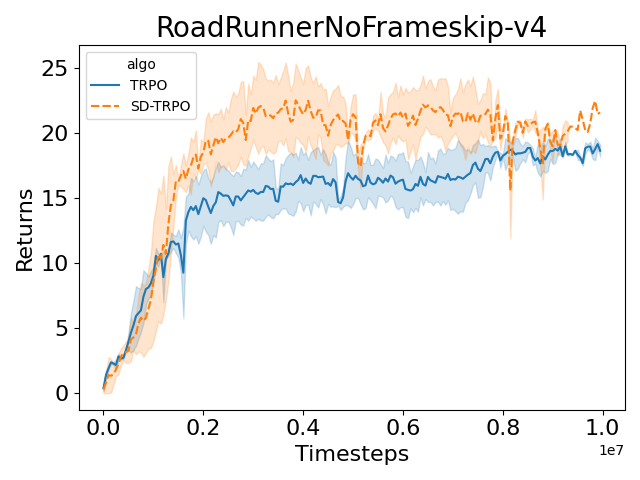}
\includegraphics[width=0.17\textwidth]{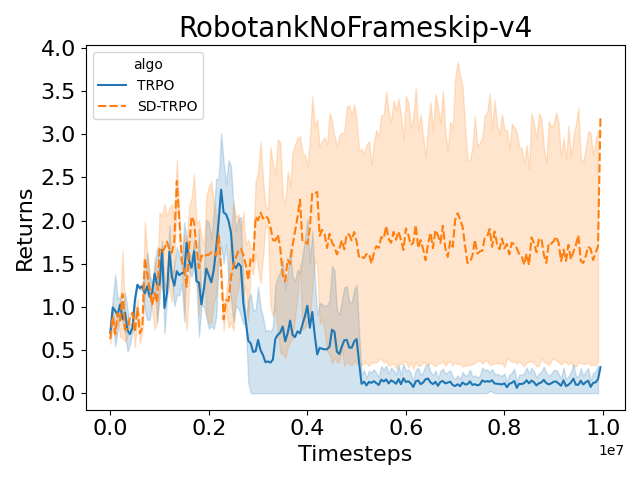}
\includegraphics[width=0.17\textwidth]{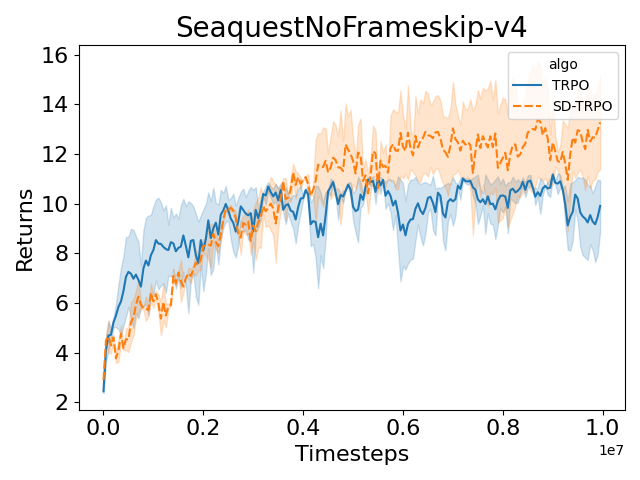}
\includegraphics[width=0.17\textwidth]{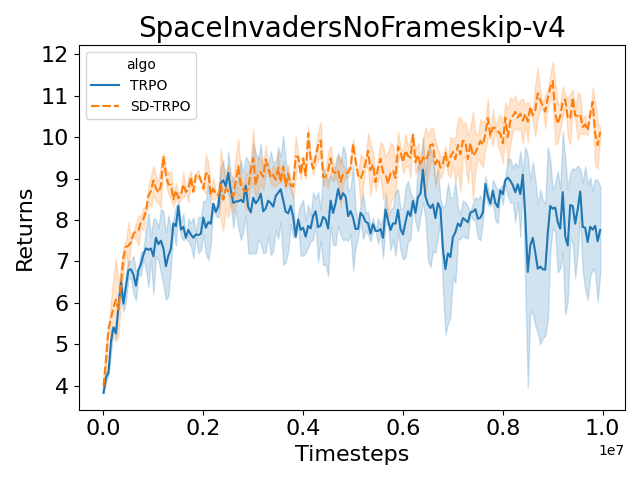}
\includegraphics[width=0.17\textwidth]{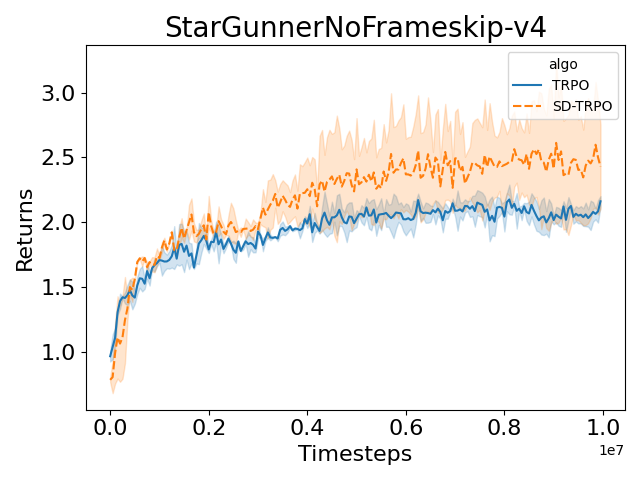}
\includegraphics[width=0.17\textwidth]{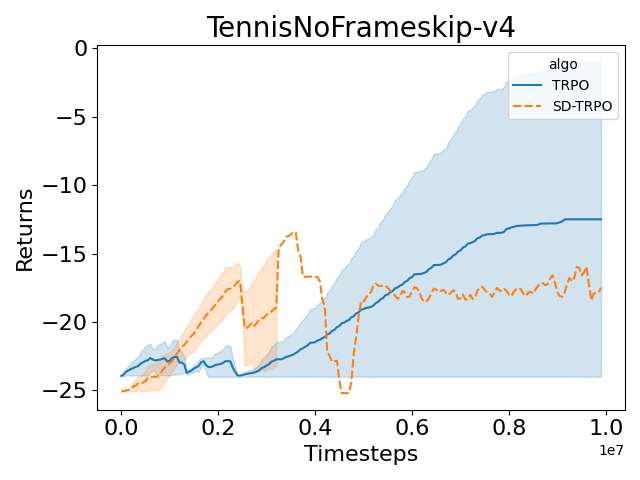}
\includegraphics[width=0.17\textwidth]{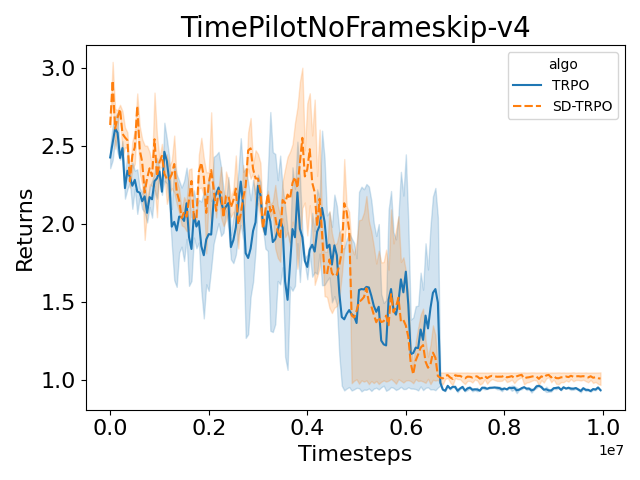}
\includegraphics[width=0.17\textwidth]{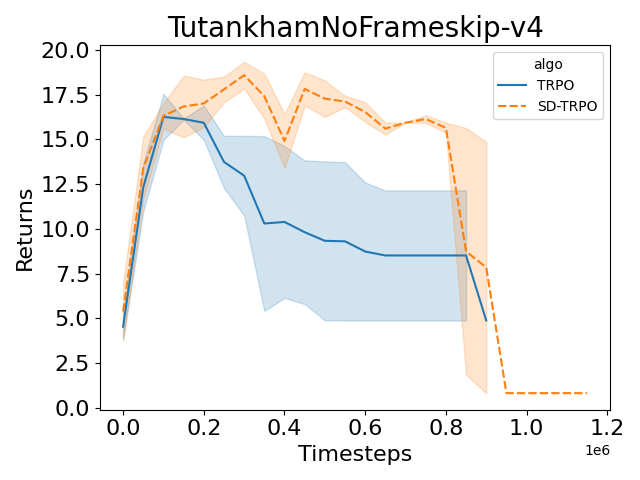}
\includegraphics[width=0.17\textwidth]{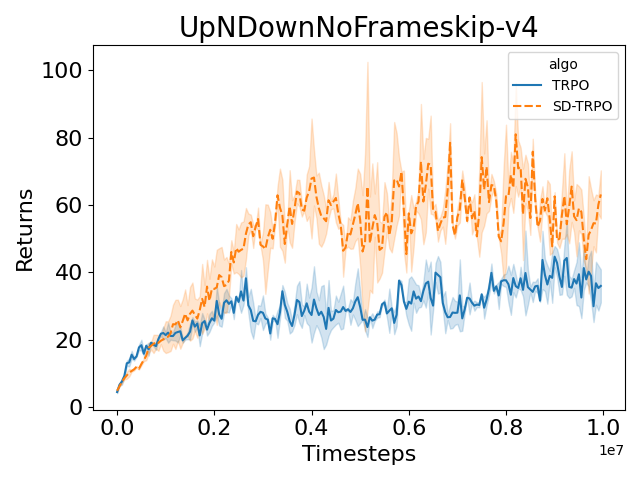}
\includegraphics[width=0.17\textwidth]{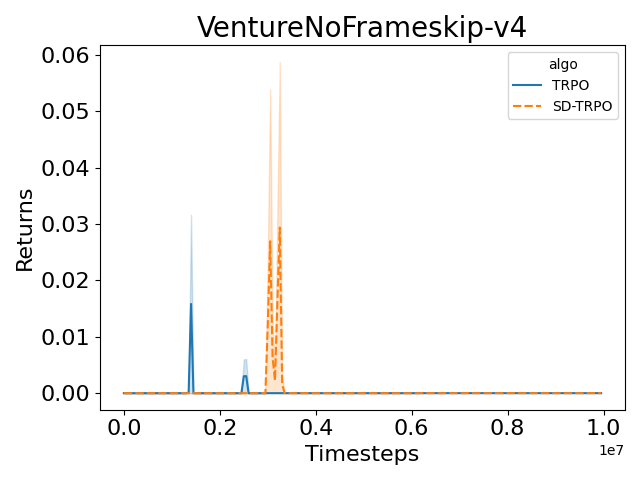}
\includegraphics[width=0.17\textwidth]{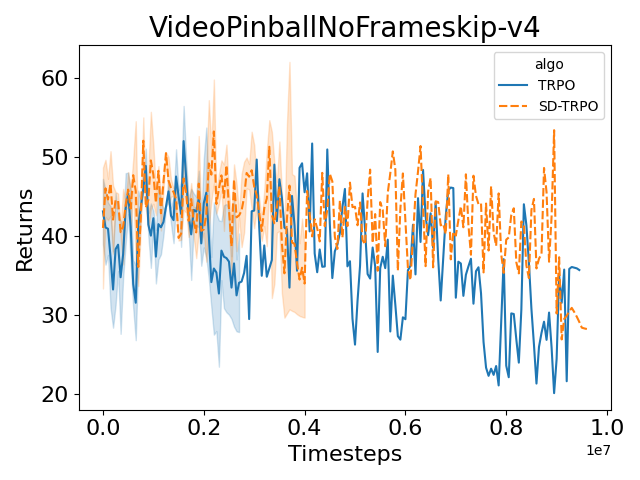}
\includegraphics[width=0.17\textwidth]{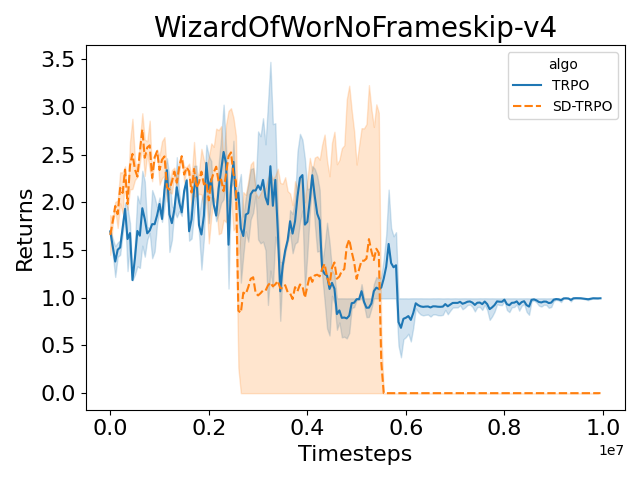}
\includegraphics[width=0.17\textwidth]{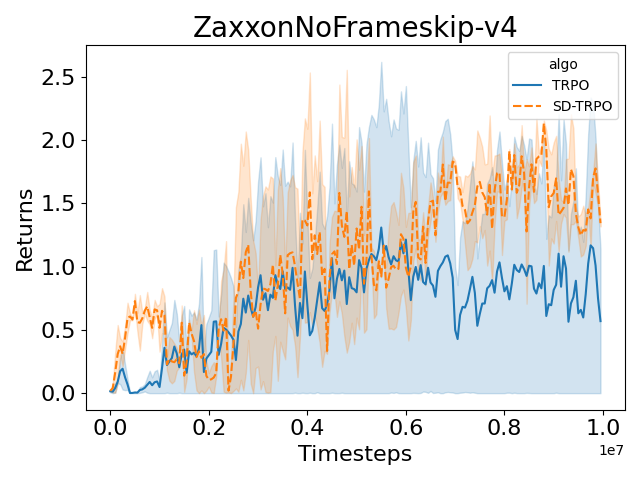}
\caption{Training curves of TRPO and SD-TRPO in Atari games. Note that the scores of TRPO and SD-TRPO in atari games are much lower than those of PPO and ESPO. For better paper presentation, we plot the training curves of atari games separately. The curves show that sample dropout successfully boosts the performance of TRPO in most tasks.}
\label{fig:atari_all_figs_trpo}
\end{figure*}

\paragraph{Atari games} We also test our algorithms in 49 atari games. 
Following common practices~\cite{van2016deep}, we compare these algorithms in training curves and human normalized scores. Specifically, we normalize the score for each game as follows:
\begin{equation}
    \textup{score}_{\textup{normalized}} =
    \frac{\textup{score}_{\textup{agent}} -  \textup{score}_{\textup{random}} }{\textup{score}_{\textup{human}} -  \textup{score}_{\textup{random}} },
\end{equation}
where $\textup{score}_{\textup{human}}$ and $\textup{score}_{\textup{random}}$ are taken from \cite{van2016deep}. Table~\ref{app:atari_norm_score} presents the comparison on the mean and median of human normalized scores at 10 million of training samples for 49 atari games. 
We find that the sample dropout technique can also improve PPO, ESPO and TRPO in the discrete-action tasks. 
Specifically, the sample dropout technique boosts the mean scores by \textbf{59\%, 110\%, 13\%} for PPO, ESPO and TRPO respectively. 
We notice that in atari games, the proximity-based algorithms (i.e., PPO, ESPO) achieves much better results than the trust-region based algorithm (i.e., TRPO).
For better paper presentation, we show the comparison of training curves in Figure~\ref{fig:atari_all_figs} (for PPO and ESPO) and Figure~\ref{fig:atari_all_figs_trpo} (for TRPO) respectively. 
Notably, ESPO struggles to learn in many environments (e.g., Kangaroo, Qbert, Amidar), while SD-ESPO learns efficiently and outperforms ESPO by a clear margin. 
Table~\ref{app:all_atari_score} shows the raw scores for each game. In general, out of 49 Atari environments, SD-PPO outperforms PPO in \textbf{43} Atari games, SD-ESPO beats ESPO in \textbf{38} Atari games, and SD-TRPO wins TRPO in \textbf{35} Atari games. 



\begin{table}[ht!]
    \centering
    \scriptsize
    \begin{tabular}{ccc|cc|cc} 
    \toprule
     & PPO & SD-PPO & ESPO & SD-ESPO & TRPO & SD-TRPO \\
    \midrule
    Mean & 615\% & \textbf{674\%} & 293\% & \textbf{403\%} & -22\% & \textbf{-9\%} \\
    Median & 112\% & \textbf{119\%} & 17\% & \textbf{56\%} & -2.7\% & \textbf{-2.5\%} \\
    \bottomrule
    \end{tabular}
\caption{Human normalized score in Atari games. The results verify the effectiveness of sample dropout in discrete action tasks.} \label{app:atari_norm_score}
\end{table}

\begin{figure*}[h]
	\centering
	\includegraphics[width=0.25\textwidth]{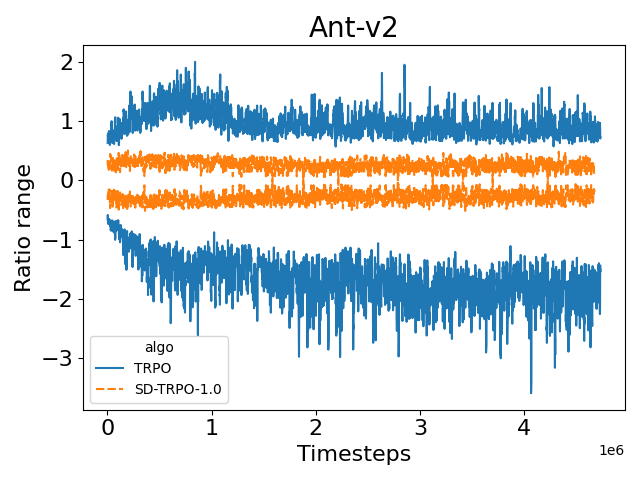}
    \includegraphics[width=0.25\textwidth]{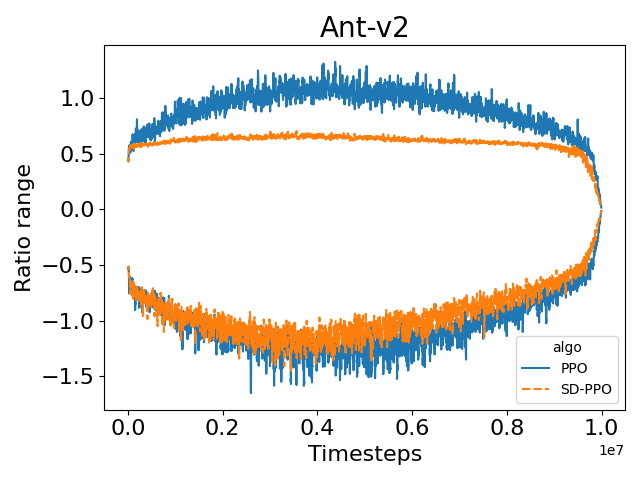}
    \includegraphics[width=0.25\textwidth]{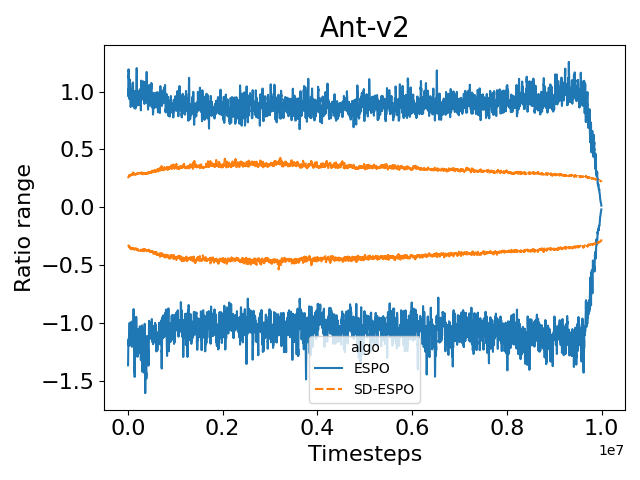}
    \includegraphics[width=0.25\textwidth]{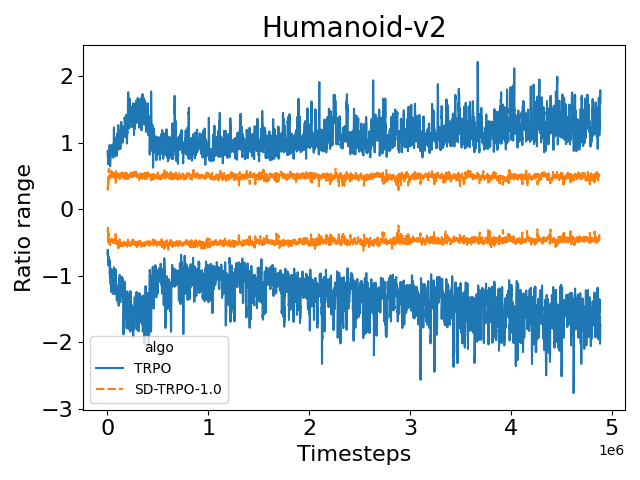}
    \includegraphics[width=0.25\textwidth]{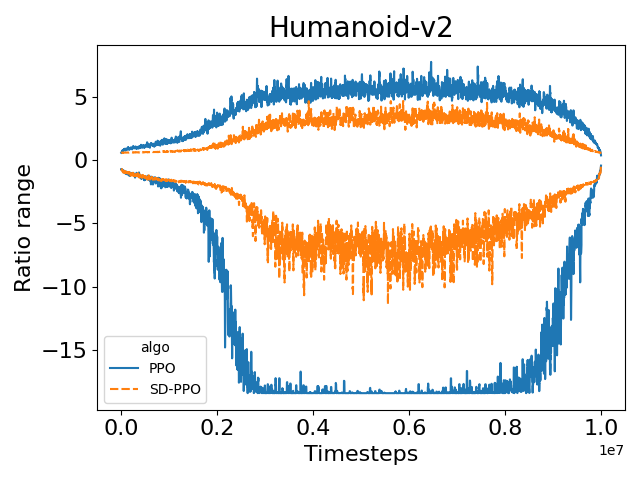}
    \includegraphics[width=0.25\textwidth]{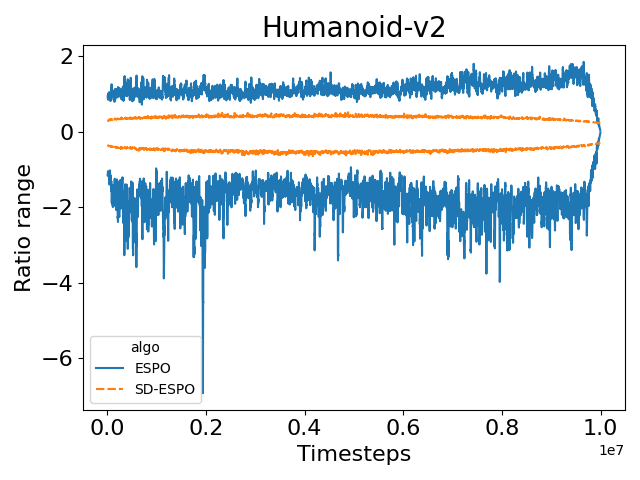}
	\caption{Ratio ranges ($\log (\frac{\tilde{\pi}}{\pi})$) for different algorithms. Algorithms with sample dropout (SD-) have a more concentrated ratio range (orange line) than algorithms without sample dropout (blue line).}
	\label{fig:ratio_range}
\end{figure*}

\subsection{Understanding the effect of sample dropout}

\subsubsection{Reducing variance of surrogate objective estimates}
To verify our theoretical analysis in Theorem~\ref{theo:variance}, we investigate the variance of surrogate objective estimate during training for ESPO and SD-ESPO, as shown in Figure~\ref{fig:reduce_variance}. We use two high-dimensional MuJoCo tasks (Ant-v2, Humanoid-v2) as a case study. 
From our experiments, we confirm that the sample dropout technique can greatly reduce the variance of surrogate objective estimate during training. While the variance of surrogate objective estimate in ESPO keeps high during the training process, sample dropout helps reduce this variance at the very early training stage, and consequently, improves sample efficiency of the agent. Particularly, we find that with sample dropout, the variance of objective remains lower than the original one along with the whole training process and consequently the agent can perform efficient optimization using the estimated surrogate objective with low variance.
The results here are consistent with our theoretical motivation in Section~\ref{sec:methodology} --- by disposing samples with large ratio deviation and keeping the ratios bounded, we can achieve a low variance when optimizing the surrogate objective. 
In addition to checking the variance, we also investigate the bias introduced by the sample dropout technique. Specifically, we measure the bias using averaged ratio deviation over a batch of sample during the multi-epoch policy optimization. Formally, we calculate the average ratio deviation as the following:
\begin{equation}
    \frac{1}{N} \sum_{i=0}^{N-1} \left| \frac{\tilde{\pi}(a_i | s_i)}{\pi(a_i | s_i)} - 1 \right|.
\end{equation}
In Figure~\ref{fig:reduce_variance}, we empirically find that the bias introduced by the sample dropout technique is small and negligible. Overall, sample dropout is a simple yet effective and low-biased technique to reduce variance for the surrogate objective estimate.

\begin{figure*}[h]
	\centering
	\includegraphics[width=0.25\textwidth]{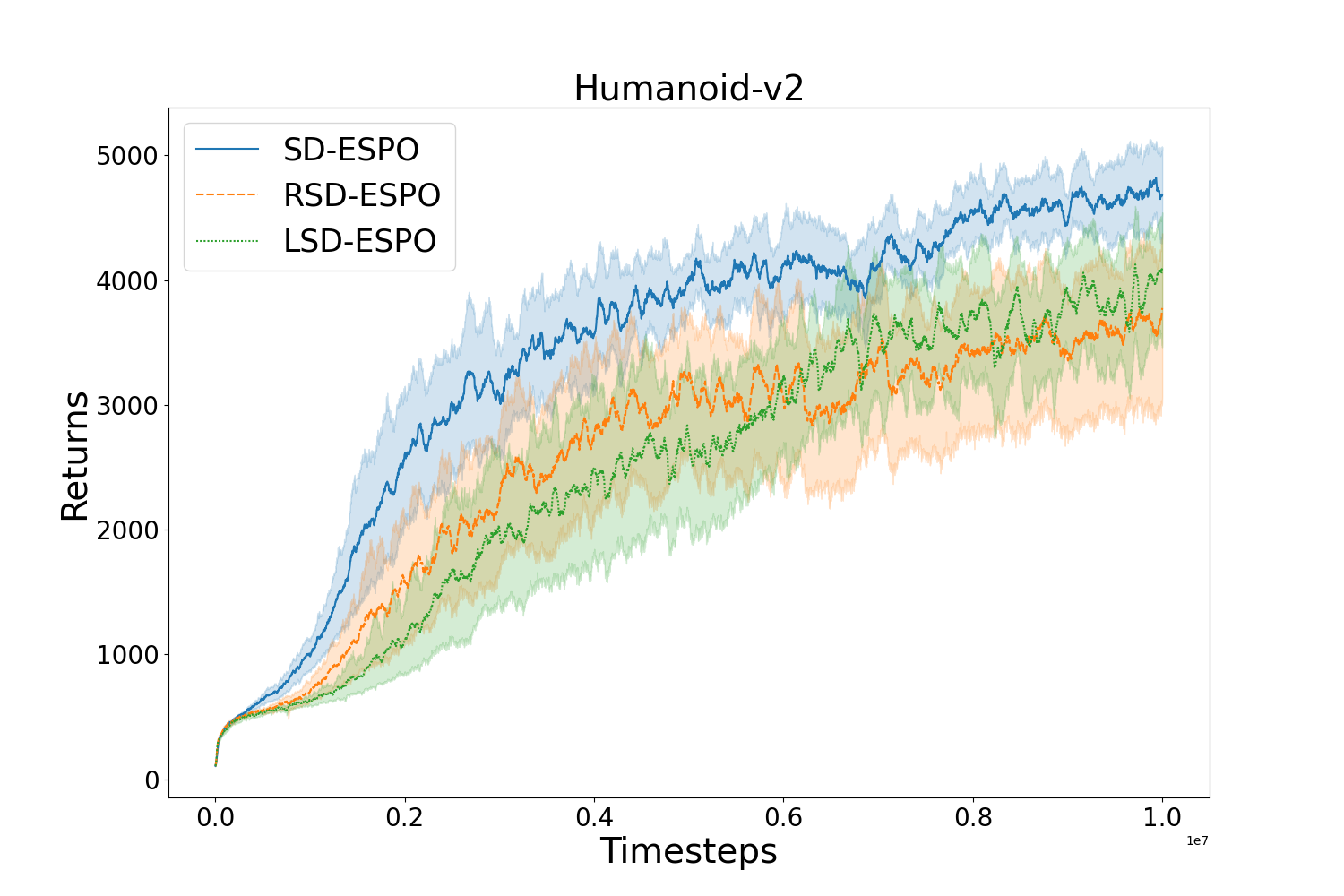}
	\includegraphics[width=0.25\textwidth]{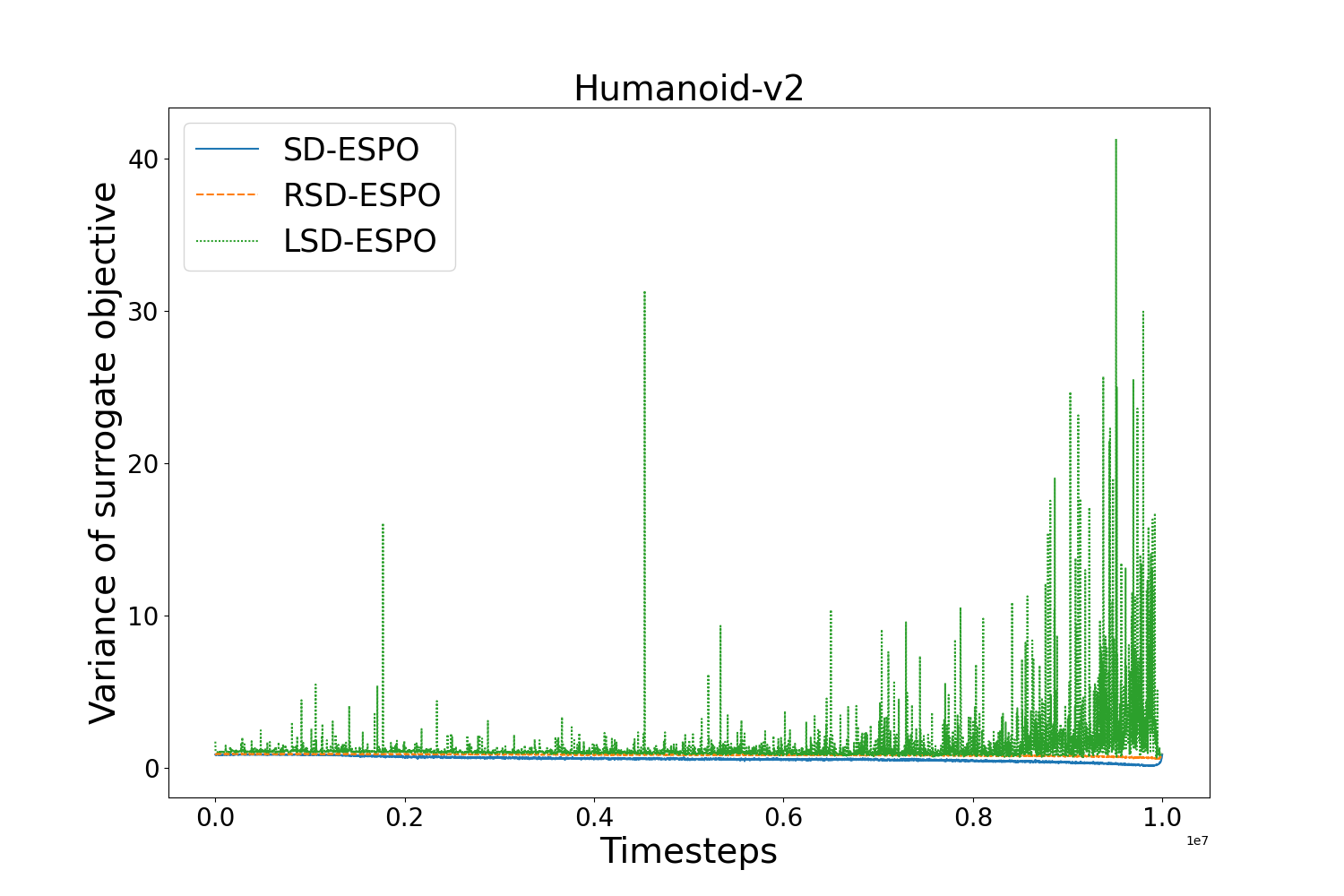}
	\includegraphics[width=0.25\textwidth]{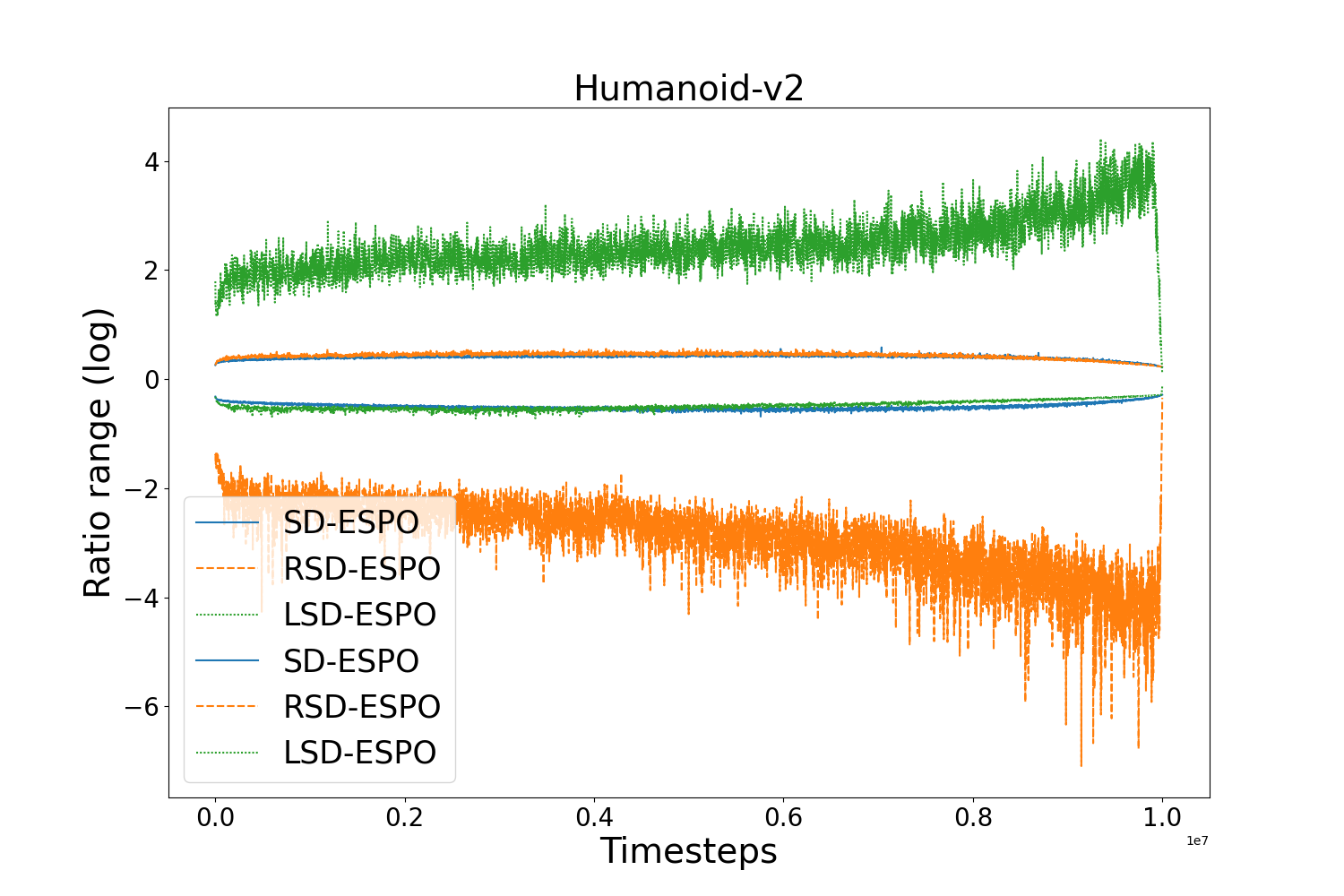}
 
	\includegraphics[width=0.25\textwidth]{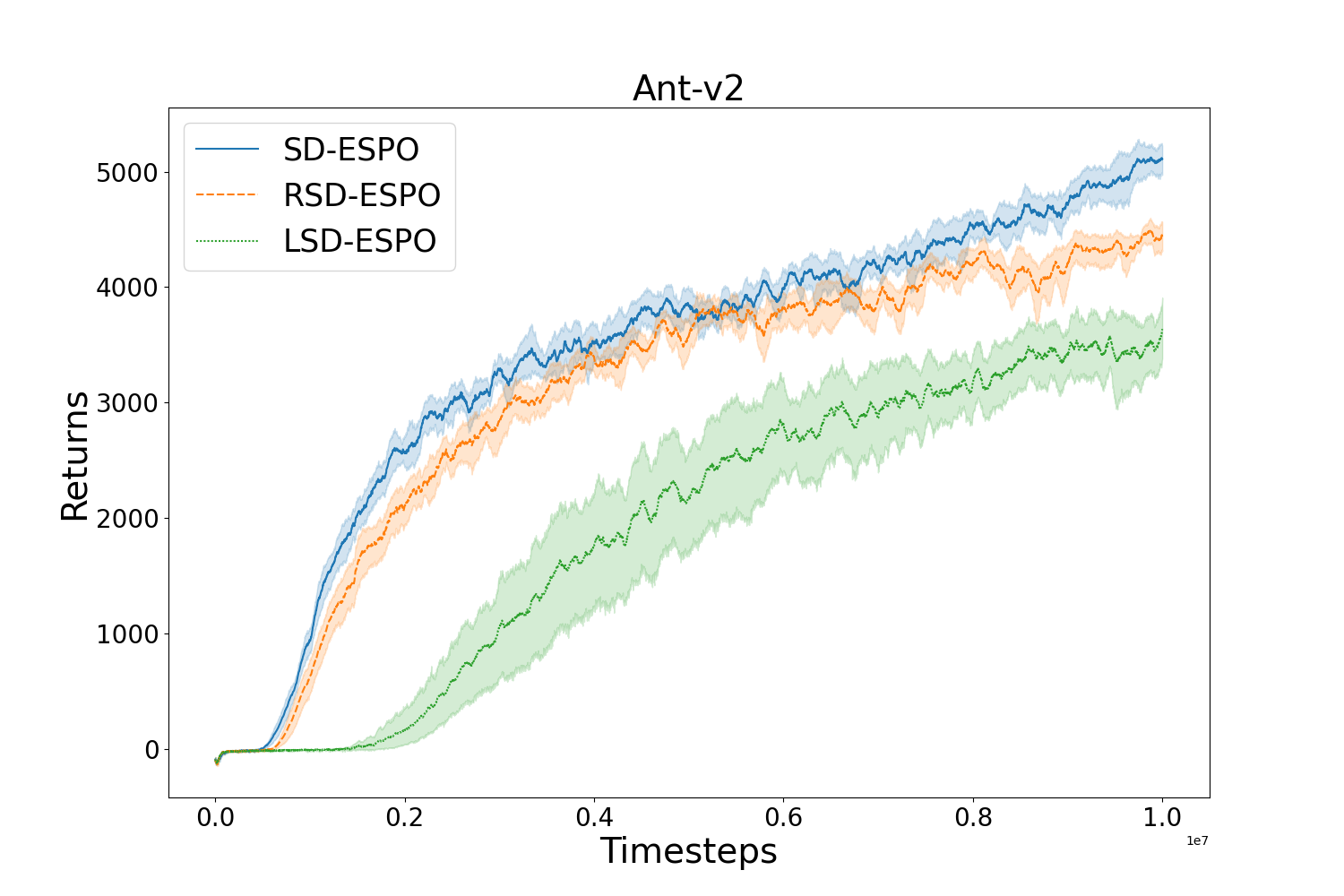}
	\includegraphics[width=0.25\textwidth]{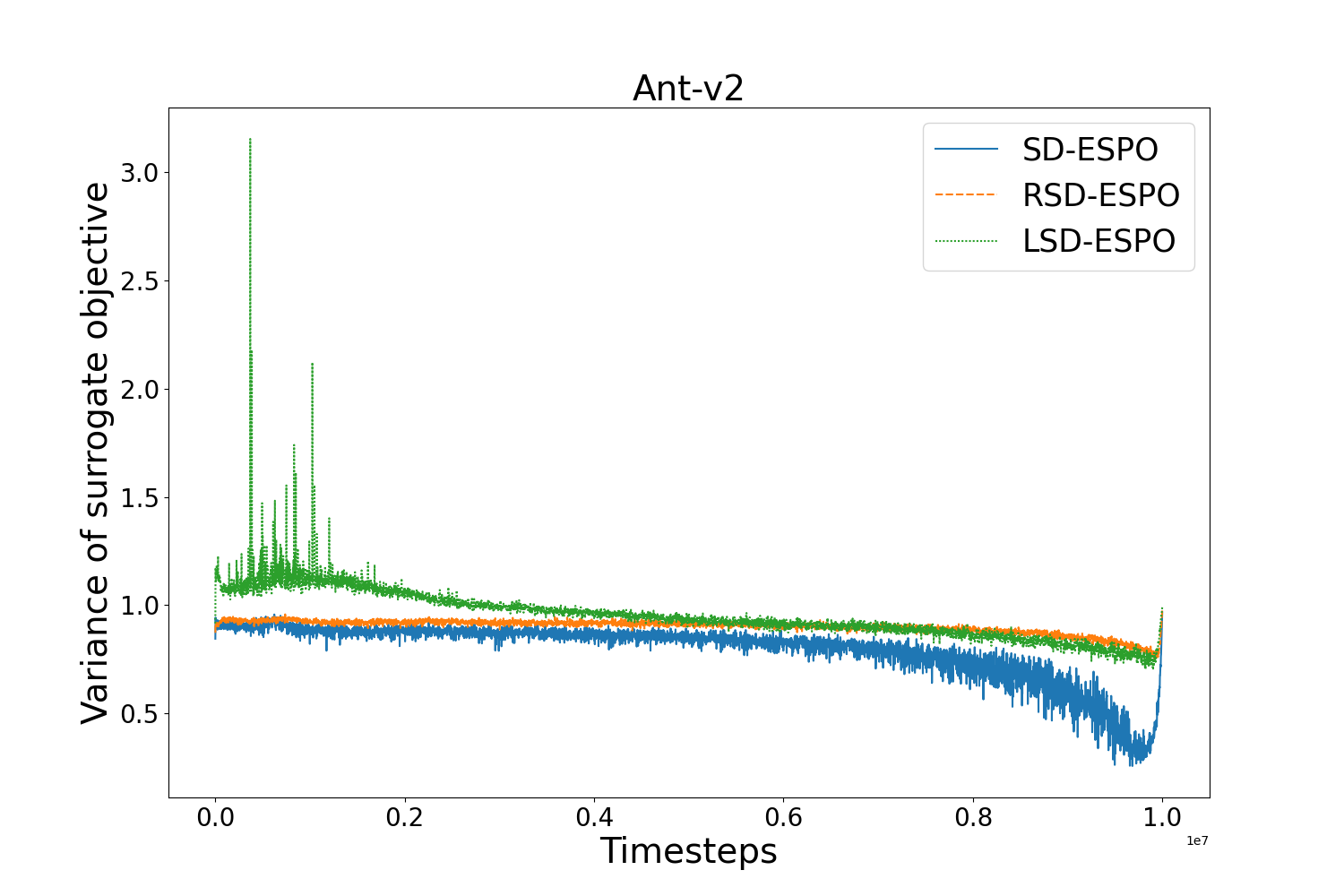}
	\includegraphics[width=0.25\textwidth]{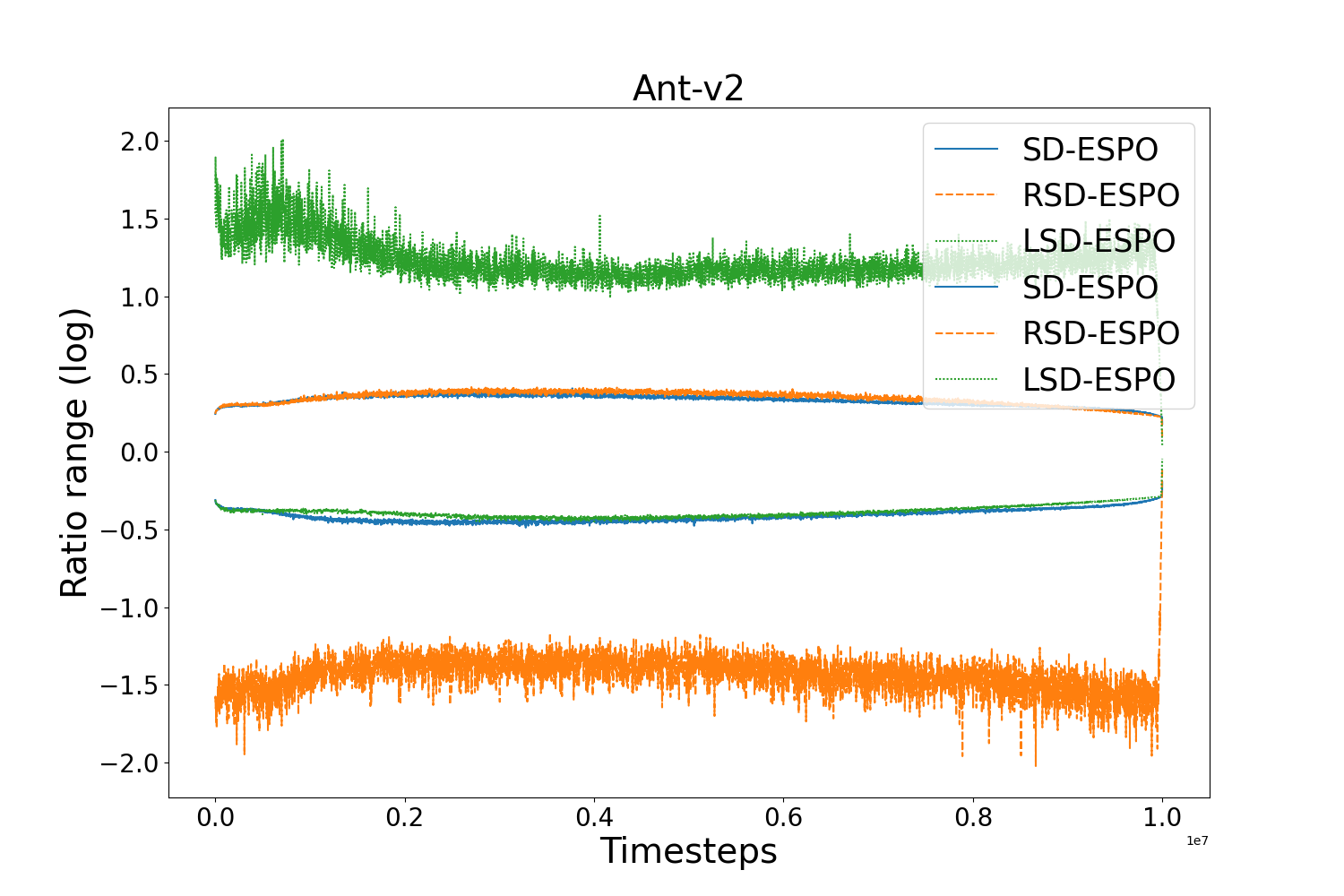}

        \caption{Comparison between two-side sample dropout and the one-side counterparts. Left Column: performance on undiscounted return. Middle column: variance of surrogate objective. Right column: Ratio range (maximum/minimum)  in the batch. The results show that both sides of sample dropout contributes to the performance boost, variance reduction and bounding ratio.}
	\label{fig:one_side}
\end{figure*}

\begin{figure*}[h]
	\centering
	\includegraphics[width=0.22\textwidth]{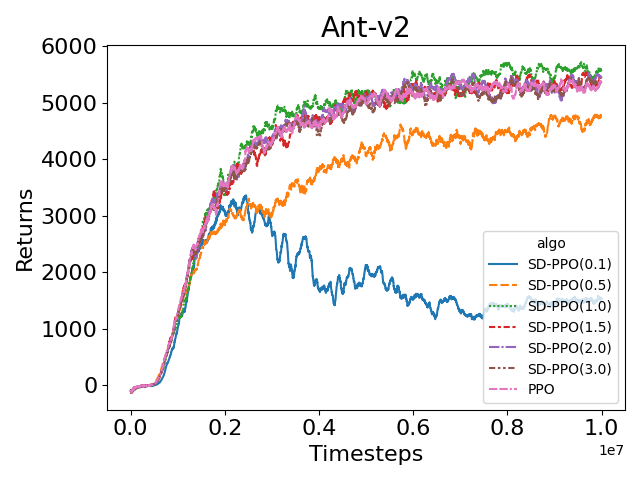}
	\includegraphics[width=0.22\textwidth]{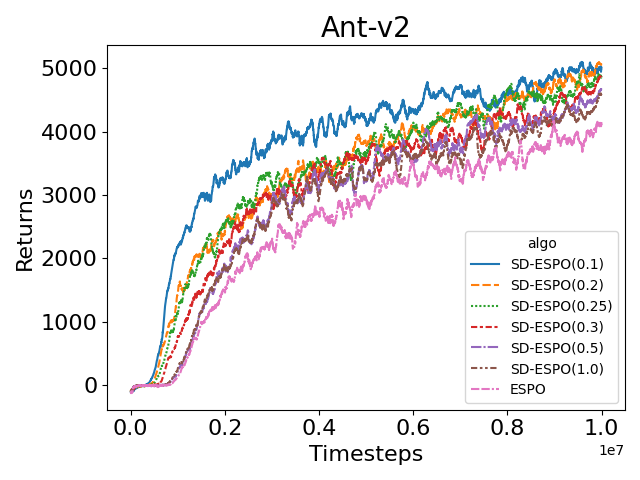}
    \includegraphics[width=0.22\textwidth]{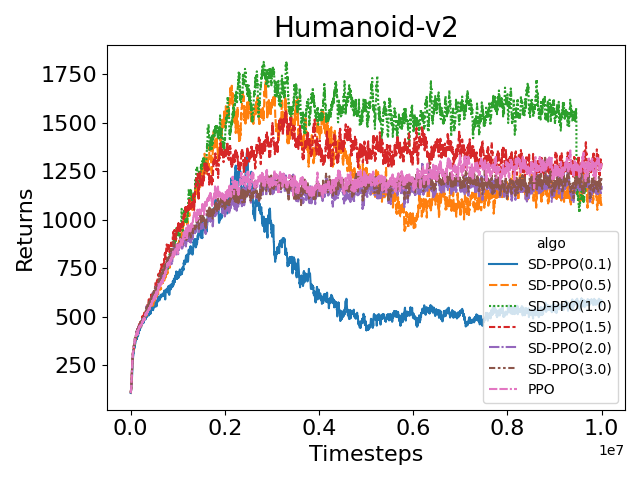}
    \includegraphics[width=0.22\textwidth]{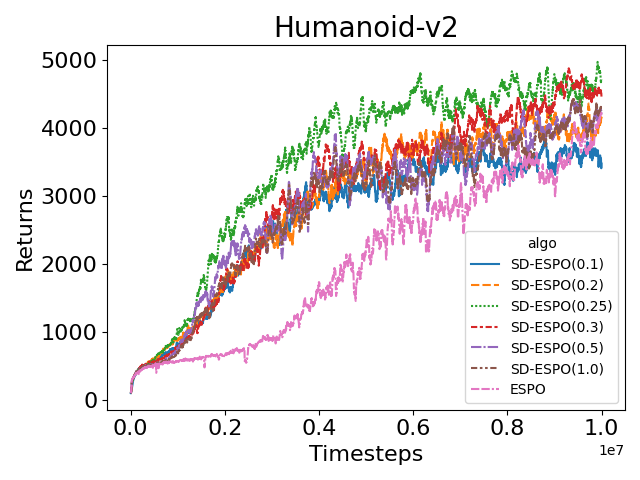}
    \caption{Comparing different degrees of sample dropout: SD-PPO with different $\delta$ and SD-ESPO with different $\delta$. The results show that the sample dropout technique is not very sensitive to the threshold hyperparameter. Sample dropout can boost performance of policy optimization methods when the threshold hyperparameter $\delta$ is set in a rational range.}
    \label{fig:degrees_ppo_espo}
\end{figure*}

\subsubsection{Enforcing the trust-region constraints and proximity}
Retrospecting the theoretical analysis in Section~\ref{sec:methodology}, the aim of sample dropout technique is to keep the ratio deviation small and bound the ratios. 
Therefore, to investigate whether our technique can enforce the trust-region constraint (in TRPO) or the proximity (in PPO and ESPO), we show the ratio range for all algorithms during training, using two MuJoCo tasks as a case study for analysis purpose.
As shown in Figure~\ref{fig:ratio_range}, we find that the ratio ranges of algorithms with sample dropout are more concentrated than those without sample dropout. Without sample dropout, the ratio could become unbounded (e.g., PPO in Humanoid) and thus could increase the variance of surrogate objective estimates. 
We also find that, compared with ESPO (the third column of Figure~\ref{fig:ratio_range}), the ratio range of SD-ESPO is much more stable --- this is in line with the superior performance of SD-ESPO in the Ant and Humanoid environments in Figure~\ref{fig:mujoco}. 
This is a clear indicator that our technique is able to help enforce the trust region/proximity and prevent the objective variance from becoming unbounded.
This also explains why our technique can bring consistent performance boosts for a wide range of policy optimization methods. 
We also notice that the ratio ranges of PPO and SD-PPO in the Ant environment (in the top middle of Figure~\ref{fig:ratio_range}) are close --- this means that the ratio range of PPO in the Ant environment is already well bounded and thus the sample dropout has little effect in this environment. This help explains why SD-PPO achieves close performance with PPO in the Ant environment in Figure~\ref{fig:mujoco}. 

\subsubsection{Two-side sample dropout outperforms the one-side counterparts}
The original proposed sample dropout technique in Section~\ref{sec:methodology} considers disposing samples with too small \textit{and} too large ratios, namely, only performing update on samples satisfying $ 1 - \delta < \frac{\tilde{\pi}}{\pi} < 1 + \delta $, which we denote as two-side sample dropout. A natural question is whether it is necessary to dispose samples from both sides and how each side of sample dropout affects the training performance. Therefore, we aim to investigate the performance of one-side sample dropout by only disposing samples with too small \textit{or} too large ratios. Specifically, we consider two variants: 1) left-side sample dropout (LSD) that disposes samples with too small ratios (i.e., $\leftcond$) and 2) right-side sample dropout (RSD) that disposes samples with too large ratios (i.e., $\rightcond$). We compare the performance on achieved returns and effects of variance reduction among SD-ESPO, LSD-ESPO and RSD-ESPO. The results are shown in Figure~\ref{fig:one_side}. 
First, we find that two-side sample dropout consistently outperforms the one-side counterparts (either left-side or right-side) in terms of sample efficiency and final performance. Second, we discover that the two-side sample dropout has the best effect on variance reduction among all methods, which explains the superior sample efficiency. We also notice that the right-side sample dropout contributes more to the variance reduction --- this is reasonable because in Section~\ref{sec:methodology} we demonstrate that the variance of estimated surrogate objective can grow quadratically with the ratios. With left-side sample dropout only, the ratio can become very large and thus have risks of increasing variance. In our experiment, we find that if we only dispose samples with small ratios but still perform updates on samples with large ratios (i.e., LSD-ESPO), the variance of estimated surrogate objective can become extremely large and hurt the performance. Third, we compare the ratio range between these three methods. On the one hand, in the case of left-side sample dropout, the minimum ratio is bounded while the maximum ratio could become unbounded. On the other hand, with right-side sample dropout, the maximum ratio is bounded and the minimum ratio tends to be unbounded. This indicates that both sides of sample dropout contributes to the performance improvement and it is necessary to consider both sides in sample dropout.



\begin{table*}[t!]
    \centering
    \begin{tabular}{ccccccc}
    \toprule
     & PPO & SD-PPO & ESPO & SD-ESPO & TRPO & SD-TRPO\\
    \midrule
    Alien  & 1363 & \textbf{1608} & 1139 & \textbf{1681} & 23.9 & \textbf{25.0} \\
    Amidar & 540 & \textbf{697} & 192 & \textbf{694} & \textbf{10.9} & 8.8 \\
    Assault & 4774 & \textbf{4916} & 2086 & \textbf{3946} & 11.0 & \textbf{16.3} \\
    Asterix & 3990 & \textbf{4932} & 1294 & \textbf{2984} & 8.6 & \textbf{9.5} \\
    Asteroids & 1542 & \textbf{1635} & \textbf{1489} & 1407 & 6.5 & \textbf{6.9} \\
    Atlantis & 1988645 & \textbf{2172901} & 1091951 & \textbf{1792711} & \textbf{139.1} & 53.5 \\
    BankHeist & \textbf{1165} & 780 & 10.8 & \textbf{11.3} & 0.0 & 0.0 \\
    BattleZone & 21007 & \textbf{23556} & 2532 & \textbf{4336} & 2.0 & \textbf{3.2} \\
    BeamRider & 1948 & \textbf{2457} & 669 & \textbf{885} & 5.6 & \textbf{6.0} \\
    Bowling & 57 & \textbf{69} & 13 & \textbf{25} & 9.9 & \textbf{10.1} \\
    Boxing & 93.5 & \textbf{98.7} & \textbf{62} & 17 & 31.7 & \textbf{60.0} \\
    Breakout & 416 & \textbf{452} & 31 & \textbf{108} & 1.1 & \textbf{1.7} \\
    Centipede & 3604 & \textbf{4138} & 2739 & \textbf{3210} & 27.6 & \textbf{31.6} \\
    Chop.Command & 815 & \textbf{836} & 685 & \textbf{697} & 1.9 & \textbf{2.5} \\
    CrazyClimber & 106472 & \textbf{126192} & 61715 & \textbf{116854} & 0.0 & 0.0 \\
    DemonAttack & 14167 & \textbf{19071} & 433 & \textbf{4349} & 5.2 & \textbf{7.6}  \\
    DoubleDunk & -13.9 & \textbf{-11.9} & -12.2 & \textbf{-11.3} & \textbf{-13.0} & -13.8  \\
    Enduro & 955 & \textbf{1075} & 686 & \textbf{729} & 0.03 & \textbf{13.6}  \\
    FishingDerby & 26.4 & \textbf{28.7} & -77.6 & \textbf{-21.4} & -84.1 & \textbf{-72.3}  \\
    Freeway & 32 & \textbf{34} & 15 & \textbf{34} & 27.0 & \textbf{33.4}  \\
    Frostbite & \textbf{1703} & 340 & 125 & \textbf{298} & \textbf{6.6} & 4.2 \\
    Gopher & 6425 & \textbf{7087} & 978 & \textbf{3472} & 7.0 & \textbf{12.6} \\
    Gravitar & 646 & \textbf{802} & 184 & \textbf{470} & 0.02 & \textbf{0.5} \\
    IceHockey & -4.8 & \textbf{-4.4} & \textbf{-5.1} & -6.5 & \textbf{-7.2} & -12.9 \\
    JamesBond & 511 & \textbf{513} & 379 & \textbf{512} & 0.5 & \textbf{1.9} \\
    Kangaroo & 8571 & \textbf{9211} & 6543 & \textbf{14397} & 3.7 & \textbf{4.1} \\
    Krull & 9362 & \textbf{10703} & 9573 & \textbf{10648} & 255.1 & \textbf{259.3} \\
    Kung-FMaster & 28552 & \textbf{32418} & 27635 & \textbf{30072} & 37.4 & \textbf{43.5} \\
    Monte.Revenge & 0.0 & \textbf{1.3} & 0.0 & \textbf{0.5} & 0.0 & 0.0 \\
    Ms.Pacman & 2279 & \textbf{2546} & \textbf{1791} & 1306 & 30.1 & \textbf{36.9} \\
    NameThisGame & \textbf{6535} & 6157 & \textbf{5113} & 4922 & \textbf{54.9} & 53.6 \\
    Pitfall & -123 & \textbf{-3.2} & -107 & \textbf{-105} & -31.3 & \textbf{-0.1}  \\
    Pong & 20.5 & \textbf{21.8} & 20.3 & \textbf{21.7} & 18.3 & \textbf{19.3} \\
    PrivateEye & 50.5 & \textbf{105} & 34 & \textbf{40} & -123.1 & \textbf{-44.4} \\
    Q*Bert & 15694 & \textbf{18807} & 3137 & \textbf{17622} & 14.5 & \textbf{15.1}\\
    RiverRaid & 8767 & \textbf{9668} & 6727 & \textbf{7666} & 12.0 & \textbf{14.0} \\
    RoadRunner & 39923 & \textbf{53261} & 13601 & \textbf{48953} & 18.7 & \textbf{20.5}\\
    Robotank & 6.9 & \textbf{8.5} & \textbf{4.3} & 3.4 & 0.15 & \textbf{1.8} \\
    Seaquest & 1378 & \textbf{1434} & \textbf{1220} & 880 & 9.9 & \textbf{13.3} \\
    SpaceInvaders & 934 & \textbf{995}& 325 & \textbf{453} & 7.8 & \textbf{10.1} \\
    StarGunner & 42747 & \textbf{46409} & 1516 & \textbf{18517} & 2.2 & \textbf{2.4} \\
    Tennis & \textbf{-16} & -17 & \textbf{-23} & -24 & \textbf{-12.5} & -20.3 \\
    TimePilot  & 5311 & \textbf{5652} & 2834 & \textbf{3936} & 0.9 & \textbf{1.0} \\
    Tutankham & 236 & \textbf{243} & 78 & \textbf{177} & \textbf{8.5} & 7.8 \\
    UpandDown & 153310 & \textbf{284985} & \textbf{87095} & 52507 & 35.93 & \textbf{63.10} \\
    Venture & 0.0 & \textbf{40.95} & \textbf{67} & 0.0 & 0.0 & 0.0 \\
    VideoPinball & \textbf{75567} & 65782 & \textbf{34862} & 34105 & \textbf{37.8} & 28.9 \\
    WizardofWor & 5671 & \textbf{7842} & 1419 & \textbf{3072} & \textbf{0.9} & 0.0 \\
    Zaxxon & \textbf{5380} & 4718 & \textbf{725} & 639 & 0.5 & \textbf{1.3} \\
    \bottomrule
    \end{tabular}
\caption{Full results in Atari games. The results show that sample dropout (SD-) achieves performance boost across a wide range of discrete action tasks on representative policy optimization algorithms including TRPO, PPO and ESPO.} \label{app:all_atari_score}
\end{table*}

\subsection{Comparing different degrees of sample dropout}
We compare different degrees of sample dropout for SD-PPO and SD-ESPO, also using two MuJoCo tasks as a case study. Specifically, we vary the threshold $\delta$ in a wide range, to see how different degrees of sample dropout affect the performance of the algorithms. 
As shown in Figure~\ref{fig:degrees_ppo_espo}, sample dropout works well in SD-PPO when the threshold $\delta$ is set in a rational range.
Meanwhile, if $\delta$ is too small (e.g., 0.1 in SD-PPO), the policy update could become too conservative and suffer from a performance drop in the later training stage. 
The results here show that the sample dropout technique is not very sensitive to this hyperparameter in SD-PPO. 
In Figure~\ref{fig:degrees_ppo_espo}, we examine the performance of SD-ESPO by varying the value of $\delta$ and discover that all tested variants outperform the baselines ESPO. We also find that if the threshold is set too small (e.g., 0.1 in SD-ESPO), the performance of SD-ESPO can vary largely across different tasks. Therefore, we set $\delta$ as 0.25 in SD-ESPO across MuJoCo tasks.

\section{Conclusions}


In this paper, we presented \textit{sample dropout}, a simple yet effective way to boost policy optimization methods. 
We provided analysis for the variance of importance sampling objectives, and
proved that the variance introduced by importance sampling to the objective estimate could become very high and thus jeopardize the effectiveness of the surrogate objective optimization. 
Motivated by the theoretical analysis, we proposed a simple yet effective technique called sample dropout to effectively bound the variance of surrogate objective. 
This technique is simple to implement and we showed that it can serve as a plug-in to improve a wide range of policy optimization methods. 
Extensive experiments across a variety of tasks demonstrated that our technique could significantly boost the performance of policy optimization methods. 
We also conducted additional experiments and analysis to show how our proposed method works. 
In the future, we plan to: 1) apply sample dropout to other policy optimization algorithms; 2) develop better dropout criteria (e.g., fully leverage the distributional information of ratio deviation); 3) cover more complex control tasks. 



\bibliographystyle{IEEEtran}
\bibliography{IEEEabrv,paper}


\ifCLASSOPTIONcaptionsoff
  \newpage
\fi

\begin{IEEEbiography}[{\includegraphics[width=1in,height=1.25in,clip,keepaspectratio]{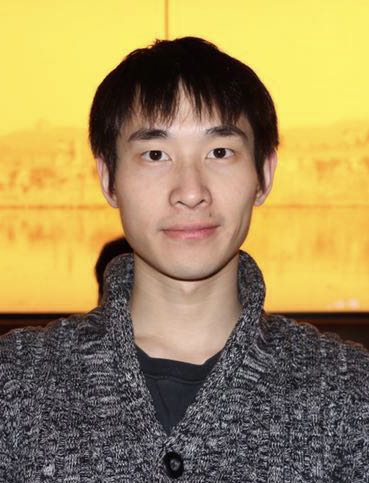}}]{Zichuan Lin} finished his Ph.D. degree from the department of Computer Science and Technology, Tsinghua University, Beijing, China, in 2021. He is now a Researcher at Tencent, Shenzhen, China, working on sample-efficient deep reinforcement learning algorithms. He is interested in machine learning, reinforcement learning as well as their applications on game AI and natural language processing. He has been serving as a PC for NeurIPS, ICML, ICLR and AAAI.
\end{IEEEbiography}

\begin{IEEEbiography}[{\includegraphics[width=1in,height=1.25in,clip,keepaspectratio]{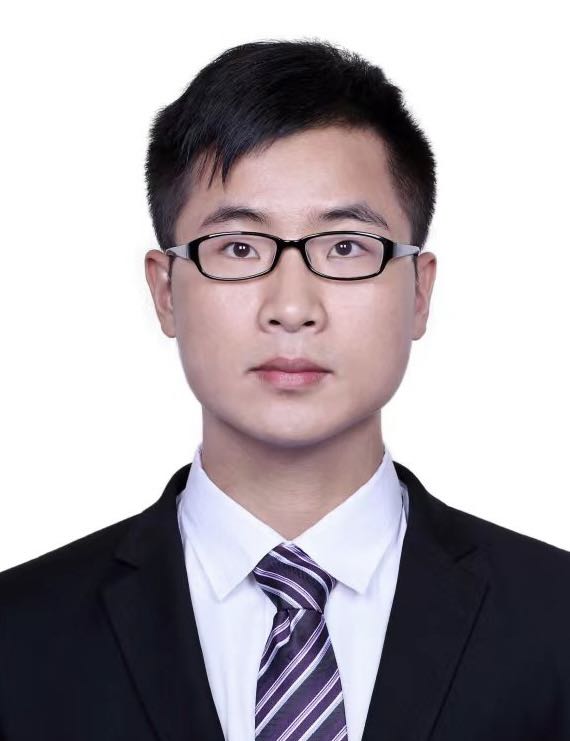}}]{Xiapeng Wu} received his M.Sc. from the Institute of Automation, Chinese Academy of Sciences (IACAS), Beijing, China, in 2020. He is currently a Researcher working on game AI at Tencent AI Lab. His research interests include reinforcement learning and deep learning.
\end{IEEEbiography}

\begin{IEEEbiography}[{\includegraphics[width=1in,height=1.25in,clip,keepaspectratio]{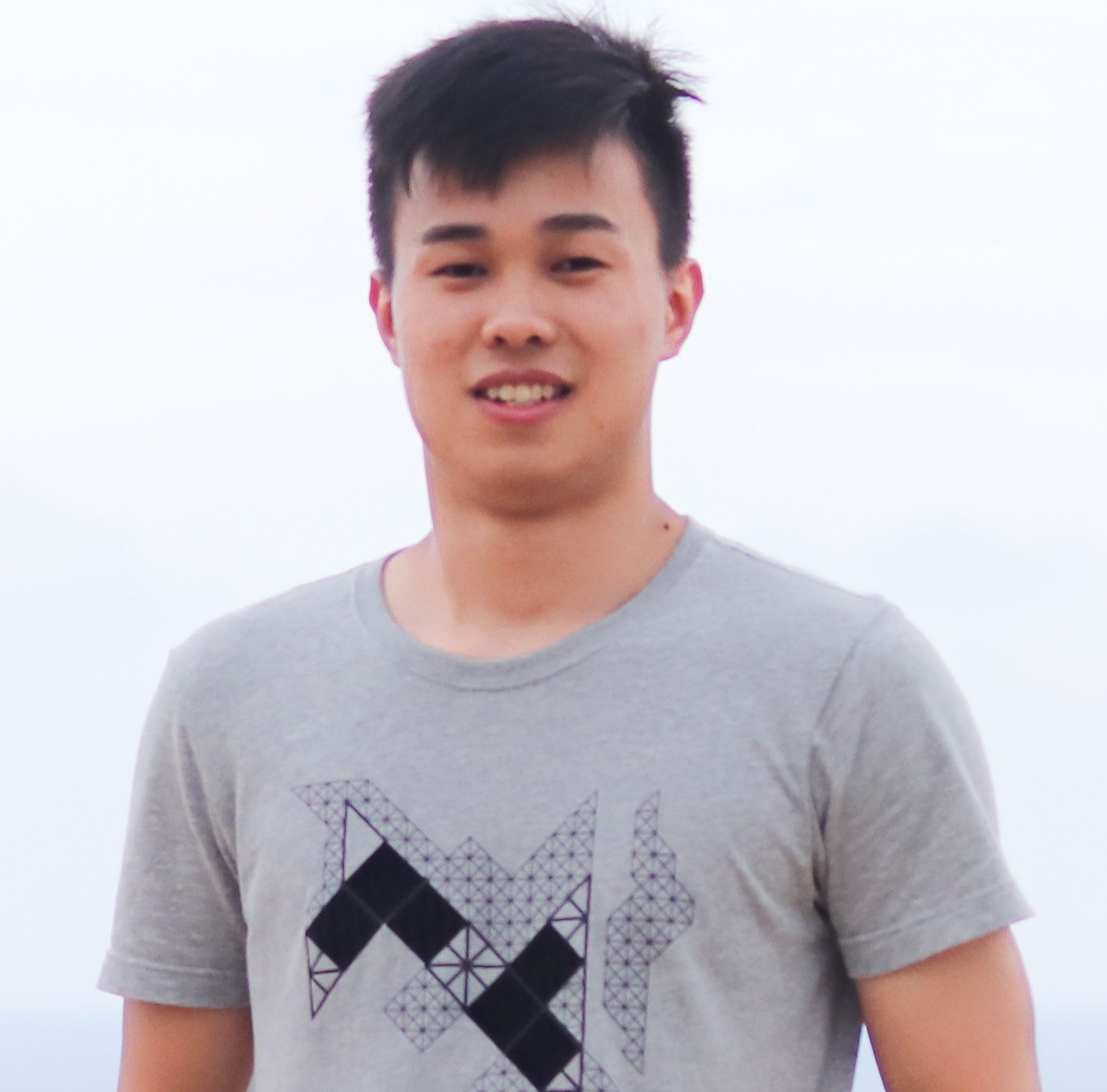}}]{Mingfei Sun} is an assistant professor at the University of Manchester. He is generally interested in artificial intelligence, with a particular focus on machine learning techniques such as deep reinforcement learning and learning from demonstrations. His research covers large-scale deep reinforcement learning, policy gradient methods, and optimization theory. He is applying these methods to problems in video games, robotics and multi-agent systems. Before joining Manchester, Mingfei Sun was a researcher at Microsoft Research Cambridge. He was also a research associate at University of Oxford. Mingfei Sun obtained his PhD degree from the Hong Kong University of Science and Technology (HKUST). 
\end{IEEEbiography}

\begin{IEEEbiography}[{\includegraphics[width=1in,height=1.25in,clip,keepaspectratio]{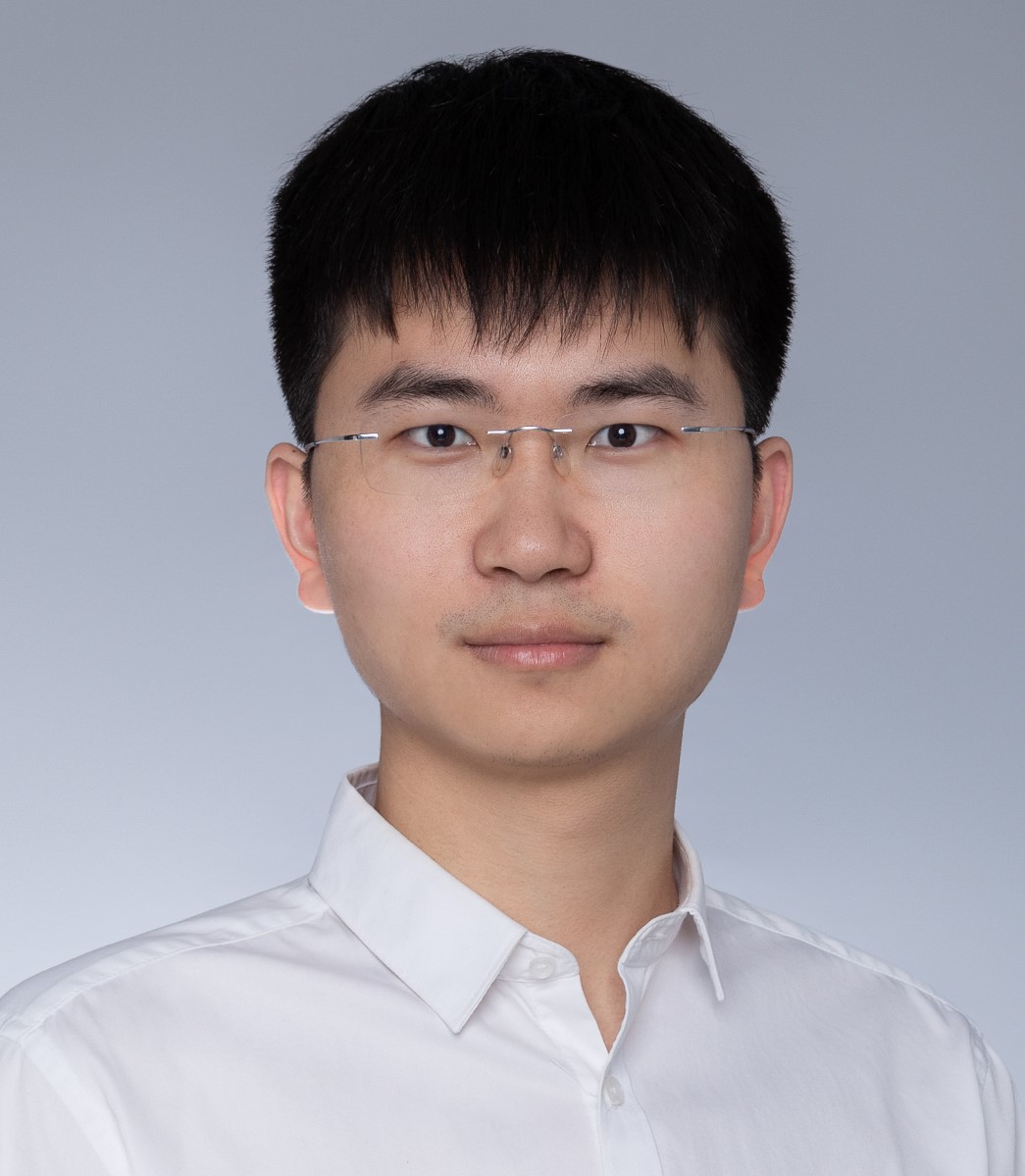}}]{Deheng Ye} finished his Ph.D. from the School of Computer Science and Engineering, Nanyang Technological University, Singapore, in 2016. He is now a Principal Researcher and Team Manager with Tencent, Shenzhen, China, where he leads a group of engineers and researchers developing large-scale learning platforms and intelligent AI agents. He is interested in applied machine learning, reinforcement learning, and software engineering. He has been serving as a PC/SPC for NeurIPS, ICML, ICLR, AAAI, and IJCAI.
\end{IEEEbiography}

\begin{IEEEbiography}[{\includegraphics[width=1in,height=1.25in,clip,keepaspectratio]{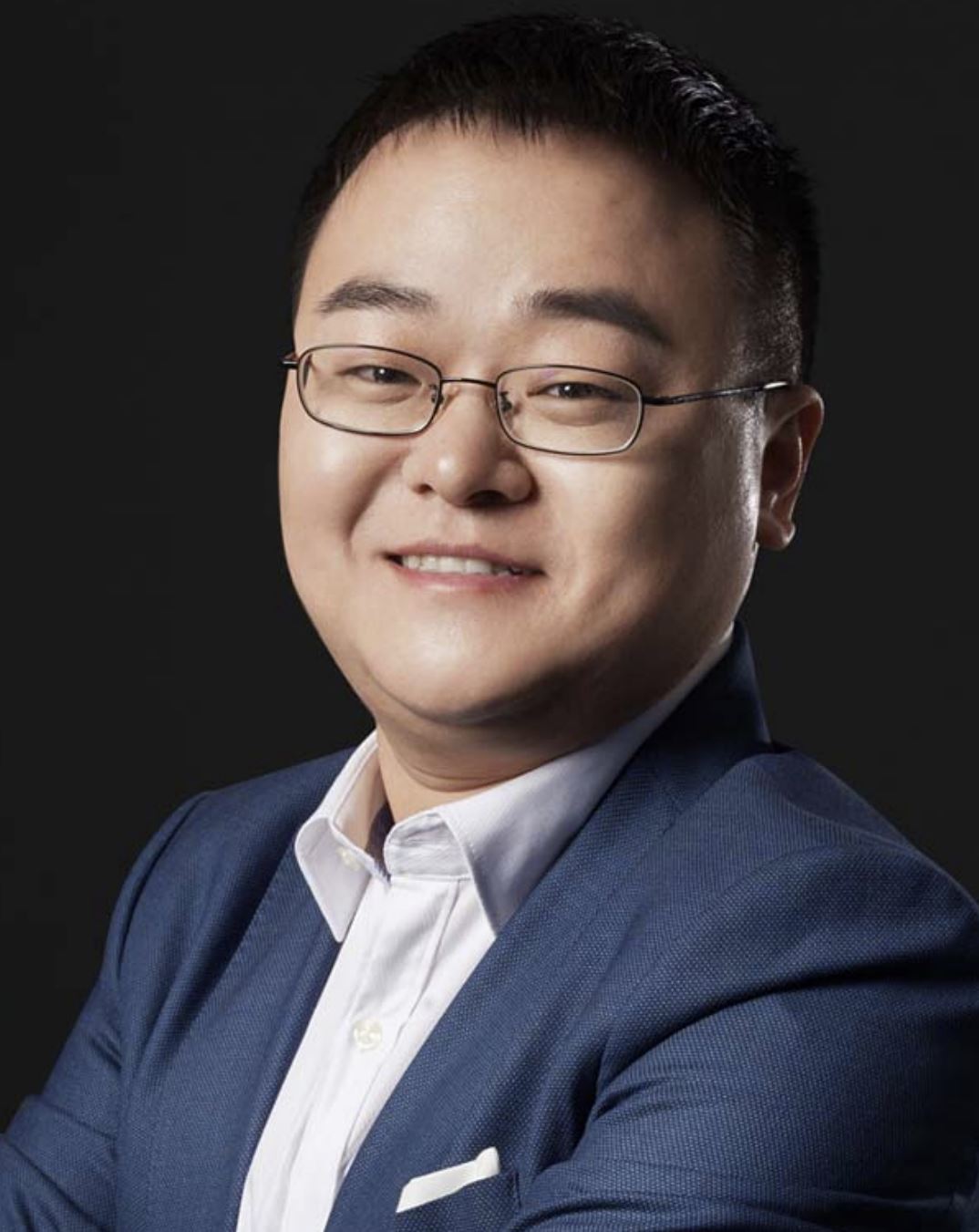}}]{Qiang Fu} received the B.S. and M.S. degrees from the University of Science and Technology of China, Hefei, China, in 2006 and 2009, respectively. He is the Director of the Game AI Center, Tencent AI Lab, Shenzhen, China. He has been dedicated to machine learning, data mining, and information retrieval for over a decade. His current research focus is game intelligence and its applications, leveraging deep learning, domain data analysis, reinforcement learning, and game theory.
\end{IEEEbiography}

\begin{IEEEbiography}[{\includegraphics[width=1in,height=1.25in,clip,keepaspectratio]{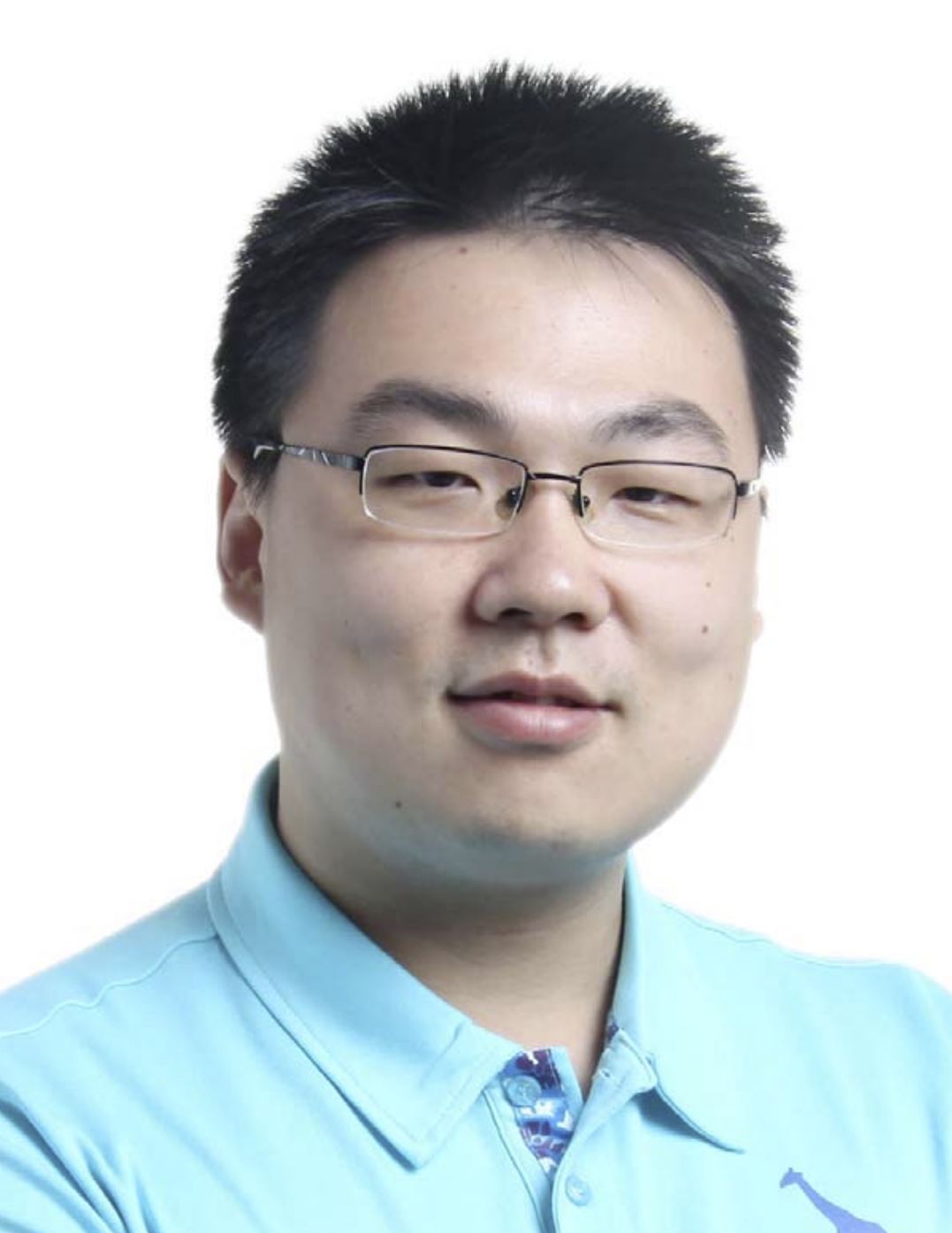}}]{Wei Yang} received the M.S. degree from the Huazhong University of Science and Technology, Wuhan, China, in 2007. He is currently the General Manager of the Tencent AI Lab, Shenzhen, China. He has pioneered many influential projects in Tencent in a wide range of domains, covering Game AI, Medical AI, search, data mining, large-scale learning systems, and so on.
\end{IEEEbiography}

\begin{IEEEbiography}[{\includegraphics[width=1in,height=1.25in,clip,keepaspectratio]{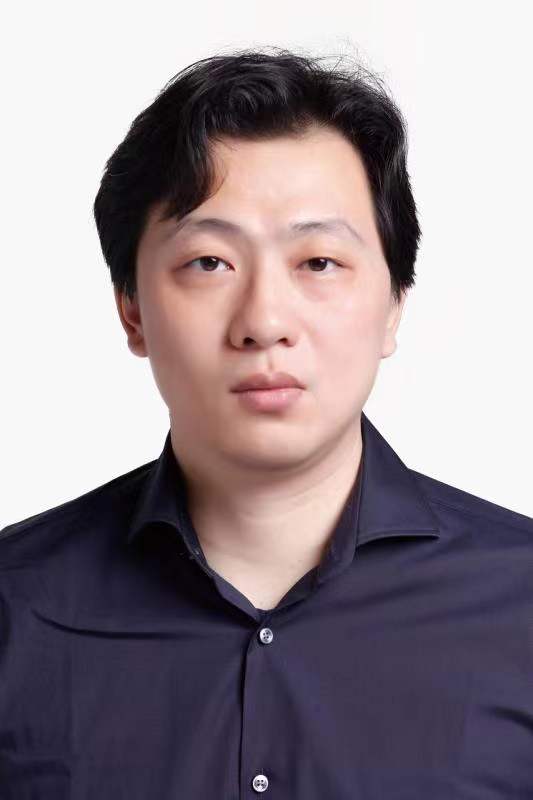}}]{Wei Liu} (M'14-SM'19) is a Distinguished Scientist of Tencent and the Director of Ads Multimedia AI at Tencent Data Platform. Dr. Liu has long been devoted to fundamental research and technology development in core fields of AI, including deep learning, machine learning, computer vision, pattern recognition, information retrieval, big data, etc. To date, he has published extensively in these fields with more than 260 peer-reviewed technical papers, and also issued 16 US patents. He currently serves on the editorial boards of IEEE TPAMI, TNNLS, IEEE Intelligent Systems, and Transactions on Machine Learning Research. He is an Area Chair of top-tier computer science and AI conferences, e.g., NeurIPS, ICML, IEEE CVPR, IEEE ICCV, IJCAI, and AAAI. Dr. Liu is a Fellow of the IEEE, IAPR, AAIA, IMA, RSA, and BCS, and an Elected Member of the ISI.
\end{IEEEbiography}




\clearpage

\end{document}